\documentclass[10pt, letter, onecolumn]{arxiv}

\usepackage{kantlipsum, lipsum}
\usepackage{dm-colors}
\usepackage{amsmath}
\usepackage{pstricks, pst-node}
\usepackage{verbatim}
\usepackage{multirow}
\usepackage{scalerel}
\usepackage{booktabs}
\usepackage{enumitem}
\usepackage{xspace}
\usepackage{bm}
\usepackage{bbm}
\usepackage{mathtools}
\usepackage{soul}
\usepackage{epsfig}
\usepackage{graphicx}
\usepackage{amssymb}
\usepackage{colortbl}
\usepackage{csquotes}
\usepackage{setspace}
\usepackage{colortbl}
\usepackage{tabularx,ragged2e}
\usepackage{placeins}
\usepackage[symbol]{footmisc}
\usepackage{nameref}
\usepackage{varioref}

\usepackage[pagebackref=false,breaklinks=false,colorlinks=true,bookmarks=true,citecolor=teal,urlcolor=red,linkcolor=blue]{hyperref}

\usepackage[capitalize]{cleveref}

\usepackage{etoc}
\usepackage[authoryear]{natbib}
\setcitestyle{authoryear,round,citesep={;},aysep={,},yysep={;}}

\graphicspath{{figures/}}

\definecolor{purple}{RGB}{100,0,200}
\definecolor{LightRed}{rgb}{1,0.92,0.92}
\definecolor{LightOrange}{rgb}{1,0.95,0.88}
\definecolor{LightYellow}{rgb}{1.0,1.0,0.84}
\definecolor{LightGreen}{rgb}{0.9,1.0,0.88}
\definecolor{LightCyan}{rgb}{0.9,1,1}
\definecolor{LightBlue}{rgb}{0.9,0.94,1}

\title{\large{How Well Does GPT-4V(ision) Adapt to Distribution Shifts? A Preliminary Investigation}}

\vspace{1cm}
\author[1$\ast$]{Zhongyi Han} 
\author[2$\ast$]{Guanglin Zhou}
\author[3$\ast$]{Rundong He}
\author[4]{Jindong Wang}
\author[5]{Tailin Wu}
\author[3]{Yilong Yin}
\author[1,6]{Salman Khan}
\author[7,2,8]{Lina Yao}
\author[9,1]{Tongliang Liu}
\author[10,1]{Kun Zhang}

\affil[1]{\normalsize Mohamed bin Zayed University of Artificial Intelligence \hspace{6pt}}
\affil[2]{\normalsize The University of New South Wales \authorcr  \vspace{2pt}}
\affil[3]{\normalsize School of Software, Shandong University \vspace{2pt}}
\affil[4]{\normalsize Microsoft Research Asia \vspace{2pt}}
\affil[5]{\normalsize Westlake University \authorcr \vspace{2pt}}
\affil[6]{\normalsize Australian National University \vspace{2pt}}
\affil[7]{\normalsize Data61, CSIRO  \vspace{2pt}}
\affil[8]{\normalsize Macquarie University  \vspace{2pt}} 
\affil[9]{\normalsize The University of Sydney \vspace{2pt}}
\affil[10]{\normalsize Carnegie Mellon University \vspace{2pt}} 

\renewcommand{\correspondingauthor}[1]{$\ast$~These authors contributed equally to this work. \\
Email: \{hanzhongyicn;jameszhou.ustc\}@gmail.com}

\begin{document}

\begin{abstract}
\section*{\centering Abstract}

In machine learning, generalization against distribution shifts---where deployment conditions diverge from the training scenarios---is crucial, particularly in fields like climate modeling, biomedicine, and autonomous driving. The emergence of foundation models, distinguished by their extensive pretraining and task versatility, has led to an increased interest in their adaptability to distribution shifts. GPT-4V(ision) acts as the most advanced publicly accessible multimodal foundation model, with extensive applications across various domains, including anomaly detection, video understanding, image generation, and medical diagnosis. However, its robustness against data distributions remains largely underexplored. Addressing this gap, this study rigorously evaluates GPT-4V's adaptability and generalization capabilities in dynamic environments, benchmarking against prominent models like CLIP, LLaVA, and Gemini. We delve into GPT-4V's zero-shot generalization across 13 diverse datasets spanning natural, medical, and molecular domains. We further investigate its adaptability to controlled data perturbations and examine the efficacy of in-context learning as a tool to enhance its adaptation. Our findings delineate GPT-4V's capability boundaries in distribution shifts, shedding light on its strengths and limitations across various scenarios. Importantly, this investigation contributes to our understanding of how AI foundation models generalize to distribution shifts, offering pivotal insights into their adaptability and robustness. The code is publicly available at \url{https://github.com/jameszhou-gl/gpt-4v-distribution-shift}.

\end{abstract}

\maketitle

\begin{figure}[h] 
\centering
\includegraphics[width=.6\textwidth]{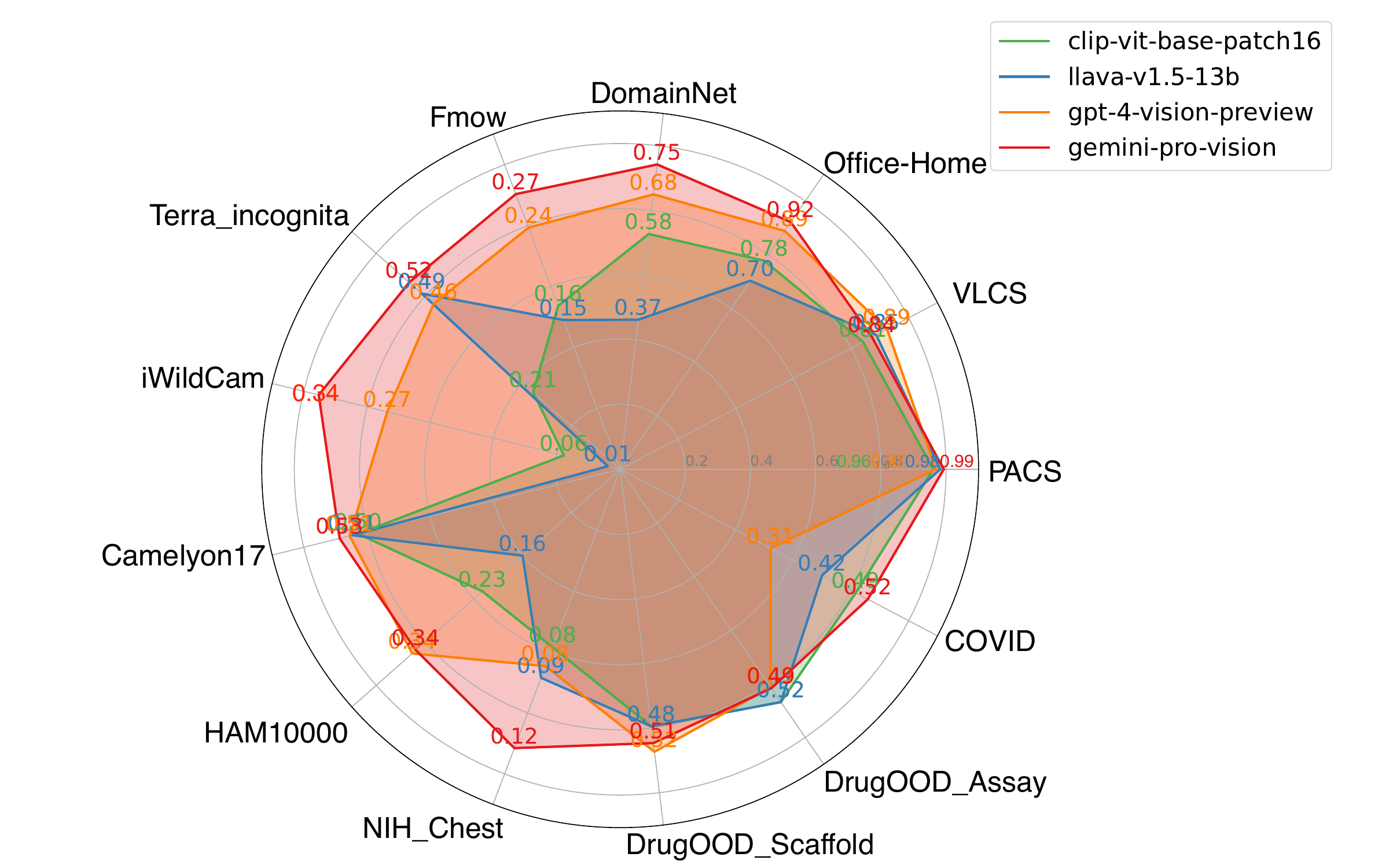}
\caption[Comparative Analysis of Zero-Shot Generalization Across Diverse Domains]
{Comparative analysis of zero-shot generalization performance across 13 distinct datasets, encompassing natural, medical, and molecular domains. The analysis features the performances of three advanced models: CLIP, LLaVA, GPT-4V and Gemini.}
\label{751719482771}
\end{figure}


\tableofcontents

\newpage

\listoffigures

\newpage


\section{Introduction}

\subsection{Motivation and Overview}
\label{sec:motivation}

In the evolving landscape of machine learning, the challenge of distribution shift emerges as a pivotal concern, often regarded as one of the main problems in improving the generalization ability. At the core of this issue lies the independent and identically distributed (i.i.d.) assumption, foundational yet frequently violated in practical scenarios. Mathematically, distribution shift denotes a divergence between the training data distribution and the data encountered in real-world applications, a discrepancy that can significantly impair model performance. This phenomenon is ubiquitously observed across diverse domains, from climate modeling \citep{knutti2010challenges, zwiers2013climate} and biomedicine \citep{park2021reliable, cascarano2023machine, stacke2020measuring, huang2021machine} to wildlife conservation \citep{tuia2022perspectives, ellis2011predicting, beery2018recognition}, autonomous driving \citep{stocco2022confidence}, and financial forecasting \citep{mashrur2020machine}. The omnipresence of distribution shifts underscores a fundamental limitation: even an algorithm capable of perfectly fitting all available data lacks practical utility without the ability to generalize in a dynamic, ever-changing world. This challenge is further compounded by the significant performance degradation observed in numerous machine learning algorithms when confronted with distribution shifts, despite their otherwise notable successes. The traditional machine learning models, while adept in stable environments, struggle with evolving or varying data distributions, highlighting an urgent need for algorithms that can adaptively maintain accuracy in the face of such shifts \citep{wang2022generalizing, yue2019domain, prakash2019structured, huang2021fsdr, qiao2020learning, liu2018unified, li2018domain, ganin2015unsupervised, ganin2016domain, rojas2018invariant, sun2021recovering, christiansen2021causal, gulrajani2020search, wiles2022a}.

Significant seminal studies have been made in addressing distribution shifts. In the realm of domain adaptation, numerous studies have demonstrated remarkable successes, particularly when the target distribution is accessible during the training process~\citep{sun2016deep,ganin2015unsupervised,cite:CVPR2017OfficeHome,cite:ICCV2019DomainNet,han2022towards,han2022learning}. This capability to adapt to known shifts has been a key driver in advancing models' robustness across varied datasets. It is not an isolated case that a plethora of domain generalization methods are designed through various strategies such as domain alignment~\citep{muandet2013domain,li2018deep}, causal representation learning~\citep{scholkopf2021toward}, stable learning~\citep{zhang2021deep}, and invariance-based optimization~\citep{liu2021heterogeneous}.

Despite the proliferation of research in distribution shifts as discussed earlier, the field is experiencing a significant trend: the emergence of foundation models as robust solutions \citep{bommasani2021opportunities, zhang2022delving, wei2023stronger,zheng2023large,tu2023many}. This wave is driven by the capabilities derived from extensive pretraining, comprehensive data understanding, substantial training data volumes, and large-scale network parameters \citep{kumar2022fine, li2022sparse, rame2023model, du2021fewshot, lee2022surgical, li2022simple}. The CLIP model \citep{radford2021learning} stands out as a prominent example, showing remarkable robustness against natural distribution shifts \citep{recht2019imagenet, wang2019learning, barbu2019objectnet, hendrycks2021natural, hendrycks2021many} primarily due to its extensive training on a wide range of image-text pairs. Notable in this context are models like BLIP~\citep{li2022blip,li2023blip}, LLaVA \citep{liu2023visual, liu2023improved}, and Gemini \citep{team2023gemini}, each contributing unique strengths to the realm of foundation models. Concurrently, foundation models are evolving from specialized tools to versatile, general-purpose assistants, demonstrating their adaptability in various downstream tasks \citep{awais2023foundational, li2023multimodal}. In this evolving landscape, GPT-4V(ision)\footnote{Hereafter referred to as "GPT-4V"} and its successor GPT emerge as cutting-edge examples of these general-purpose foundation models \citep{gpt4v}, particularly in their ability to handle distribution shifts. Central to our research is the pivotal question: {\bf "How well does GPT-4V(ision) adapt to distribution shifts?"}. 
 
Despite the growing fascination with GPT-4V and its wide-ranging applications, a significant gap is evident in current research: the assessment of its adaptability to distribution shifts remains underexplored. This gap is particularly striking, considering GPT-4V's extensive deployment across various domains, each presenting unique data challenges. The importance of robust performance in out-of-distribution scenarios cannot be overstated; failures in these contexts can lead to critical errors, especially in high-stakes fields such as medical diagnosis or autonomous driving, where accurate and reliable predictions are imperative. While recent studies have showcased GPT-4V's effectiveness in anomaly detection \citep{cao2023towards}, optical character recognition \citep{shi2023exploring}, video understanding \citep{lin2023mm}, image generation \citep{yang2023idea2img}, zero-shot visual recognition~\citep{wu2023gpt4vis}, and medical diagnosis \citep{wu2023can}, the critical question of its performance under distribution shifts has not been addressed. This study is orthogonal to these existing works, focusing on an uncharted yet crucial aspect of machine learning robustness: GPT-4V's ability to adapt to evolving data environments.

This paper is the first to evaluate GPT-4V on distribution shifts. Through our investigation, we seek to unravel the intricacies of GPT-4V's performance in various distribution shift scenarios and various applications, thereby providing some insights into its robustness and adaptability. This exploration is not just about probing the depths of GPT-4V’s capabilities; it also aims to broaden our understanding of the potential and limitations of multimodal foundation models in navigating the complexities of real-world data scenarios. This exploration is poised to contribute to the evolution of AI systems, paving the way for more resilient and versatile applications.

Our investigation of GPT-4V in distribution shift scenarios is guided by the following questions:

\begin{enumerate}
    \item \emph{How effectively does GPT-4V manage distribution shifts across diverse domains?} 
    We seek to measure the extent of GPT-4V's zero-shot adaptability to the unique distribution shifts inherent in diverse domains. We aim to evaluate how GPT-4V understands and reacts to changes in data distribution, benchmarking its performance against models like CLIP \citep{radford2021learning}, known for robustness to natural distribution shifts, LLaVA \citep{liu2023visual, liu2023improved}, an open-sourced multimodal foundation model, and Gemini \citep{team2023gemini}, a competitive LMM with GPT-4V.

    \item \emph{How does GPT-4V react to deliberate alternations in data distribution?}
    The traditional approach to distribution shifts typically considers a model's ability to generalize from a source domain to a target domain, with an inherent shift in data distribution.
    However, in a zero-shot context, this adaptability may diverge from the conventional settings, as the test data could largely differ from GPT-4V's pre-training data.
    Given the opacity of its pre-training data, we investigate its reaction to distribution shifts that we engineer.
    We first select images from domains where GPT-4V has initially exhibited promising performance, indicative of these images likely to align with or be part of its pre-training distribution.
    Subsequently, we intend to introduce Gaussian noise and implement stylistic transformation to these chosen images. 
    These manipulations serve as a means to create specific distribution shifts, allowing us to assess the model's generalization capabilities under these controlled perturbations.

    \item \emph{Is in-context learning an effective method to augment GPT-4V's adaptation to distribution shifts?} 
    Conventional approaches to distribution shifts in foundation models typically involve tuning model parameters, often through methods such as efficient tuning or fine-tuning \citep{hu2021lora}. 
    Considering the impracticality of fine-tuning GPT-4V's vast parameters, we turn to the potential of in-context learning, a technique at the heart of emergent abilities in the large language models \citep{brown2020language}, as an alternative to simulate traditional domain generalization methods. 
    This apporach entails utilizing representative images from source domain classes as in-context examples, followed by introducing a test image from a novel target domain. 
    This investigation centers on the capacity of in-context examples to improve GPT-4V's performance in the face of distribution shifts.

\end{enumerate}

\subsection{Our Approach in Exploring GPT-4V}
\subsubsection{How Do We Treat Distribution Shifts in This Work?}

In the realm of machine learning, distribution shifts pose a formidable challenge, particularly when deploying models in real-world scenarios that differ from the training environment. 
Traditional approaches to this issue involve fine-tuning pre-trained foundation models on source domain data to adapt them to target domains. 
However, when it comes to massive models like GPT-4V, this conventional approach encounters significant hurdles. 
The vast scale of GPT-4V's architecture makes standard fine-tuning methods impractical, while the opacity nature of its pre-training data adds layers of uncertainty regarding its performance in novel scenarios. 
In response to these challenges, our study adopts a nuanced and multifaceted approach, aiming to thoroughly examine GPT-4V's adaptability to distribution shifts. This involves employing three distinct evaluation strategies: 
(1) \textbf{Zero-shot Generalization:} 
In Section \ref{353244693620}, we evaluate GPT-4V's inherent zero-shot generalization capabilities.
Similar to models like CLIP, we assess GPT-4V's performance across different domains without prior tuning or exposure to relevant domain data, reflecting a purview into the model's natural adaptability.
(2) \textbf{Response to Data Perturbations:} 
In Section \ref{057362393941}, our study investigates GPT-4V's robustness when faced with artificially induced shifts in data characteristics, focusing on domains where it shows initial high performance. 
(3) \textbf{In-Context Learning as a Domain Bridge:} 
In Section \ref{018520640299}, we assess GPT-4V's use of in-context learning to simulate conventional domain generalization paradigms, highlighting its ability to apply contextual understanding from the source domain to interpret data from the target one.

This multifaceted strategy is designed to illustrate GPT-4V's adaptability comprehensively, from its generalization capacity in comparison with baselines to its performance under artificially induced shifts and its ability to utilize contextual learning as a means to bridge domain gaps.

\subsubsection{Sample Selection Guidance for GPT-4V Evaluation}
\label{944004671334}
To conduct a meaningful evaluation of GPT-4V within the constraints of the OpenAI API’s rate limits, we have devised a sample selection strategy that prioritizes both diversity and informative value. 
Our selection process is guided by the following principles.  

\textbf{Random Sampling for Diversity.} 
Our process begins with a random selection of samples from each class across all domains within our 13 datasets, intending to capture the inherent diversity of each domain, reflecting varying complexities and content types.
To ensure comprehensive coverage, we employ two distinct sample sizes for each dataset: 180 and 1800. This decision aligns with OpenAI's revised rate limit policies, increasing daily requests from 100 to 500 as of December 2, 2023.
Consequently, our sampling strategy, constrained by the limits of 100 and 500 requests per day for each account, strategically includes approximately 180 and 1800 random selections. 
Although these numbers might appear limited for certain datasets, they represent a balance between operational efficiency and practical feasibility under existing constraints. 
Notably, our preliminary findings indicate a consistent performance trend when comparing the two sample sizes, as shown in Tables~\ref{617085393024} and~\ref{medical_science}. 
Our goal is to minimize selection bias and provide a thorough evaluation of GPT-4V’s performance across a broad spectrum of data.
    
\textbf{Inclusion of Failure Cases From CLIP.} 
To further enrich our evaluation, we have deliberately chosen to incorporate 180 instances for each dataset, where the CLIP model exhibits underperformance. 
This focused selection is driven by a specific objective: to assess how GPT-4V handles challenges that have proven difficult for a well-established model like CLIP. 
By analyzing GPT-4V's performance in these particular scenarios, we aim to gain deeper insights into its relative strengths and adaptability compared to CLIP.
It is noteworthy that failure cases are sourced from CLIP due to its established role as a baseline model, particularly noted for its zero-shot robustness against distribution shifts. While a similar analytical approach using LLaVa's failure cases presents a valuable avenue for future research, it remains outside the scope of our current study.

Recognizing the continuous evolution of LMMs, the cases we have selected are designed to function as a benchmark for evaluating and tracking the adaptability of state-of-the-art foundation models to distribution shifts. This benchmark not only serves our current study but also contributes to the broader research community\footnote{The test cases employed in our study have been publicly available at \url{https://huggingface.co/datasets/jameszhou-gl/gpt-4v-distribution-shift}}.

\begin{figure}[tb!]
\centering
\includegraphics[width=.9\textwidth]{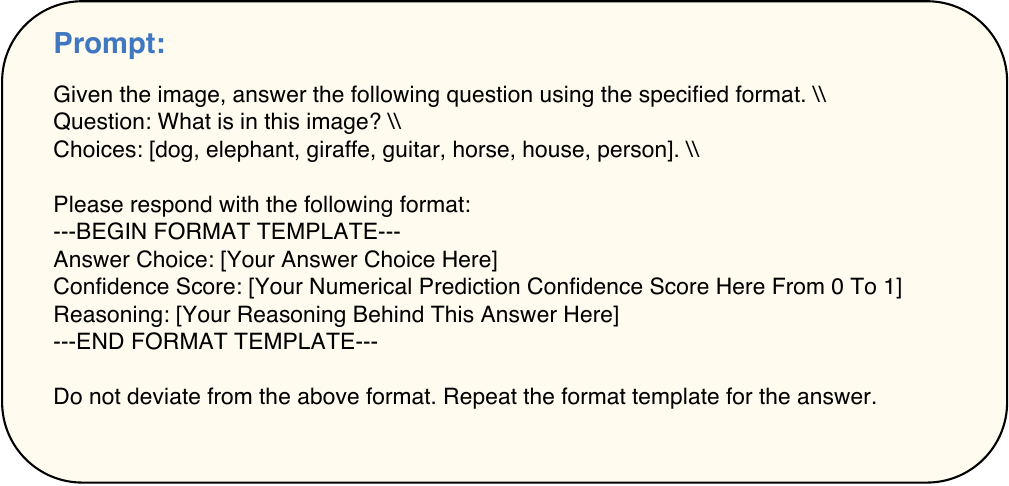}
\caption[Illustration of a structured prompt format]
{An illustration of a structured prompt format used in the PACS dataset, showcasing a specific approach for image-based questioning and response formatting.
The format includes a question about the image's content, a list of answer choices, and a template for answering, including an answer, confidence score, and the reasoning process.}
\label{291135271633}
\end{figure} 

\subsubsection{Prompt Designs}
In transforming conventional classification tasks into a visual question answering (VQA) format, our focus has been to devise a prompt template that is straightforward yet sufficiently informative. 
This approach seeks to exploit the expressive capabilities of language, a strength evidenced in previous models such as GPT-2 \citep{radford2019language} and GPT-3 \citep{brown2020language}.  
Crucially, our prompt design is tailored to a fair comparison of the inference abilities of GPT-4V and LLaVA. 
Specifically, we have developed a prompt that pairs an image with a clear, direct question, such as `What is in this image?' followed by a set of answer choices.
This design is intended to maintain simplicity, focusing primarily on the model’s ability to interpret and accurately respond to visual content.
Moreover, GPT-4V and LLaVA are prompted not just to select an answer option but also to provide a confidence score and a rationale for the decision, enhancing the depth of our analysis.

As exemplified in Figure \ref{291135271633}, our structured prompt serves several key purposes:
\begin{itemize}
    \item \textbf{Simplicity:} By employing a straightforward template that contextualizes the image with basic question and answer choices, we ensure minimal complexity in the prompt design.
    \item \textbf{Uniformity:} The approach ensures consistency and standardization in the model's responses, which is vital for comparative analysis across diverse test scenarios.
    \item \textbf{Insight into Reasoning:} The inclusion of confidence scoring and rationale requirements leverages GPT-4V's ability to output the decision-making process, thereby providing valuable insights into its reasoning and improving the interpretability of its outputs.
\end{itemize}

\subsection{Contributions of This Report}
\begin{itemize}
\item \textbf{First Exploration:} This paper marks the first comprehensive study into the adaptability of GPT-4V to distribution shifts. Our research fills a notable gap in the existing literature, as no prior studies have systematically explored this aspect of GPT-4V. This investigation not only highlights the novelty of our approach but also sets a precedent for future research in evaluating and improving the robustness of multimodal foundation models in the face of data variability.

\item \textbf{Quantitative Benchmark Results and Comparison:}  Before our work, explorations into the performance of models like GPT-4V were largely confined to qualitative or case-study approaches. Our study stands out as the first to provide a quantitative evaluation of GPT-4V's adaptability to distribution shifts. This quantitative approach allows us to present concrete, measurable insights into the model's capabilities, setting a benchmark for future research and development in this area. Our work also provides a quantitative comparison between GPT-4V and pioneering foundation models.

\item \textbf{Deep Analysis and Insights:} We have conducted rigorous experiments across various domains and scenarios to assess GPT-4V's performance under distribution shifts. These experiments provide valuable insights into the model's zero-shot adaptability, its response to engineered distribution shifts, and the effectiveness of in-context learning and style transfer in enhancing its robustness. Our empirical analysis offers a detailed understanding of how GPT-4V navigates complex, real-world data scenarios, contributing significantly to the field of AI and machine learning.

\end{itemize}

\subsection{Limitations of This Report}

Here, we discuss several limitations in our evaluation of GPT-4V:
\begin{itemize}
    \setlength\itemsep{1em}
    \item \textbf{Sample Bias.} One of the primary limitations in our assessment of GPT-4V, as well as CLIP and LLaVA, is the presence of sample bias. While we employed random sampling to enhance diversity and reduce selection bias, as detailed in Section \ref{944004671334}, eliminating sample bias remains challenging. Our sampling process ensured a comprehensive representation of each class across domains. However, due to the rate limits imposed by the OpenAI API, our usage of GPT-4V\footnote{Namely, the gpt-4-vision-preview model via the OpenAI API.} was restricted to a finite number of requests per day. 
    Consequently, we limited our selection to 1,800 cases per dataset. 
    This constraint might result in the derived performance metrics not fully capturing the models' true capabilities or weaknesses, particularly in processing novel, complex, or varied inputs. Such a limitation is significant, as it potentially leads to overestimations or underestimations of the model's practical utility and robustness in real-world scenarios.

    \item \textbf{Potential Challenge in VQA Format Transformation.} 
    Another limitation arises from converting conventional classification tasks into a VQA format. 
    As illustrated in Figure~\ref{291135271633}, this approach requires inputting all class names into GPT-4V and prompting it to select one from the list. 
    However, this method becomes impractical for datasets with a large number of classes, such as the ImageNet dataset, which contains 1000 classes \citep{recht2019imagenet,wang2019learning,shankar2021image,hendrycks2021natural}. 
    The VQA format in such cases would necessitate an excessively high token count, posing significant challenges in terms of feasibility and efficiency.

    \item \textbf{Failed query may happen in GPT-4V.} When verifying the robustness of GPT-4V to distributional shift via the OpenAI API, it is possible that a query failure may occur. The reasons include: 1) Using the API may lead to query failure due to reaching usage limits, API service malfunctions, or network issues. 2) The input data has a significant distributional shift, causing the model to fail to understand or respond correctly, leading to query failure. This query failure leads to a difference in the denominator when conducting quantitative assessments, between GPT-4V and other models like CLIP and LLaVA. For example, in Table~\ref{617085393024}'s random test, both CLIP and LLaVA are at 180, whereas GPT-4V is less than 180 due to query failures.
            
\end{itemize}

In summary, while our evaluation may not be exhaustive, we believe that this analysis offers valuable insights for both researchers and medical professionals, it sheds light on the current capabilities of the multimodal foundational model and may inspire future work towards building medical foundation models.

\section{Observations}
In this section, we summarize our primary observations from the extensive experiments conducted on GPT-4V, addressing the adaptability of this advanced multimodal foundation model to distribution shifts. 

\begin{itemize}
    \item \textbf{General Performance Across Domains:} In Section \ref{353244693620}, across various domains, GPT-4V showcased robust performance, particularly evidencing resilience to natural image distribution shifts. Nevertheless, its proficiency waned in more specialized fields like medicine and chemistry, signaling potential areas for enhancement. This was notably apparent in datasets such as Camelyon17, NIH-Chest, DrugOOD Assay, and DrugOOD Scaffold, where GPT-4V's classification outcomes resembled random guesses, as detailed in Table~\ref{medical_science}. This pattern suggests a need for targeted improvements in these domain-specific contexts.
    
    \item \textbf{Adaptability to Controlled Data Perturbations:} The experiments in Section~\ref{057362393941} utilizing ControlNet-generated and random noise-induced data distributions presented GPT-4V with entirely novel domains, distinct from those involved in its pretraining phase. This setup rigorously tests the model's generalization capabilities in handling out-of-distribution scenarios. As demonstrated in Table~\ref{control}, GPT-4V almost surpassed other methods in its performance, excelling particularly with challenging samples and in situations where CLIP encountered failures. These results underscore GPT-4V's exceptional stability and reliability when confronted with controlled perturbations and novel data distributions, highlighting its robust generalization abilities.

    \item \textbf{In-context Learning Is an Effective Method:} The experiments detailed in Section~\ref{018520640299} illuminate the efficacy of in-context learning in enhancing GPT-4V's adaptability to distribution shifts. Notably, in the case studies depicted in Figure~\ref{351022164706}, GPT-4V demonstrates its capability to accurately identify the class of pathological images by discerning differences compared to two source images. This adaptability was consistently mirrored across four distinct datasets, reinforcing the utility of in-context learning strategies in navigating distribution shifts. Looking forward, there is a promising avenue for developing more sophisticated in-context learning methods, aiming to further bolster GPT-4V's robustness across diverse data distributions.
    
    \item \textbf{Detail-Oriented Classification Rationale:} The classification rationale provided by GPT-4V reflects a nuanced and detailed understanding of image elements, illustrating its sophisticated content comprehension. For instance, as exemplified in Figure~\ref{057815033513}, GPT-4V's capability outshines that of LLaVA by accurately recognizing distinct characteristics such as a robust body, short tail, and tufted ears. These instances clearly demonstrate GPT-4V's advanced ability to discern and articulate finer details in images, further reinforcing its superiority in complex image classification tasks under distribution shifts.

    \item \textbf{Higher Confidence in Predictions:} GPT-4V consistently displayed higher and more justified confidence levels in its predictions, indicative of a confident and precise decision-making process. As illustrated in Figure~\ref{057815033513}, GPT-4V's detail-oriented classification rationale contributes to its generating higher confidence scores compared to LLaVA. For instance, in Figure~\ref{169039403841}, GPT-4V achieves a peak confidence score with a descriptive analysis: ``The image shows a metal kettle with a spout, handle, and thermometer on the top, which is a common design for a kettle used to heat water." Conversely, in medical imaging scenarios, such as depicted in Figure~\ref{HAM10000_random_case}, GPT-4V's confidence scores are more moderate, often accompanied by recommendations for further clinical testing, reflecting a prudent approach in high-stakes contexts.

    \item \textbf{Need for Domain-Specific Fine-Tuning:} GPT-4V's performance in fields requiring specialized knowledge, such as medicine, chemistry, and biology, highlights the need for further fine-tuning using domain-specific data. While GPT-4V often provides rational and contextually appropriate reasoning, it can still yield incorrect classifications or diagnoses. A case in point is Figure~\ref{289347172711}, where GPT-4V accurately describes an image labeled as a guitar, stating that ``the image displays a stylized depiction of a guitar ... leading to high confidence in this identification,'' yet it incorrectly classifies the image as a person. This example underscores the critical need for domain-specific fine-tuning, especially in areas where precision and reliability are paramount. Incorporating domain-specific knowledge and data into GPT-4V could substantially improve its accuracy, ensuring that its sophisticated reasoning consistently aligns with accurate contextual interpretations and decisions.
    
    \item \textbf{Consistency in Challenging Samples:} GPT-4V showcased remarkable consistency in handling challenging samples, particularly in scenarios where CLIP encountered errors. Its performance was notably superior to that of LLaVA, exhibiting enhanced adaptability and precision. This is clearly evidenced in Tables~\ref{617085393024} and~\ref{medical_science}, where, in instances of failure cases, GPT-4V almost outperforms both LLaVA and CLIP by a significant margin. These findings highlight GPT-4V's robustness and efficacy in dealing with complex samples, especially those involving significant distribution shifts.

    \item \textbf{Limitations in Applicability for Certain Tasks:} GPT-4V struggles with classification tasks when labels lack semantic information. This limitation becomes evident in scenarios such as activity identification tasks involving chemical molecular structures. In these cases, where sample labels are simply `active' or `inactive,' both GPT-4V and LLaVA tend to perform no better than random guessing. The provided reasoning, such as ``The image shows a chemical structure, which does not have an active or inactive state in the context of physical motion or activity," as highlighted in Table~\ref{medical_science} and Figure~\ref{drugood_assay_random_case}, reveals a gap in context comprehension. Similarly, tasks with numerical labels also pose a challenge for GPT-4V’s zero-shot classification capabilities. These findings underscore the need for additional adaptation or fine-tuning for downstream tasks that involve non-semantic labels.    
    \item \textbf{Consistent Improvements as the Model Evolves:} Initially, our evaluation focused on GPT-4V, a leading-edge LMM. During this phase, the introduction of Gemini presented an opportunity to broaden our analysis. Consequently, we integrated Gemini, facilitating a more comprehensive comparison. This progression, from CLIP, LLAVA, and GPT-4V to Gemini, represents a remarkable trajectory of continuous enhancement in zero-shot generalization capabilities across a wide range of natural domains. Particularly, Gemini showcases near state-of-the-art performance, as evidenced in Tables \ref{617085393024} and \ref{table:all_methods}. While our focus remains on GPT-4V's performance, the inclusion of Gemini enriches our analysis. Additionally, the codes and datasets are publicly available, encouraging ongoing validation and study of the latest advancements in LMMs.
\end{itemize}

\section{Zero-shot Generalization Across Varied Domains}
\label{353244693620}
This section delineates our findings on the zero-shot generalization capabilities of GPT-4V in the context of distribution shifts, as enumerated in Table \ref{617085393024} and \ref{medical_science}. 
We compare the performance of GPT-4V with baseline models such as CLIP\footnote{{https://huggingface.co/openai/clip-vit-base-patch16}}, LLaVA\footnote{{https://huggingface.co/liuhaotian/llava-v1.5-13b}}, and Gemini Pro Vision\footnote{https://ai.google.dev/models/gemini}, highlighting its effectiveness and limitation across a variety of domains.
Our investigation categorizes the datasets into three distinct groups: natural visuals, medical images, and molecular images. 
For each category, we first provide an overview of the collective results, showcasing GPT-4V's generalization performance. 
This is followed by in-depth case studies, where we delve into specific instances to uncover nuanced insights about the model's performance in diverse and challenging scenarios.

\subsection{Natural Images}

\begin{table*}[htb]
\setlength{\abovecaptionskip}{0.cm}
  \caption{Summary of zero-shot generalization performance across various natural datasets, showcasing the comparative results of GPT-4V (gpt-4-vision-preview) with CLIP (clip-vit-base-patch16), LLaVA (llava-v1.5-13b) and Gemini (gemini-pro-vision) models.
}
  \label{617085393024}
\centering
\scalebox{0.75}{
\begin{tabular}{lccccccc}
\toprule
Dataset & PACS & VLCS & Office-Home & DomainNet  & Fmow & TerraIncognita & iWildCam\\  
\midrule
Category & natural & natural & natural & natural & natural & natural & natural\\
Prediction & animal species & animal species & everyday items & objects, creatures  & land use & animal species & animal species\\
Domain & artistic media & image repositories & visual categories & artistic styles & time, region & camera trap & location\\
\#domains & 4 & 4 & 4 & 6 & 6 & 4 & 206\\
\#classes & 7 & 5 & 65 & 345 & 62 & 10 & 323 \\
Examples & \includegraphics[width=2cm]{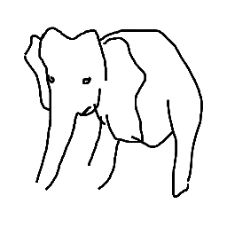} & \includegraphics[width=2cm]{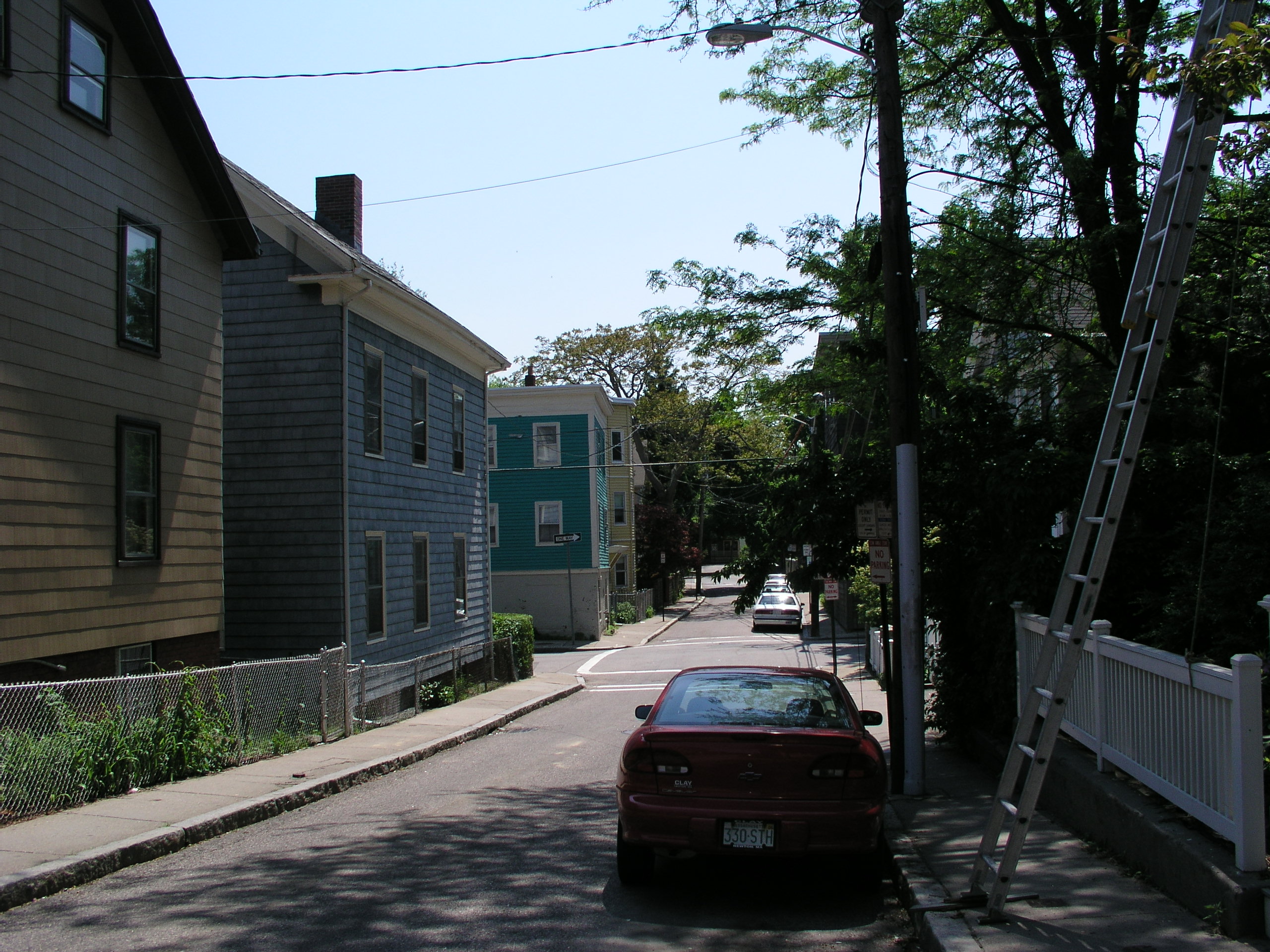}& \includegraphics[width=2cm]{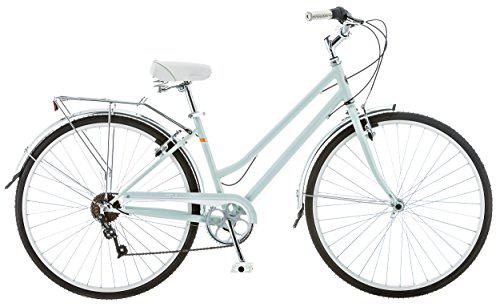} & \includegraphics[width=2cm]{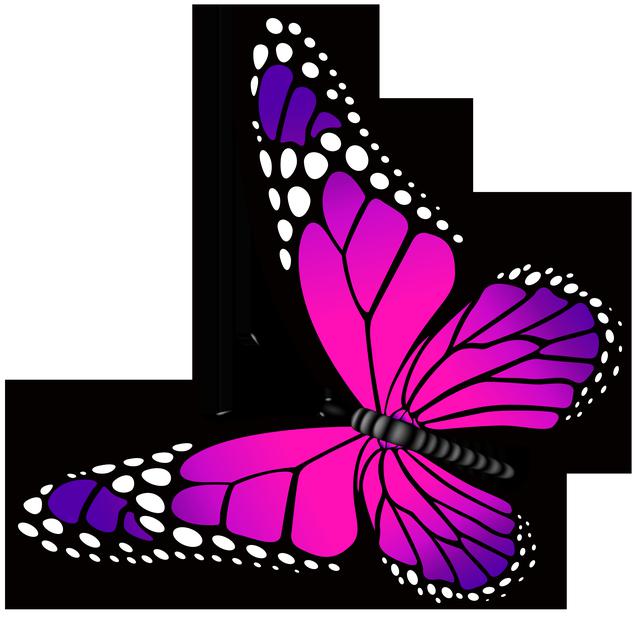}& \includegraphics[width=2cm]{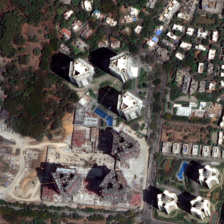}& \includegraphics[width=2cm]{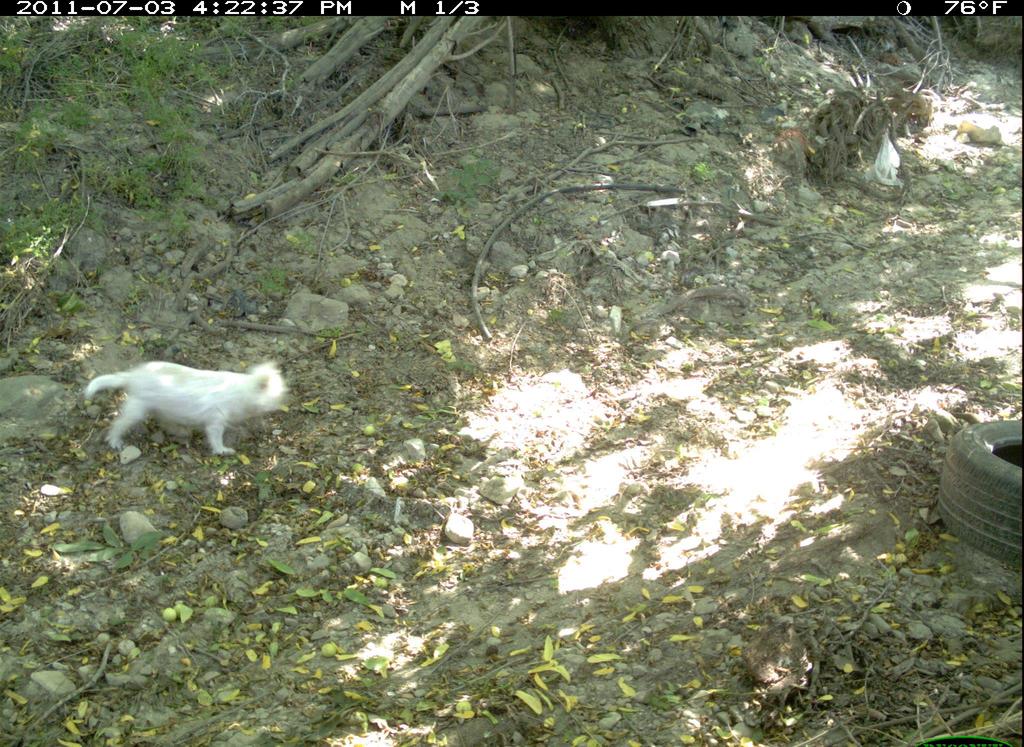} & \includegraphics[width=2cm]{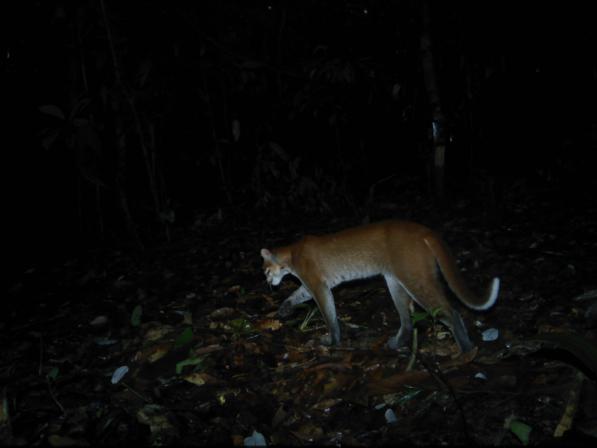}\\

\midrule
& \multicolumn{7}{>{\columncolor{LightBlue}}c}{\textit{random samples (180 cases)}} \\ \midrule
\multirow{2}{*}{CLIP} 
& \cellcolor{gray!20} 0.967 & \cellcolor{gray!20} 0.833 & \cellcolor{gray!20} 0.800 & \cellcolor{gray!20} 0.572 & \cellcolor{gray!20} 0.111 & \cellcolor{gray!20} 0.194 & \cellcolor{gray!20} 0.061 \\
 & \cellcolor{gray!5} 174/180 & \cellcolor{gray!5} 150/180 & \cellcolor{gray!5} 144/180 & \cellcolor{gray!5} 103/180 & \cellcolor{gray!5} 20/180 & \cellcolor{gray!5} 35/180 & \cellcolor{gray!5} 11/180\\
\midrule
\multirow{2}{*}{LLaVA}  & \cellcolor{gray!20}  0.994 & \cellcolor{gray!20} 0.894 & \cellcolor{gray!20} 0.650 & \cellcolor{gray!20} 0.306 & \cellcolor{gray!20} 0.128 & \cellcolor{gray!20} 0.539 & \cellcolor{gray!20} 0.006\\
  &  \cellcolor{gray!5} 179/180 & \cellcolor{gray!5} 161/180 & \cellcolor{gray!5} 117/180 & \cellcolor{gray!5} 55/180 & \cellcolor{gray!5} 23/180 & \cellcolor{gray!5} 97/180 & \cellcolor{gray!5} 1/180\\
  \midrule
\multirow{2}{*}{GPT-4V} & \cellcolor{gray!20} 0.978 &\cellcolor{gray!20}  0.797 &\cellcolor{gray!20}  0.936 & \cellcolor{gray!20} 0.833 & \cellcolor{gray!20} 0.220 & \cellcolor{gray!20} 0.500 & \cellcolor{gray!20} 0.309 \\
 & \cellcolor{gray!5} 175/179 & \cellcolor{gray!5} 141/177 & \cellcolor{gray!5} 160/171 & \cellcolor{gray!5} 135/162 &  \cellcolor{gray!5} 39/177 & \cellcolor{gray!5} 90/180 & \cellcolor{gray!5} 55/178\\
\midrule

\multirow{2}{*}{Gemini} & \cellcolor{gray!20} 0.983 &\cellcolor{gray!20}  0.871 &\cellcolor{gray!20}  0.963 & \cellcolor{gray!20} 0.910 & \cellcolor{gray!20} 0.333 & \cellcolor{gray!20} 0.483 & \cellcolor{gray!20} 0.396 \\
 & \cellcolor{gray!5} 173/176 & \cellcolor{gray!5} 148/170 & \cellcolor{gray!5} 155/161 & \cellcolor{gray!5} 142/156 &  \cellcolor{gray!5} 56/168 & \cellcolor{gray!5} 87/180 & \cellcolor{gray!5} 71/179\\
\midrule

& \multicolumn{7}{>{\columncolor{LightBlue}}c}{\textit{random samples (1800 cases)}} \\
\multirow{2}{*}{CLIP} 
& \cellcolor{gray!20} 0.961 & \cellcolor{gray!20} 0.808 & \cellcolor{gray!20} 0.778 & \cellcolor{gray!20} 0.582 & \cellcolor{gray!20} 0.161 & \cellcolor{gray!20} 0.214 & \cellcolor{gray!20} 0.064 \\
 & \cellcolor{gray!5} 1730/1800 & \cellcolor{gray!5} 1455/1800 & \cellcolor{gray!5} 1400/1800 & \cellcolor{gray!5} 1048/1800 & \cellcolor{gray!5} 290/1800 & \cellcolor{gray!5} 385/1800 & \cellcolor{gray!5} 116/1800\\
\midrule

\multirow{2}{*}{LLaVA}  & \cellcolor{gray!20}  0.982 & \cellcolor{gray!20} 0.852 & \cellcolor{gray!20} 0.703 & \cellcolor{gray!20} 0.370 & \cellcolor{gray!20} 0.147 & \cellcolor{gray!20} 0.488 & \cellcolor{gray!20} 0.014\\
  &  \cellcolor{gray!5} 1768/1800 & \cellcolor{gray!5} 1534/1800 & \cellcolor{gray!5} 1265/1800 & \cellcolor{gray!5} 666/1800 & \cellcolor{gray!5} 264/1800 & \cellcolor{gray!5} 879/1800 & \cellcolor{gray!5} 25/1800\\
  \midrule
\multirow{2}{*}{GPT-4V} & \cellcolor{gray!20} 0.969 &\cellcolor{gray!20}  0.888 &\cellcolor{gray!20}  0.889 & \cellcolor{gray!20} 0.680 & \cellcolor{gray!20} 0.238 & \cellcolor{gray!20} 0.459 & \cellcolor{gray!20} 0.265 \\
 & \cellcolor{gray!5} 1742/1797 & \cellcolor{gray!5} 1455/1799 & \cellcolor{gray!5} 1599/1800 & \cellcolor{gray!5} 1162/1710 &  \cellcolor{gray!5} 428/1800 & \cellcolor{gray!5} 827/1800 & \cellcolor{gray!5} 473/1787\\
\midrule

\multirow{2}{*}{Gemini} & \cellcolor{gray!20} 0.993 &\cellcolor{gray!20}  0.838 &\cellcolor{gray!20}  0.922 & \cellcolor{gray!20} 0.754 & \cellcolor{gray!20} 0.271 & \cellcolor{gray!20} 0.519 & \cellcolor{gray!20} 0.343 \\
 & \cellcolor{gray!5} 1770/1782 & \cellcolor{gray!5} 1445/1724 & \cellcolor{gray!5} 1528/1658 & \cellcolor{gray!5} 1214/1611 &  \cellcolor{gray!5} 473/1743 & \cellcolor{gray!5} 931/1794 & \cellcolor{gray!5} 600/1750\\
\midrule

& \multicolumn{7}{>{\columncolor{LightOrange}}c}{\textit{failure cases}} \\

\multirow{2}{*}{CLIP} & \cellcolor{gray!20} 0.000 & \cellcolor{gray!20} 0.000 &\cellcolor{gray!20} 0.000 &\cellcolor{gray!20} 0.000 &\cellcolor{gray!20} 0.000 &\cellcolor{gray!20} 0.000 &\cellcolor{gray!20} 0.000\\
     & \cellcolor{gray!5} 0/173 & \cellcolor{gray!5} 0/180 & \cellcolor{gray!5} 0/180 & \cellcolor{gray!5} 0/180 & \cellcolor{gray!5} 0/180 & \cellcolor{gray!5} 0/180 & \cellcolor{gray!5} 0/180\\
     \midrule
\multirow{2}{*}{LLaVA} & \cellcolor{gray!20} 0.751 & \cellcolor{gray!20} 0.517 &\cellcolor{gray!20} 0.406 &\cellcolor{gray!20} 0.128  & \cellcolor{gray!20} 0.083 & \cellcolor{gray!20} 0.517 &\cellcolor{gray!20} 0.016\\
       & \cellcolor{gray!5} 130/173 &\cellcolor{gray!5} 93/180 &\cellcolor{gray!5} 73/180 &\cellcolor{gray!5} 23/180 &\cellcolor{gray!5} 15/180 &\cellcolor{gray!5} 93/180 &\cellcolor{gray!5} 3/180\\
       \midrule
\multirow{2}{*}{GPT-4V} & \cellcolor{gray!20} 0.732& \cellcolor{gray!20} 0.651 & \cellcolor{gray!20} 0.774 &\cellcolor{gray!20} 0.523  &\cellcolor{gray!20} 0.192 &\cellcolor{gray!20} 0.411 &\cellcolor{gray!20} 0.285\\
       & \cellcolor{gray!5} 120/164 &\cellcolor{gray!5} 112/172 &\cellcolor{gray!5} 127/164 &\cellcolor{gray!5} 78/149  &\cellcolor{gray!5} 32/167 & \cellcolor{gray!5} 74/180 &\cellcolor{gray!5} 51/179\\
\midrule
\multirow{2}{*}{Gemini} & \cellcolor{gray!20} 0.848 &\cellcolor{gray!20}  0.650 &\cellcolor{gray!20}  0.860 & \cellcolor{gray!20} 0.736 & \cellcolor{gray!20} 0.266 & \cellcolor{gray!20} 0.458 & \cellcolor{gray!20} 0.431 \\
 & \cellcolor{gray!5} 140/165 & \cellcolor{gray!5} 104/160 & \cellcolor{gray!5} 141/164 & \cellcolor{gray!5} 106/144 &  \cellcolor{gray!5} 45/169 & \cellcolor{gray!5} 82/179 & \cellcolor{gray!5} 75/174\\

\bottomrule

\end{tabular}}
\end{table*}

\begin{table*}[t]
\centering
\scalebox{0.9}{
\begin{tabular}
{lcccccc} \toprule
 Method  & Office-Home  & PACS   & DomainNet    & TerraIncognita    & VLCS    & {Avg.}         \\ \midrule
MMD \citep{li2018mmd}                                                & 0.663 & 0.847 & 0.234 & 0.422          & 0.775 & 0.588           \\
Mixstyle \citep{zhou2021domain}                                     & 0.604 & 0.852 & 0.340 & 0.440          & 0.779 & 0.603           \\
GroupDRO \citep{sagawa2019distributionally}                          & 0.660 & 0.844 & 0.333 & 0.432          & 0.767 & 0.607           \\
IRM \citep{arjovsky2019invariant}                                    & 0.643 & 0.835 & 0.339 & 0.476          & 0.785 & 0.616           \\
CDANN \citep{li2018domain}                                           & 0.658 & 0.826 & 0.383 & 0.458          & 0.775 & 0.620           \\
DANN \citep{ganin2016domain}                                         & 0.659 & 0.836 & 0.383 & 0.467          & 0.786 & 0.626           \\
MTL \citep{blanchard2021domain}                                      & 0.664 & 0.846 & 0.406 & 0.456          & 0.772 & 0.629           \\
Mixup \citep{xu2020adversarial} & 0.681 & 0.846 & 0.392 & 0.479          & 0.774 & 0.634           \\
MLDG \citep{li2018learning}                                          & 0.668 & 0.849 & 0.412 & 0.477          & 0.772 & 0.636           \\
ERM \citep{vapnik1999overview}                                      & 0.676 & 0.842 & 0.440 & 0.478          & 0.773 & 0.642           \\
SagNet \citep{nam2021reducing}                                       & 0.681 & 0.863 & 0.403 & 0.486          & 0.778 & 0.642           \\
SelfReg \citep{kim2021selfreg}                                                 & 0.679 & 0.856 & 0.428 & 0.470          & 0.778 & 0.642           \\
CORAL \citep{sun2016deep}                                            & 0.687 & 0.862 & 0.415 & 0.476          & 0.788 & 0.645           \\
mDSDI \citep{bui2021exploiting}                                                & 0.692 & 0.862 & 0.428 & 0.481          & 0.790 & 0.651           \\
ERM + MIRO \citep{cha2022domain}                                                                        & 0.705 & 0.854 & 0.443 & 0.504          & 0.790 & 0.659           \\
ERM + SWAD \citep{cha2021swad}                                                         & 0.706 & 0.881 & 0.465 & 0.500          & 0.791 & 0.669           \\
CORAL + SWAD \citep{cha2021swad}                                                     & 0.713 & 0.883 & 0.468 & 0.510          & 0.789 & 0.673           \\
DIWA \citep{rame2022diverse} & 0.728 & 0.890 & 0.477 & 0.519 & 0.786 & 0.680  \\
ERM + MIRO + SWAD  \citep{cha2021swad}                                                                  & 0.724 & 0.884 & 0.470 & \textbf{0.529} & 0.796 & 0.681           \\
ERM++ \citep{teterwak2023erm++}   & 0.747  & 0.898   & 0.508   & 0.512   & 0.780    & 0.689 \\ 
\midrule

CLIP \citep{radford2021learning} & 0.778 & 0.961 & 0.582 & 0.214 & 0.808 & 0.669 \\
LLaVA \citep{liu2023visual, liu2023improved} & 0.703 & 0.982 & 0.370 & 0.488 & 0.852 & 0.679 \\
GPT-4V \citep{gpt4v} & 0.889 & 0.969 & 0.680 & 0.459 & \textbf{0.888} & 0.777 \\
Gemini \citep{team2023gemini} & \textbf{0.922} & \textbf{0.993} & \textbf{0.754} & 0.519 & 0.838 & \textbf{0.805}\\
\bottomrule
\end{tabular}}
\caption{\textbf{Zero-shot Generalization Performance of GPT-4V on DomainBed:} In the DomainBed benchmark for domain generalization, GPT-4V demonstrates superior zero-shot generalization capabilities, surpassing traditional approaches and marking a significant advancement in the field. The results highlight GPT-4V's effectiveness across diverse domains, showcasing its potential for robust and versatile applications.} 
\label{table:all_methods}
\end{table*}

\subsubsection{Task Introduction}
The category of natural visuals encompasses an extensive array of real-world imagery, capturing the myriad facets of nature and everyday life. 
This domain is characterized by its inherent diversity and complexity, presenting scenes and objects that are commonly encountered in daily experiences.

In our study, we examine the following natural datasets, each with its distinct characteristics and challenges:

\begin{itemize}
    \item \textbf{PACS~\citep{li2017deeper}:} Comprising images from four different styles - art painting, cartoon, photo, and sketch - this dataset challenges models to generalize across artistic mediums, testing their ability to recognize the same objects in vastly different visual representations.

    \item \textbf{VLCS~\citep{fang2013unbiased}:} This dataset is a collection from four different image repositories. It poses a challenge in terms of variations in image quality, lighting, and backgrounds, requiring robust feature extraction for successful classification.

    \item \textbf{Office-Home~\citep{venkateswara2017deep}:} Featuring everyday objects from office and home environments, this dataset includes images from diverse categories such as Art, Clipart, Product, and Real World, offering a testbed for models to generalize across everyday items.

    \item \textbf{DomainNet~\citep{peng2019moment}:} Encompassing a variety of artistic styles and objects, DomainNet is a large-scale dataset that tests a model’s ability to generalize across different visual domains and a vast array of object classes.

    \item \textbf{Fmow~\citep{christie2018functional}:} This dataset focuses on land use and land cover classification, presenting a challenge with its time-series satellite imagery, which includes temporal and regional variations.

    \item \textbf{TerraIncognita~\citep{beery2018recognition}:} Composed of wildlife images captured by camera traps in various locations, it tests models' abilities to recognize animal species across different environmental conditions and camera settings.

    \item \textbf{iWildCam~\citep{beery2021iwildcam}:} The iWildCam dataset offers a unique challenge in the realm of wildlife conservation and ecological studies. Comprised of images captured by camera traps set up in diverse wilderness locations, it is tailored to evaluate the ability of models to identify and classify a wide range of animal species. 
\end{itemize}

These datasets not only cover a wide range of natural scenes and objects but also introduce various types of distribution shifts, making them ideal for evaluating the zero-shot generalization capabilities of GPT-4V, in comparison with CLIP, LLaVA, and Gemini. 
Each dataset presents its unique set of challenges, from artistic style variations in PACS to environmental differences in TerraIncognita, providing a comprehensive testbed for assessing model robustness in natural settings.
Table~\ref{617085393024} firstly provides an overview of each natural dataset, detailing key aspects such as the type of prediction, domain characteristics, the number of domains and classes, and illustrative examples. 
This table serves as a quick reference to understand the diversity and scope of challenges posed by these datasets in our evaluation.

\subsubsection{Comparative Accuracies Across Datasets and Domains}
Table~\ref{617085393024} outlines the accuracies and correct-to-total case ratios for four models (CLIP, LLaVA, Gemini, and GPT-4V) across seven natural datasets, incorporating both random samples and failure cases identified in CLIP.
This subsection is dedicated to examining GPT-4V's zero-shot generalization abilities within natural datasets.

\textbf{GPT-4V's Performance in Random Samples:}
Focusing first on datasets with a large variety of domains and classes, such as Office-Home and DomainNet, GPT-4V demonstrates a notable capacity for generalization. Its high accuracy rates in Office-Home (0.889) and DomainNet (0.680) suggest a robust understanding and adaptability to diverse natural visuals, including a broad range of everyday items and varied artistic styles.
Additionally, in uncommon datasets like Fmow and TerraIncognita, GPT-4V significantly surpasses CLIP's performance (0.238 vs 0.161 in Fmow and 0.459 vs 0.214 in TerraIncognita).
In the PACS and VLCS datasets, all three models perform well, with accuracies exceeding 0.8.
This consistency suggests that these domains may have been included in the pre-training data of these three models.  

\textbf{GPT-4V in Handling CLIP's Failure Cases:}
To assess GPT-4V's capabilities in more challenging scenarios, we examine its performance on CLIP's failure cases. In datasets with a diverse range of classes, such as DomainNet and Office-Home, GPT-4V shows remarkable resilience. For instance, in Office-Home, GPT-4V achieves an accuracy of 0.774, surpassing LLaVA's 0.406. 
Similarly, in DomainNet, GPT-4V records 0.523 accuracy, significantly higher than LLaVA's 0.128. This trend is also evident in Fmow, where GPT-4V's performance (0.192) markedly exceeds LLaVA's (0.083). These results indicate GPT-4V's robustness in handling complex and challenging visuals, even in scenarios where CLIP struggled.

\begin{figure*}[htbp]
  \centering

  \begin{subfigure}[t]{\textwidth}
    \centering
    \includegraphics[width=0.8\textwidth]{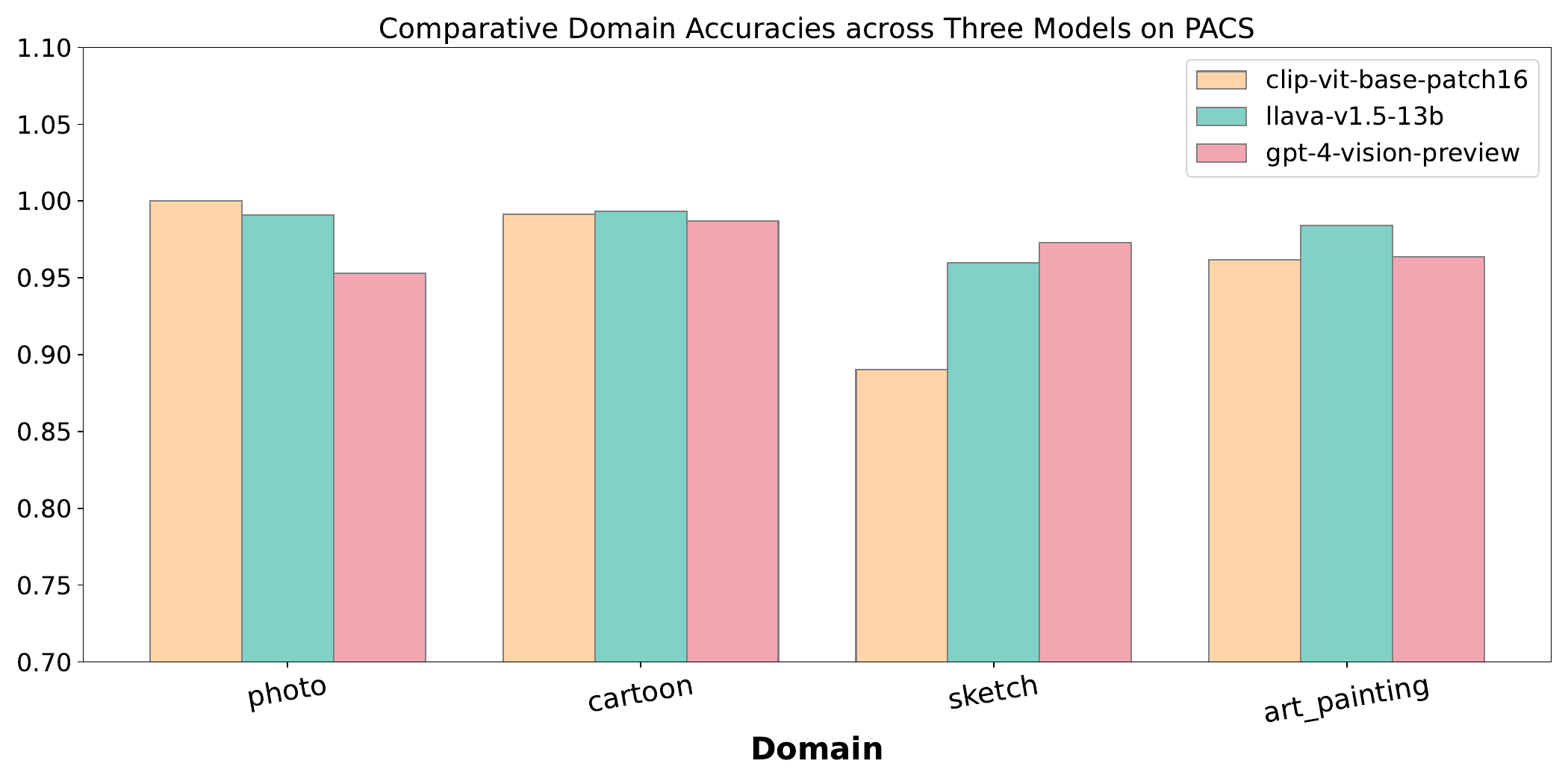}
    \caption{Comparative Domain Accuracies across Three Models on PACS}
    \label{fig:sub1}
  \end{subfigure}


  \begin{subfigure}[t]{\textwidth}
    \centering
    \includegraphics[width=0.8\textwidth]{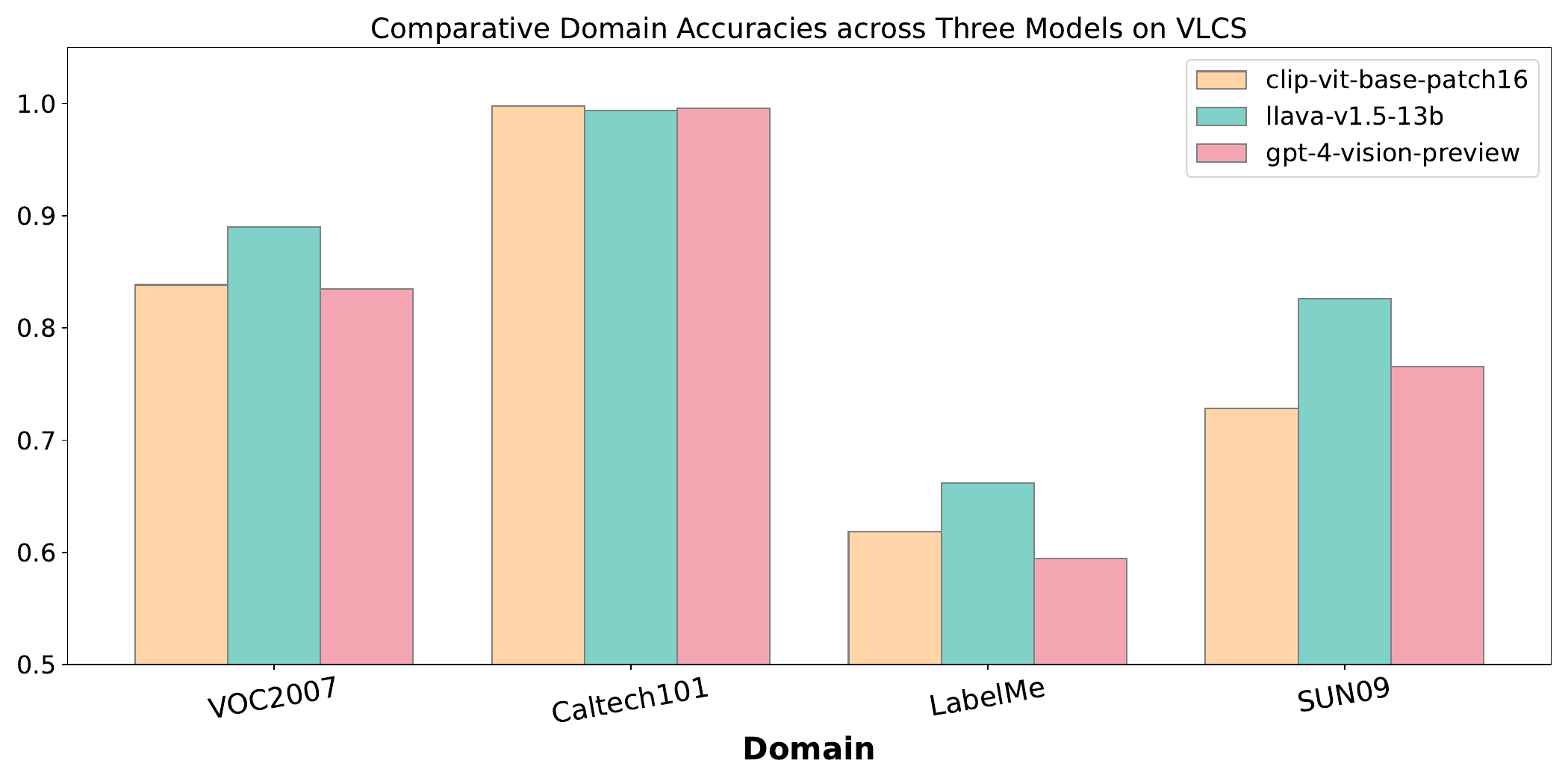}
    \caption{Comparative Domain Accuracies across Three Models on VLCS}
    \label{fig:sub2}
  \end{subfigure}


  \begin{subfigure}[t]{\textwidth}
    \centering
    \includegraphics[width=0.8\textwidth]{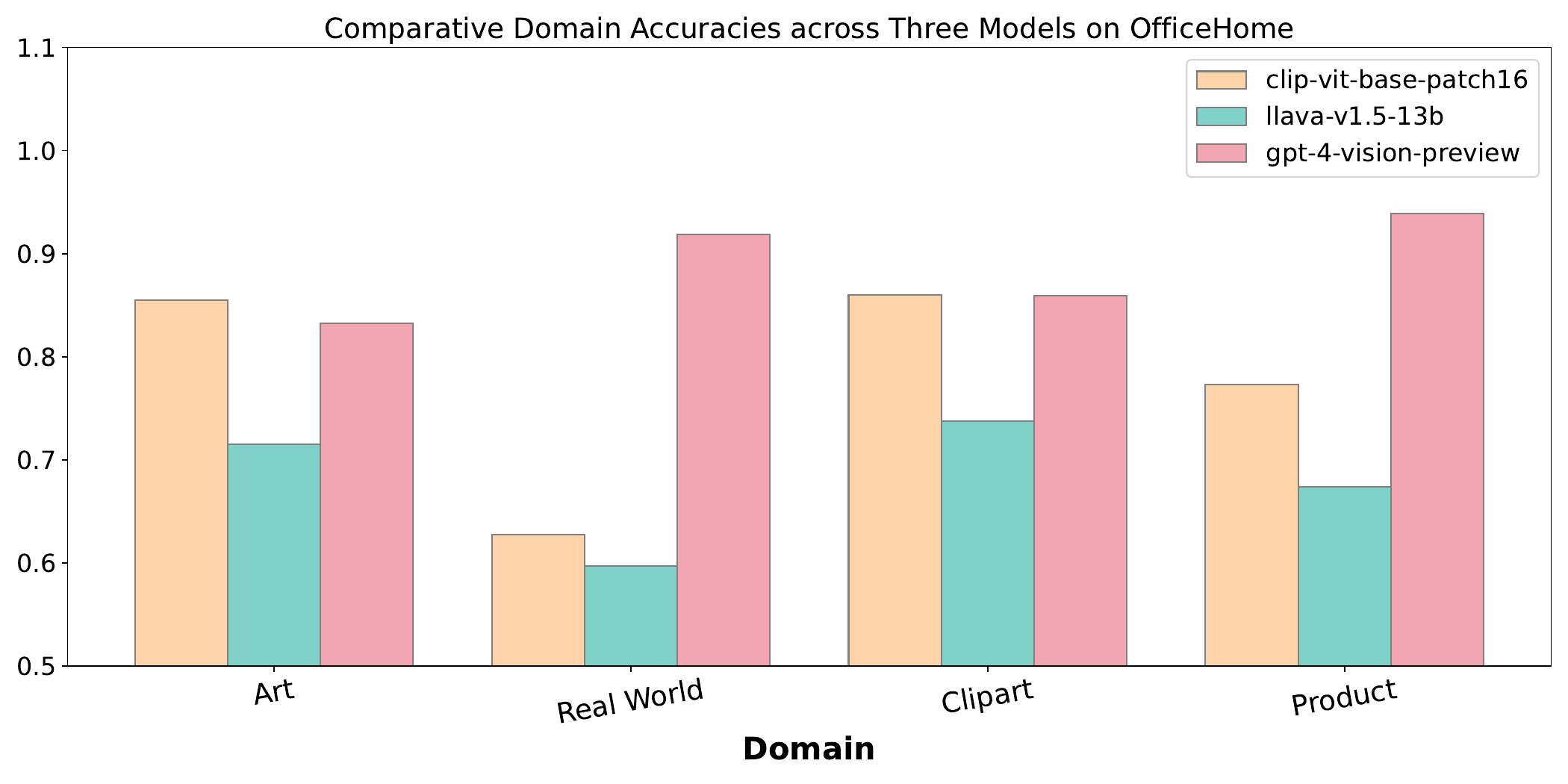}
    \caption{Comparative Domain Accuracies across Three Models on OfficeHome}
    \label{fig:sub3}
  \end{subfigure}

  \caption[Comparative Domain Accuracies on PACS, VLCS and Office-Home]{Comparative accuracies of three models across domains in the PACS, VLCS, Office-Home datasets.}
  \label{767892400375}
\end{figure*}

\begin{figure*}[htbp]
  \centering

  \begin{subfigure}[t]{\textwidth}
    \centering
    \includegraphics[width=0.8\textwidth]{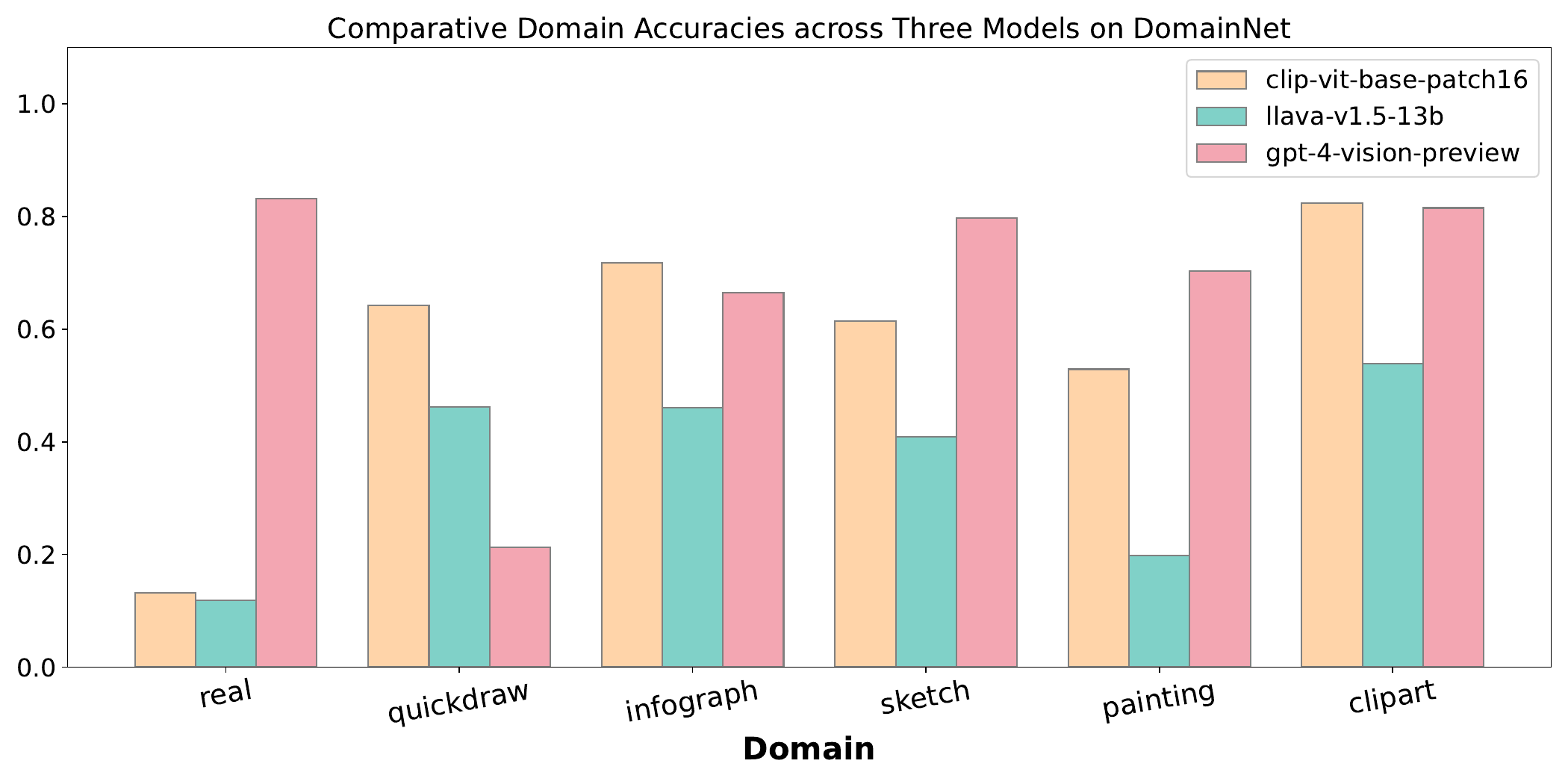}
    \caption{Comparative Domain Accuracies across Three Models on DomainNet}
  \end{subfigure}

  \begin{subfigure}[t]{\textwidth}
    \centering
    \includegraphics[width=0.8\textwidth]{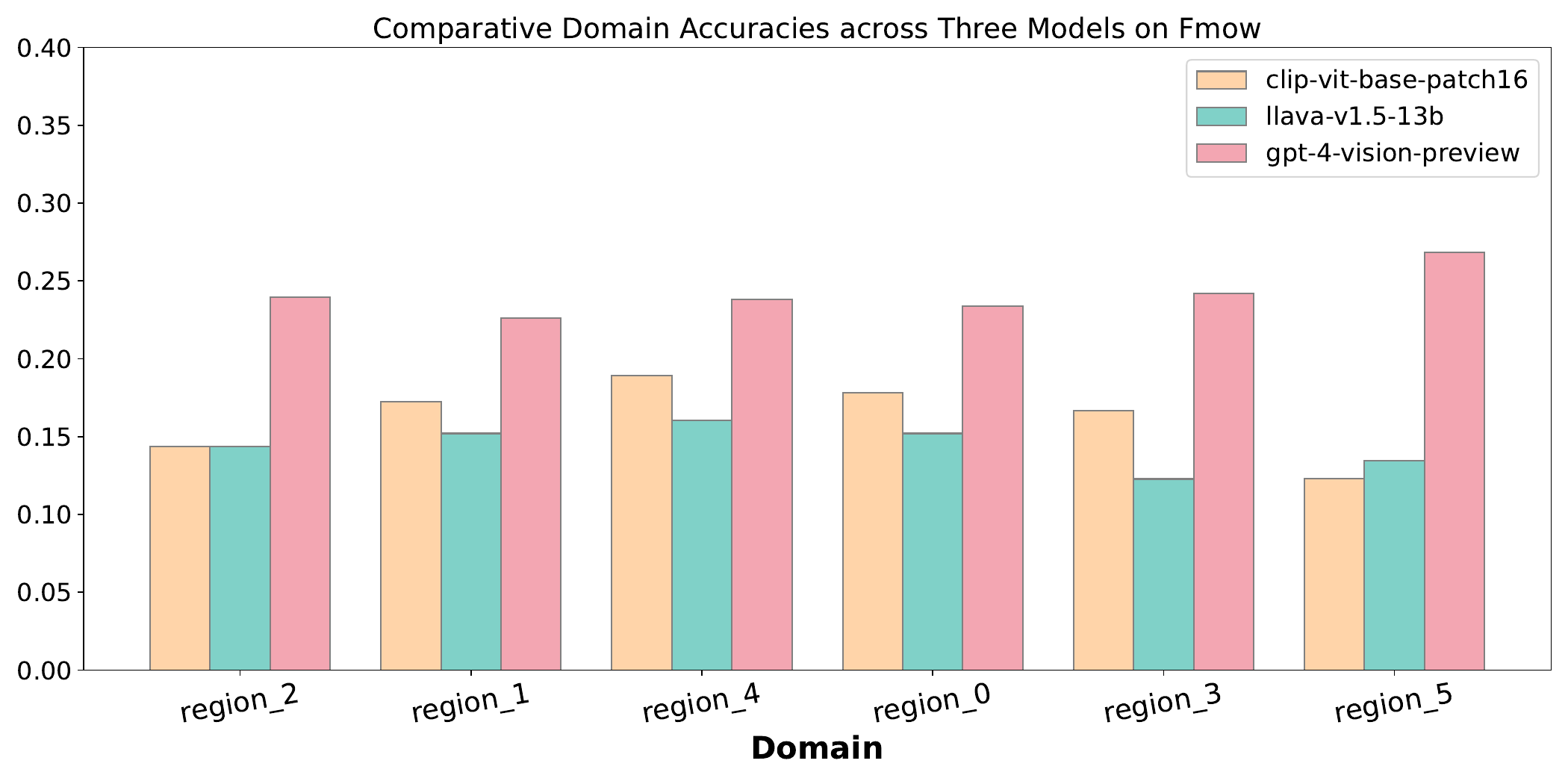}
    \caption{Comparative Domain Accuracies across Three Models on Fmow}
  \end{subfigure}


  \begin{subfigure}[t]{\textwidth}
    \centering
    \includegraphics[width=0.8\textwidth]{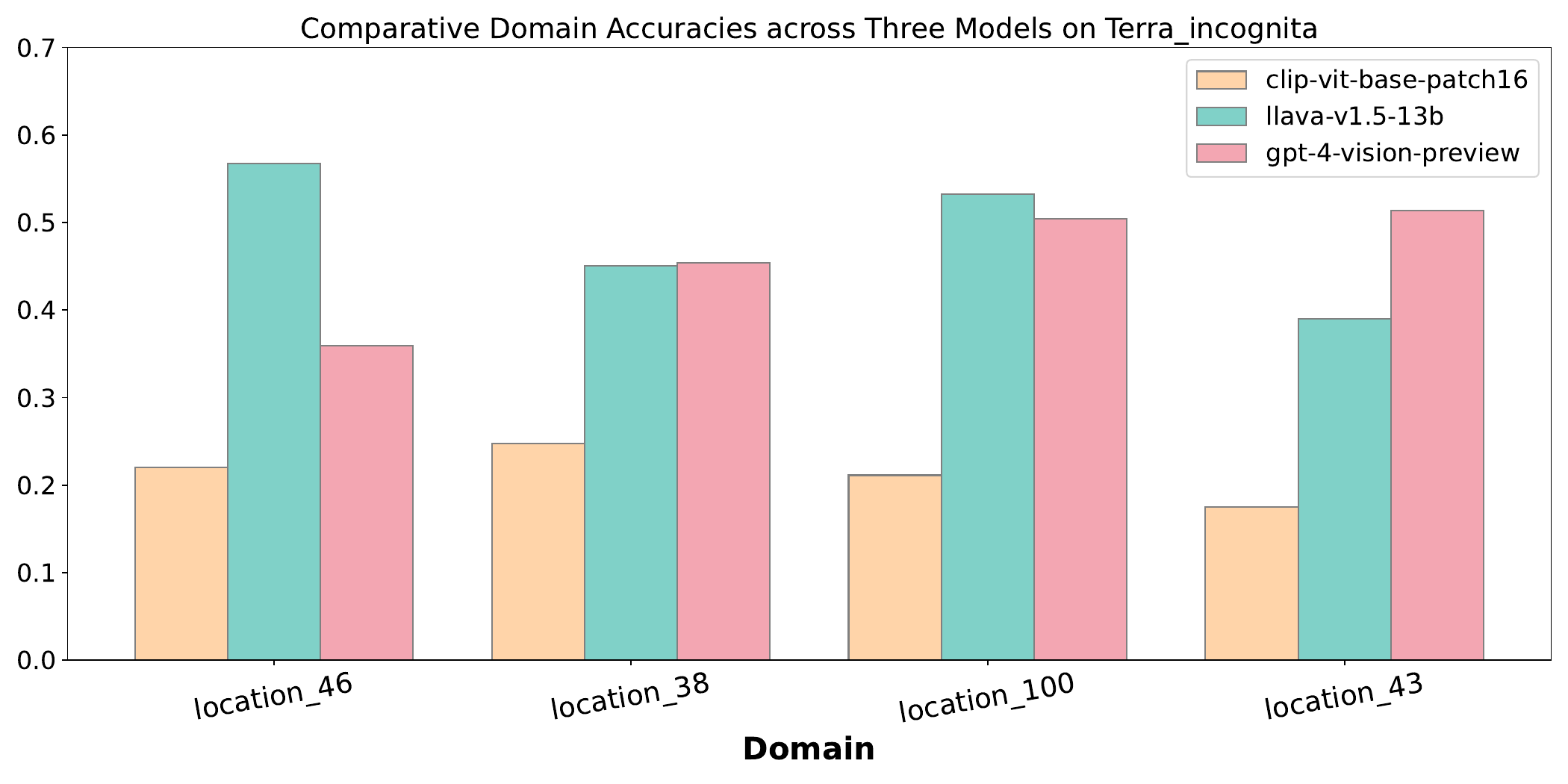}
    \caption{Comparative Domain Accuracies across Three Models on TerraIncognita}
  \end{subfigure}

  \caption[Comparative Domain Accuracies on DomainNet, Fmow and TerraIncognita]{Comparative accuracies of three models across domains in the DomainNet, Fmow, TerraIncognita datasets.}
  \label{636022932930}
\end{figure*}

\textbf{GPT-4V's Performance Across Individual Domains: }
While Table~\ref{617085393024} provides an overall view of the accuracies for the three models across various datasets, a more granular look at their performance in specific domains is essential for a comprehensive understanding.
To this end, we have detailed comparative domain accuracies for each model within the PACS, VLCS, Office-Home, DomainNet, Fmow, and TerraIncognita datasets.
These comparisons are illustrated in Figures~\ref{767892400375}, \ref{636022932930}.
These figures illuminate the relative strengths and weaknesses of each model across different domains within the datasets and help to understand the extent of GPT-4V's generalization capabilities and how it compares to CLIP and LLaVA in diverse contexts.

\textbf{Highlighting GPT-4V's Superiority in DomainBed:} In the context of DomainBed~\citep{gulrajani2020search}, the popular benchmark for domain generalization, Table~\ref{table:all_methods} provides a clear illustration of the strides made by GPT-4V. It achieves unparalleled zero-shot generalization performance, significantly outpacing traditional domain generalization methods. Its exceptional performance across the board is indicative of its sophisticated understanding and the ability to adapt to new, unseen domains. This achievement is not just a reflection of GPT-4V's powerful architecture but also an indicator of its potential to revolutionize how models tackle the challenge of domain generalization.

\subsubsection{Case Demonstration}
\begin{figure*}[htb!]
\centering
\includegraphics[width=0.96\textwidth]{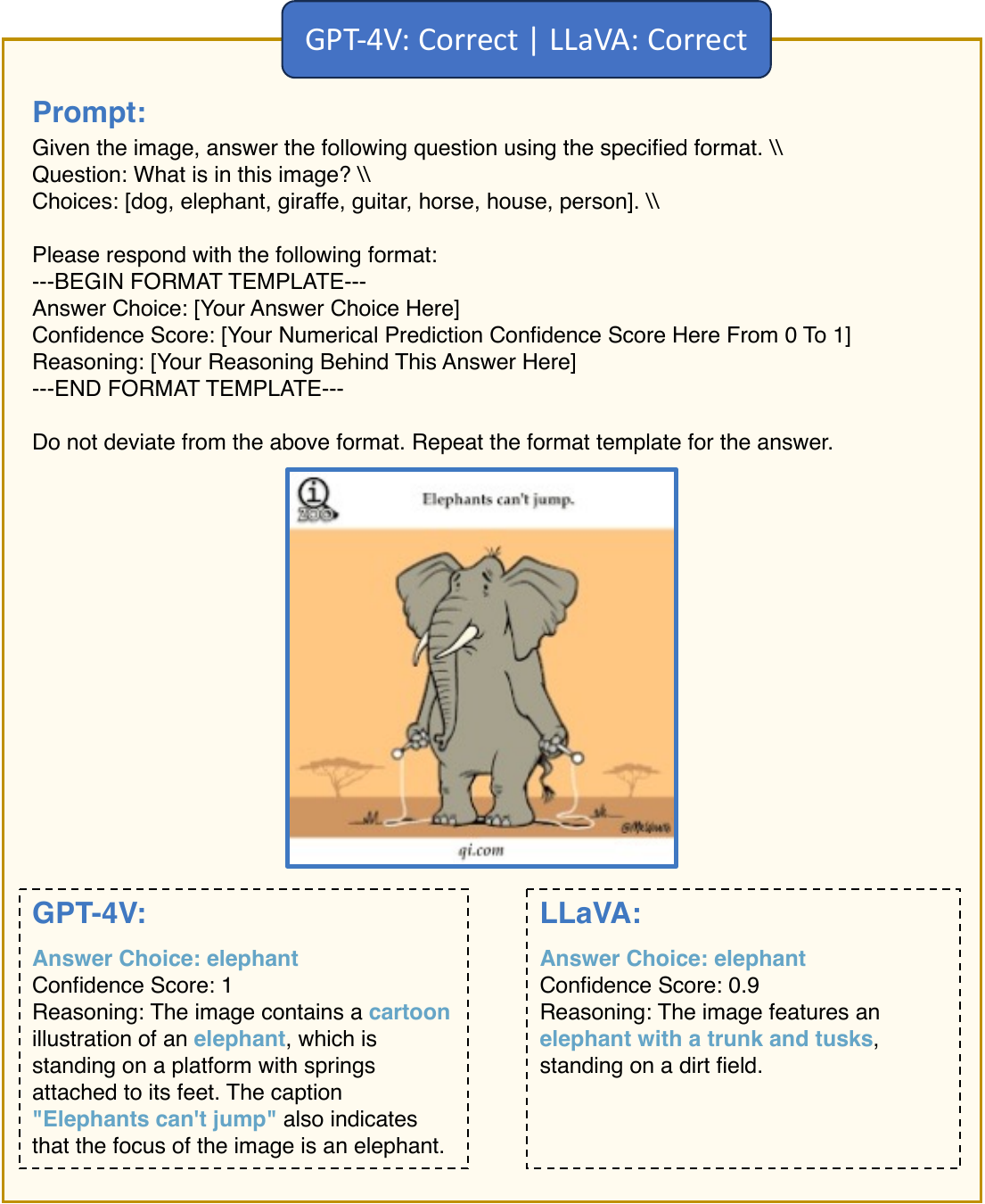}
\caption[Natural Distribution Shift: Case 1 on PACS]{Natural Distribution Shift: Case 1 - Elephant category in the Cartoon Domain of PACS Dataset.
In this instance, both GPT-4V and LLaVA are provided with the same text prompt alongside an image.
Both models successfully generate an answer choice, a confidence score, and their reasoning.
Notably, GPT-4V demonstrates a capability for detail recognition, accurately identifying the text 'Elephants can't jump.' in the image.
This case exemplifies GPT-4V's advanced ability to discern and interpret finer details within visual inputs, compared to LLaVA.
}
\label{775225593585}
\end{figure*}

\begin{figure*}[htb!]
\centering
\includegraphics[width=0.96\textwidth]{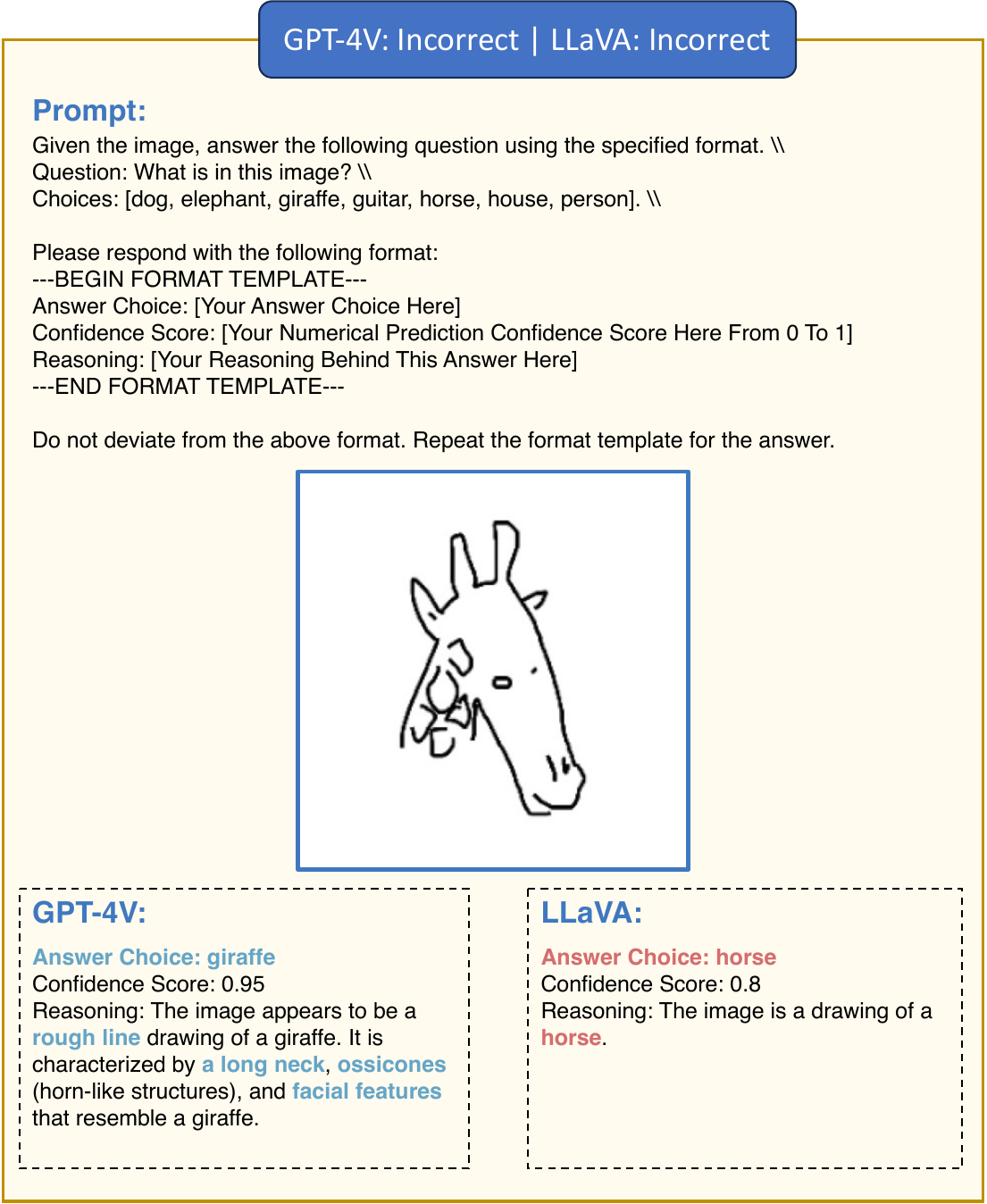}
\caption[Natural Distribution Shift: Case 2 on PACS]{Natural Distribution Shift: Case 2 - Giraffe category in the Sketch Domain of PACS Dataset.
This example showcases the performance of GPT-4V and LLaVA when given the same text prompt with an image depicting a giraffe in a sketch style. 
GPT-4V successfully identifies the giraffe, providing detailed reasoning and demonstrating a nuanced understanding of the image's content, such as long neck, horn-like structures. 
In contrast, LLaVA fails to correctly identify the giraffe, offering limited reasoning in its response.
}
\label{896371250907}
\end{figure*}

\begin{figure*}[htb!]
\centering
\includegraphics[width=0.96\textwidth]{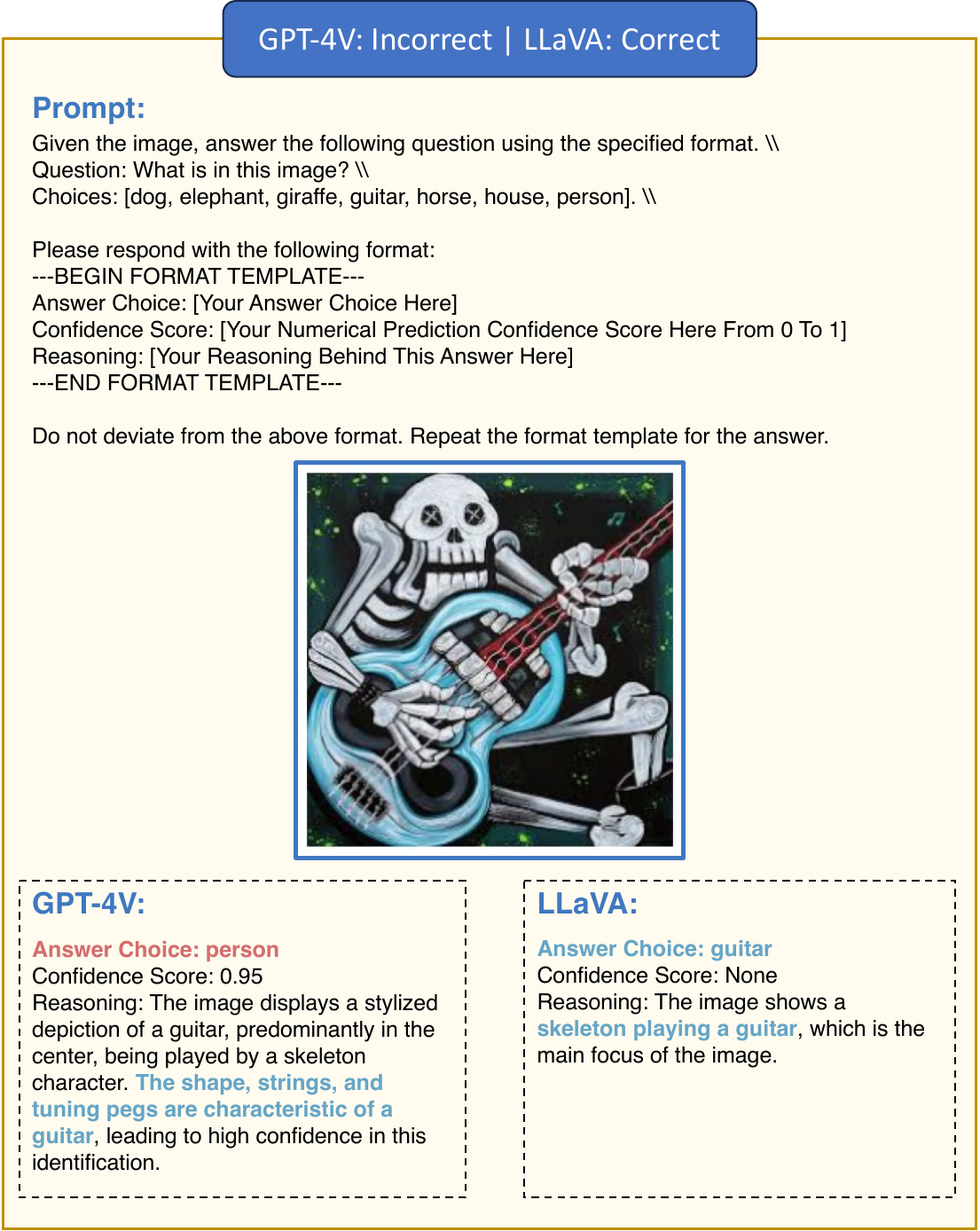}
\caption[Natural Distribution Shift: Case 3 on PACS]{Natural Distribution Shift: Case 3 - Guitar category in the Art\_painting Domain of PACS Dataset.
While LLaVA accurately classifies the image, GPT-4V fails to identify the correct class. However, an interesting observation emerges in the rationale provided by GPT-4V. Despite the incorrect classification, GPT-4V articulates a highly reasoned and contextually relevant explanation, offering a detailed and accurate description of the ground\_truth class label: guitar.
}
\label{289347172711}
\end{figure*}

\begin{figure*}[htb!]
\centering
\includegraphics[width=0.96\textwidth]{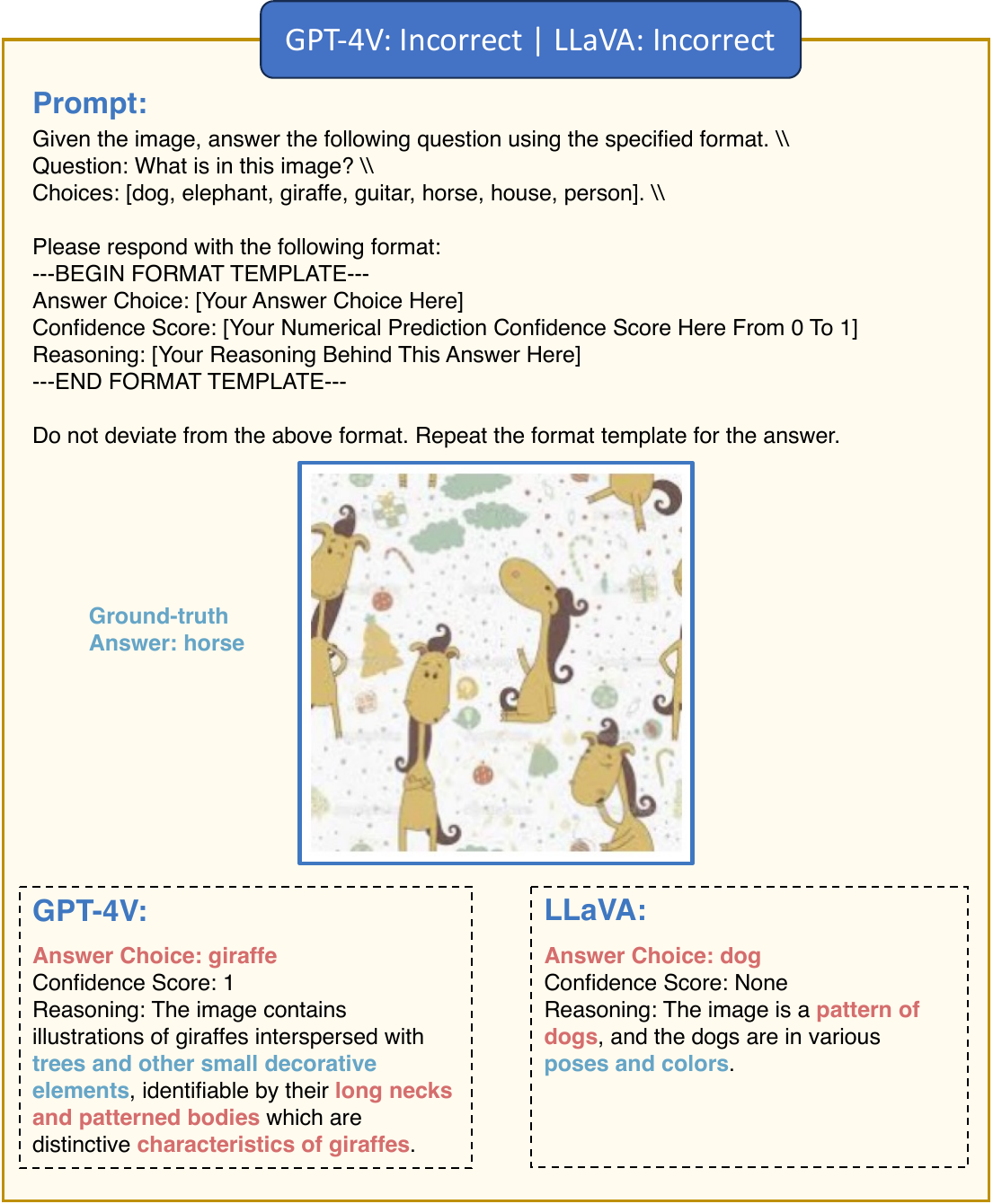}
\caption[Natural Distribution Shift: Case 4 on PACS]{Natural Distribution Shift: Case 4 - Horse category in the Cartoon Domain of PACS Dataset.
In this example, both GPT-4V and LLaVA incorrectly identify the subject in the image. The cartoon domain often features abstract styles where certain aspects of objects are exaggerated, as seen in the elongated necks of the horses in the image. 
GPT-4V incorrectly classifies the subject as a giraffe, likely influenced by the exaggerated neck feature. 
Compared to LLaVA, which provides limited reasoning, GPT-4V's rationale, though leading to an incorrect conclusion, is more detailed, noting the distinctive long neck as a key characteristic for its prediction. 
}
\label{830473335836}
\end{figure*}

\begin{figure*}[htb!]
\centering
\includegraphics[width=0.96\textwidth]{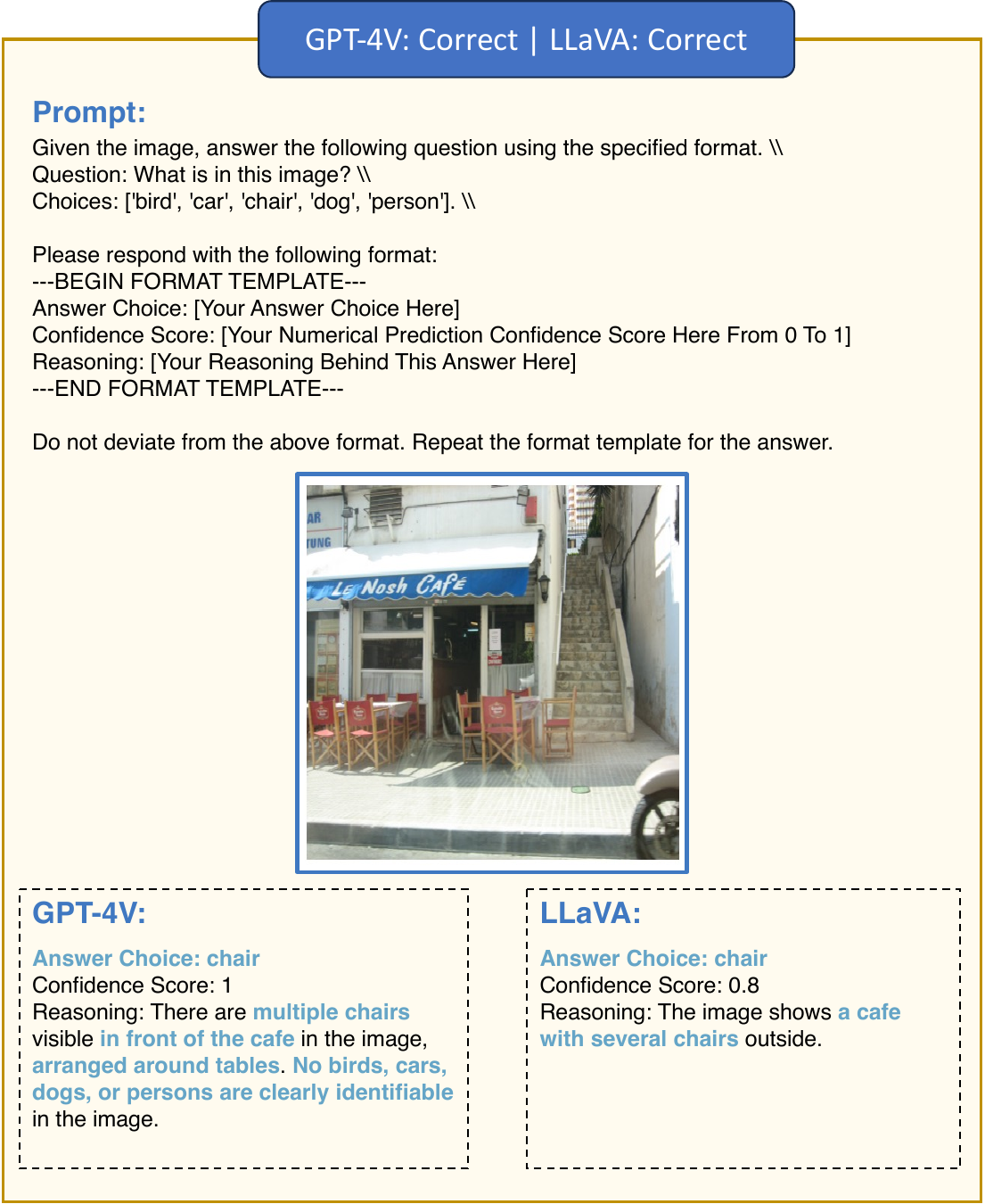}
\caption[Natural Distribution Shift: Case 5 on VLCS]{Natural Distribution Shift: Case 5 - Chair category in the LabelMe Domain of VLCS Dataset.
This case illustrates the proficiency of both GPT-4V and LLaVA models in accurately identifying multiple chairs within the scene. GPT-4V, in particular, stands out for its detailed and comprehensive description, offering nuanced insights that surpass the more straightforward analysis provided by LLaVA.
}
\label{015926979441}
\end{figure*}

\begin{figure*}[htb!]
\centering
\includegraphics[width=0.96\textwidth]{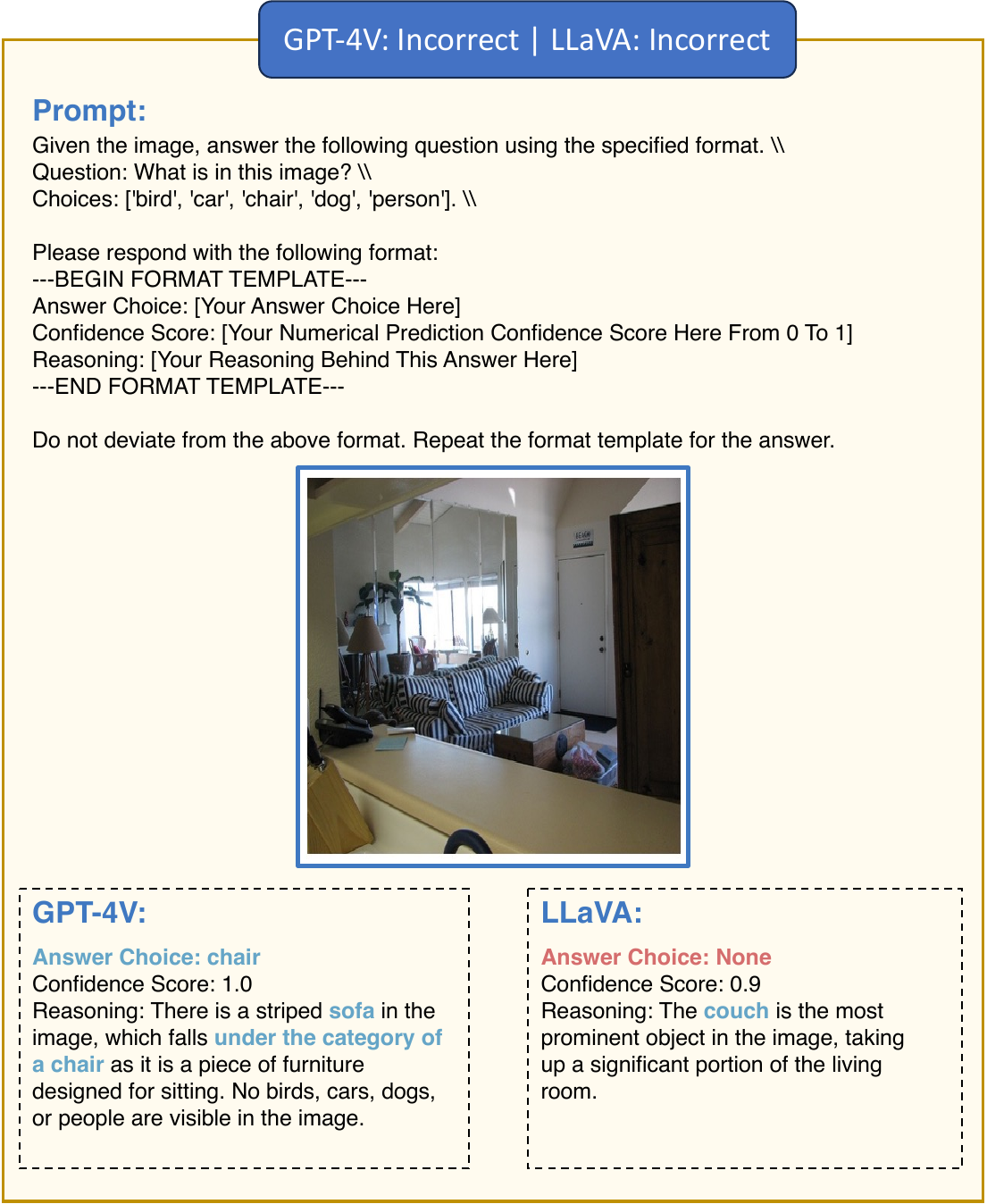}
\caption[Natural Distribution Shift: Case 6 on VLCS]{Natural Distribution Shift: Case 6 - Chair category in the LabelMe Domain of VLCS Dataset.
In this scenario, both GPT-4V and LLaVA models are presented with an image of a sofa/couch. 
GPT-4V demonstrates adaptability by categorizing the sofa as a type of chair, aligning with the limitations of the provided answer choices, and thus delivering an accurate classification. 
In contrast, LLaVA struggles to make the correct inference within the given constraints, highlighting a notable difference in their interpretative flexibility.
}
\label{546069757003}
\end{figure*}

\begin{figure*}[htb!]
\centering
\includegraphics[width=0.96\textwidth]{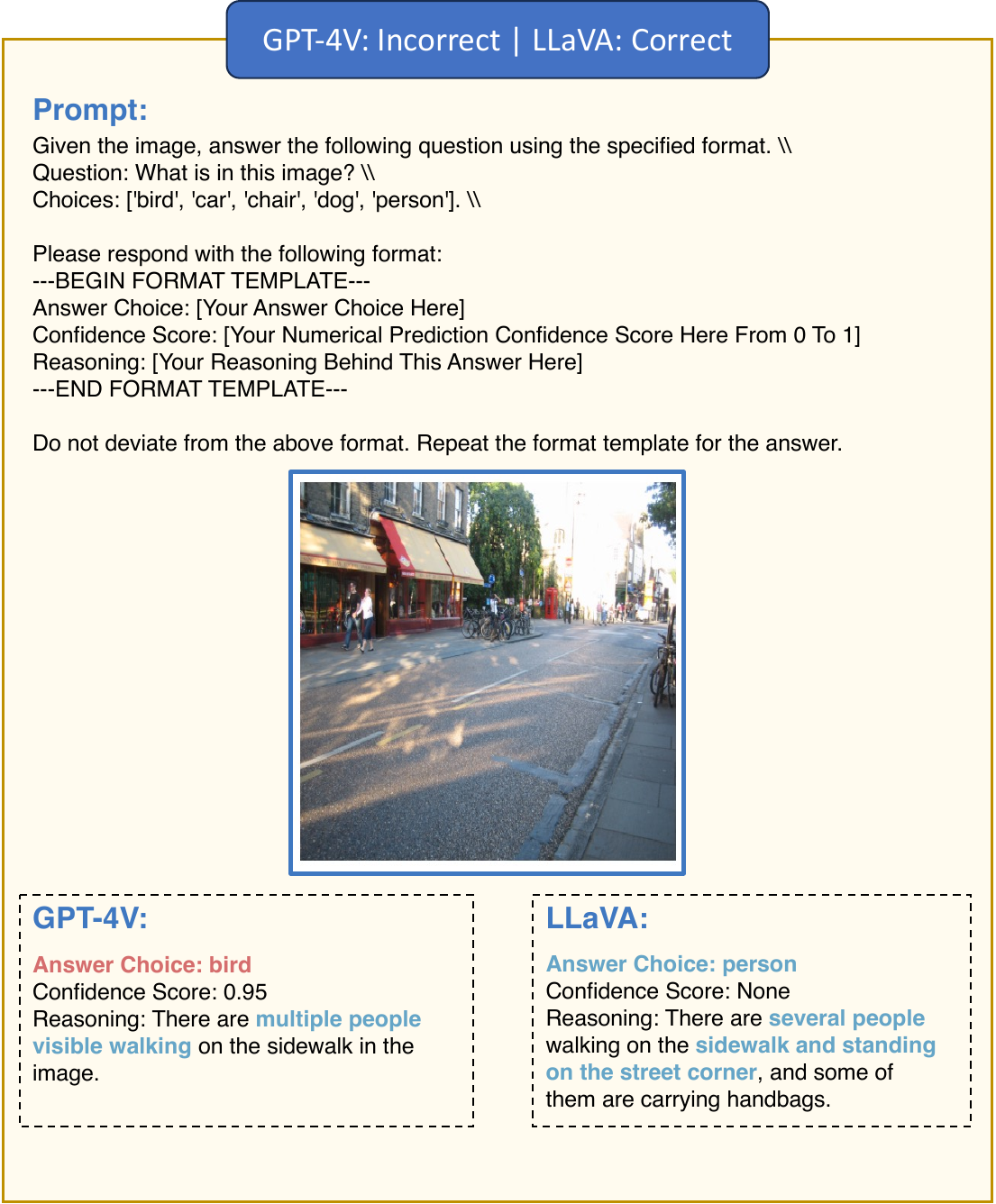}
\caption[Natural Distribution Shift: Case 7 on VLCS]{Natural Distribution Shift: Case 7 - Person category in the LabelMe Domain of VLCS Dataset.
In this instance, despite GPT-4V providing a logically sound reasoning process, it paradoxically arrives at an incorrect conclusion. This case highlights an intriguing aspect of GPT-4V's performance, where accurate analysis and reasoning do not always lead to the correct classification.
}
\label{340345078522}
\end{figure*}

\begin{figure*}[htb!]
\centering
\includegraphics[width=0.96\textwidth]{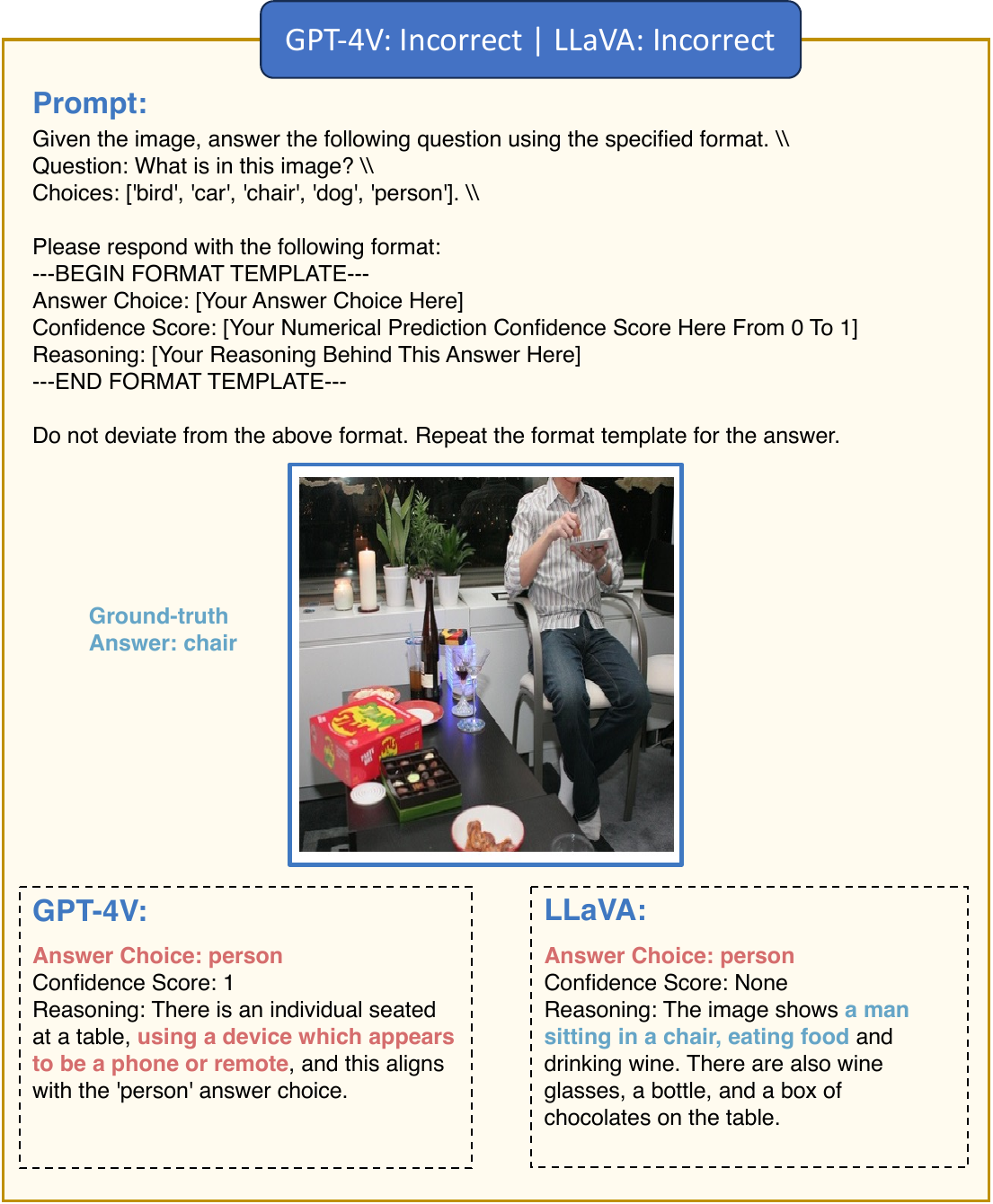}
\caption[Natural Distribution Shift: Case 8 on VLCS]{Natural Distribution Shift: Case 8 - Chair category in the VOC2007 Domain of VLCS Dataset.
This scenario illustrates the challenge faced by models like GPT-4V and LLaVA in accurately classifying images with multiple objects. 
Despite providing rational explanations, these models struggle to pinpoint the correct class when presented with complex scenes containing various elements. 

}
\label{728390971249}
\end{figure*}

\begin{figure*}[htb!]
\centering
\includegraphics[width=0.96\textwidth]{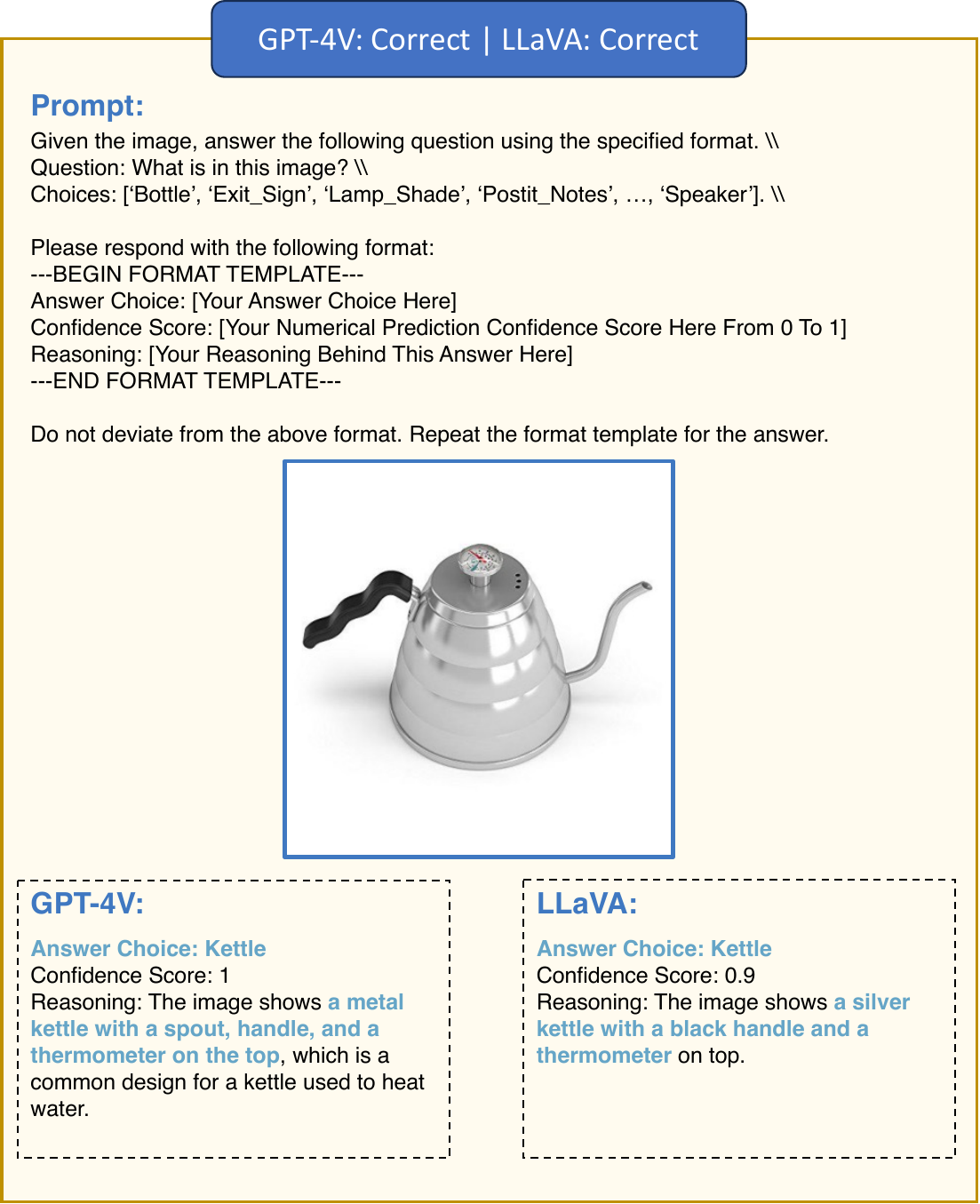}
\caption[Natural Distribution Shift: Case 9 on Office-Home]{Natural Distribution Shift: Case 9 - Kettle category in the Product Domain of Office-Home Dataset.
In this case study, both GPT-4V and LLaVA models are tasked with responding to an identical text prompt accompanied by an image.
It is noteworthy that GPT-4V demonstrates a more nuanced understanding, particularly in its ability to detail specific features such as the kettle's metallic nature and the presence of a spout. 
Additionally, GPT-4V enhances its answer with a summary that emphasizes typical design characteristics, thereby lending greater confidence to its response.
}
\label{169039403841}
\end{figure*}

\begin{figure*}[htb!]
\centering
\includegraphics[width=0.96\textwidth]{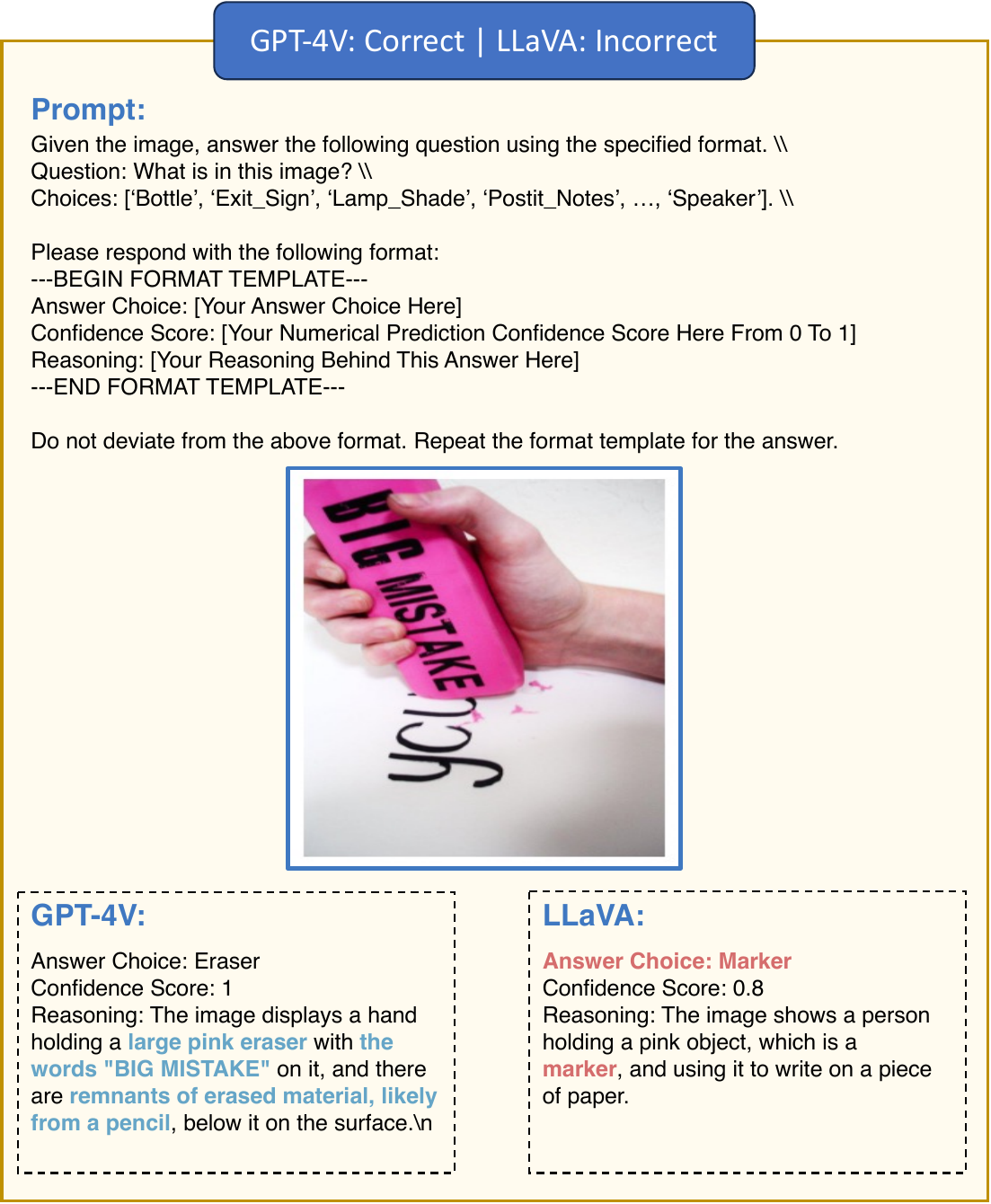}
\caption[Natural Distribution Shift: Case 10 on Office-Home]{Natural Distribution Shift: Case 10 - Analyzing the 'Eraser' Category in the Art Domain of the Office-Home Dataset.
This figure presents an intriguing instance where the depicted 'Eraser' might be initially mistaken for a 'Marker', a common perceptual challenge. GPT-4V remarkably identifies the correct object, utilizing cues from the text in the image, as well as the object's size and color. Notably, GPT-4V correctly interprets the action of erasing, in contrast to LLaVA, which interprets the action as writing. This demonstrates GPT-4V's advanced reasoning capabilities in distinguishing subtle contextual differences.
}
\label{293135981812}
\end{figure*}

\begin{figure*}[htb!]
\centering
\includegraphics[width=0.96\textwidth]{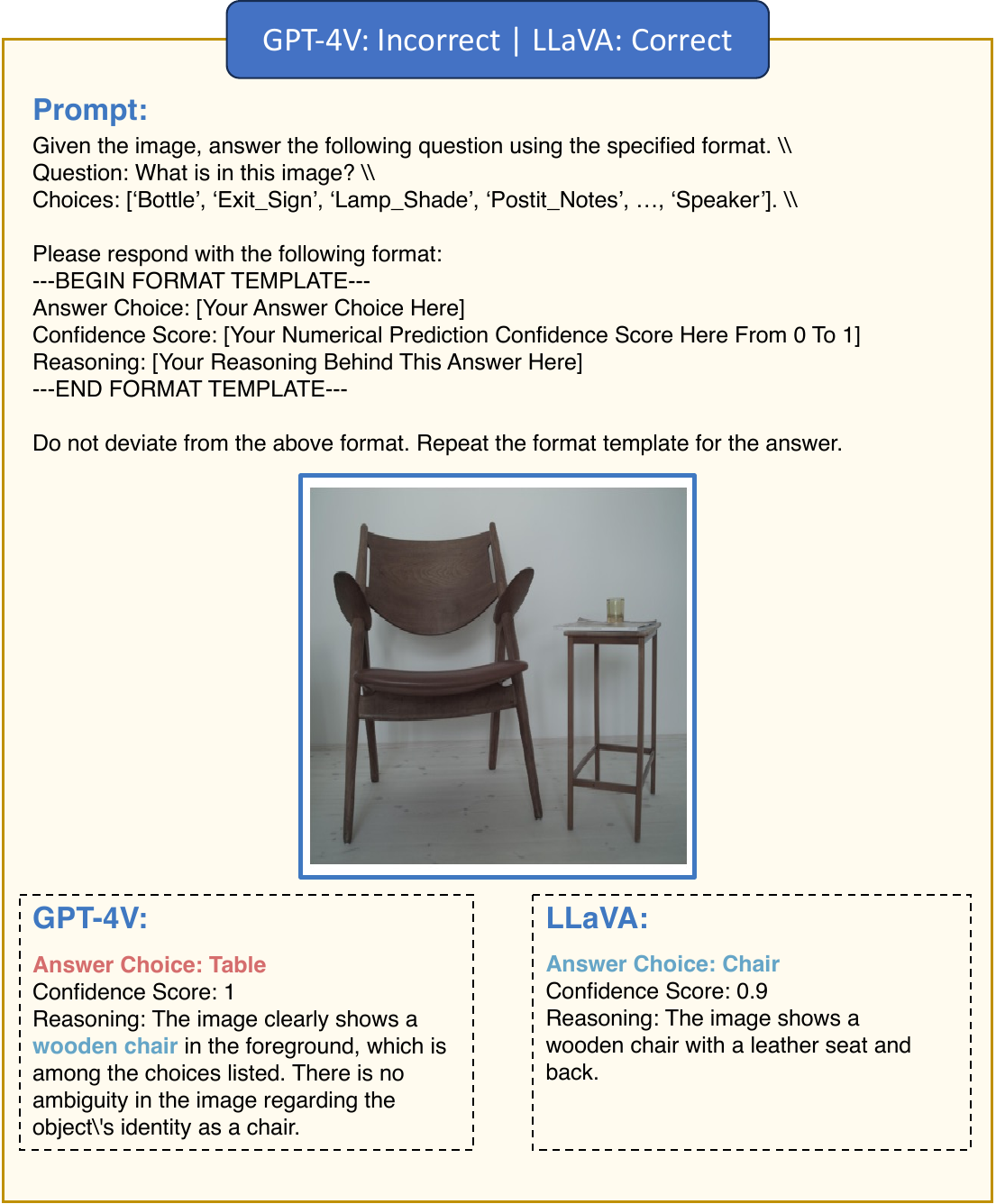}
\caption[Natural Distribution Shift: Case 11 on Office-Home]{Natural Distribution Shift: Case 11 - Chair category in the Real World Domain of Office-Home Dataset.
In this example, GPT-4V exhibits details and accuracy in its description of the image. 
Despite this, the model ultimately arrives at an incorrect classification.
}
\label{044345318906}
\end{figure*}

\begin{figure*}[htb!]
\centering
\includegraphics[width=0.96\textwidth]{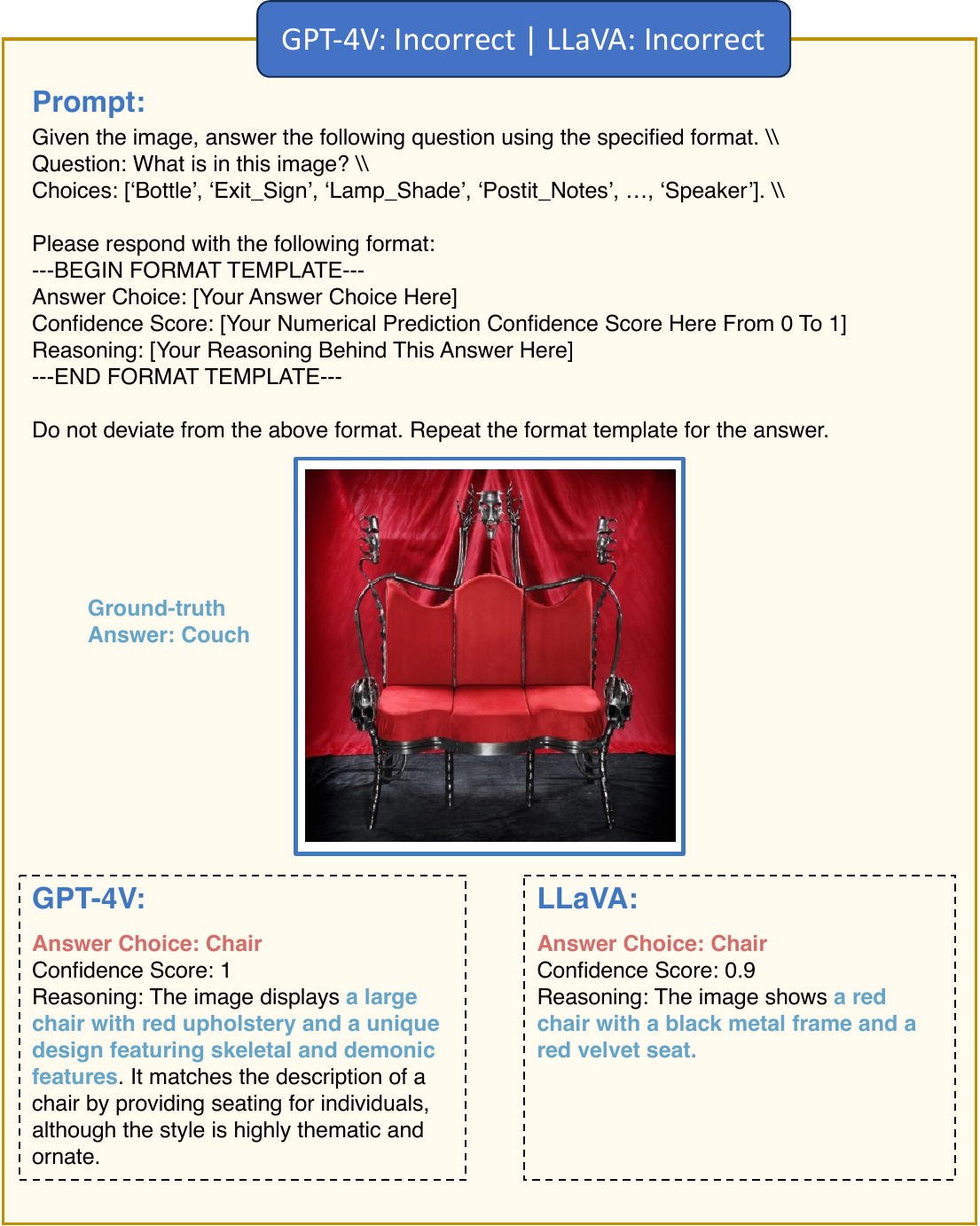}
\caption[Natural Distribution Shift: Case 12 on Office-Home]{Natural Distribution Shift: Case 12 - Couch category in the Art Domain of Office-Home Dataset.
In this instance, both GPT-4V and LLaVA demonstrate detailed and accurate descriptions of the image, yet both models misclassify the object. 
This misclassification arises from the overlapping categories of 'couch' and 'chair' in the dataset, showcasing the challenge models face when distinct class labels share similarities. 
This case highlights the complexity models encounter in accurately categorizing objects within overlapping or closely related classes.
}
\label{060139264231}
\end{figure*}

\begin{figure*}[htb!]
\centering
\includegraphics[width=0.96\textwidth]{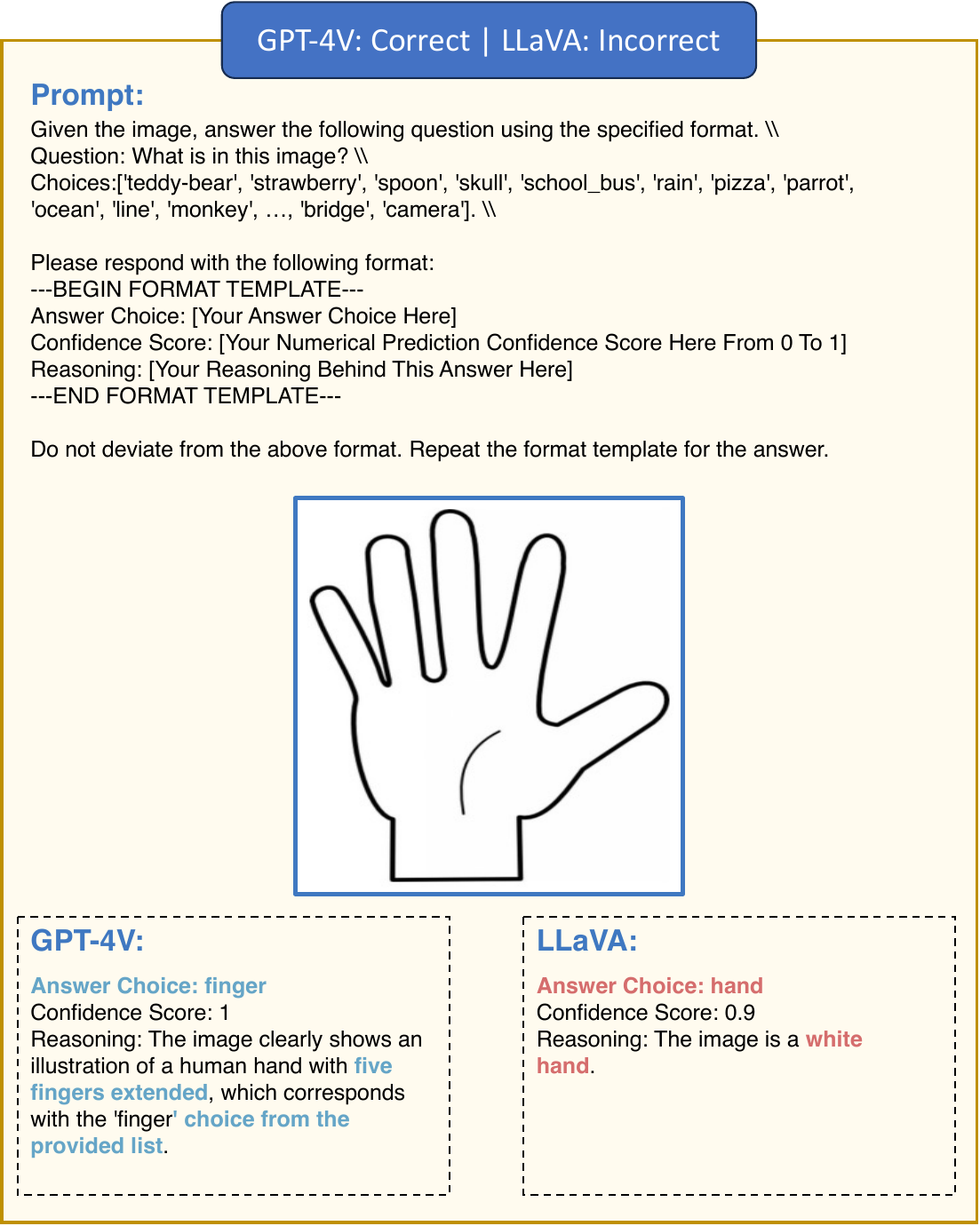}
\caption[Natural Distribution Shift: Case 13 on DomainNet]{Natural Distribution Shift: Case 13 - Analyzing the 'Finger' Category in the Clipart Domain of the DomainNet Dataset.
This figure showcases a challenging example where differentiation between 'Hand' and 'Finger' is crucial. GPT-4V accurately identifies the object as 'Finger', recognizing the illustration of a human hand with all five fingers extended, which aligns with the 'Finger' choice. In contrast, LLaVA identifies the object as 'Hand', providing a less precise classification and a somewhat unusual description of a 'white hand'.
}
\label{211039753973}
\end{figure*}

\begin{figure*}[htb!]
\centering
\includegraphics[width=0.96\textwidth]{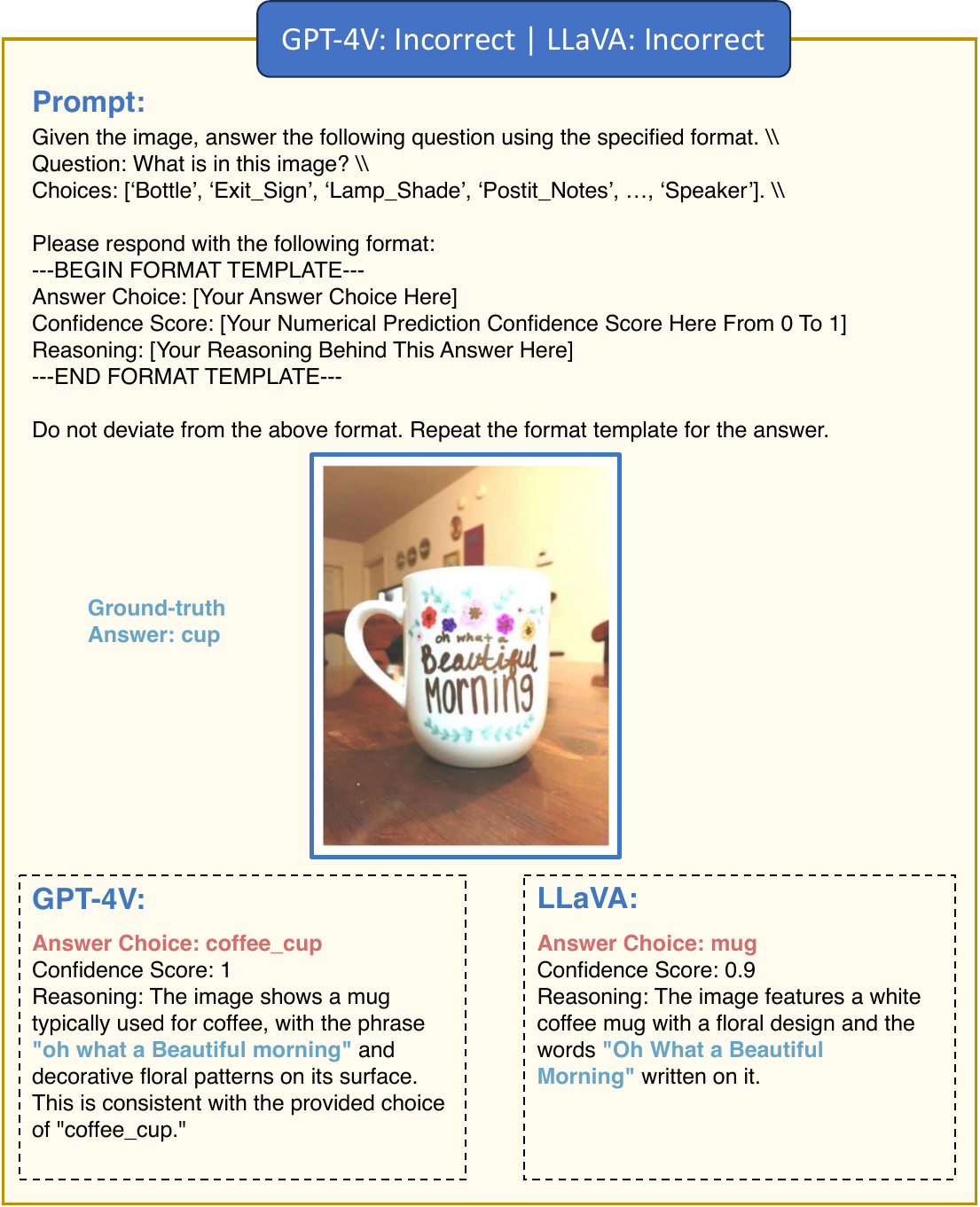}
\caption[Natural Distribution Shift: Case 14 on DomainNet]{Natural Distribution Shift: Case 14 - Analyzing the 'Cup' Category in the Painting Domain of the DomainNet Dataset.
 Despite both GPT-4V and LLaVA present accurate OCR capability, neither GPT-4V nor LLaVA successfully distinguishes the correct category among these closely related concepts, leading to incorrect classifications. 
 This scenario underscores the complexity inherent in nuanced visual recognition tasks, particularly when dealing with objects that share similar characteristics and uses.
}
\label{351223127016}
\end{figure*}

\begin{figure*}[htb!]
\centering
\includegraphics[width=0.96\textwidth]{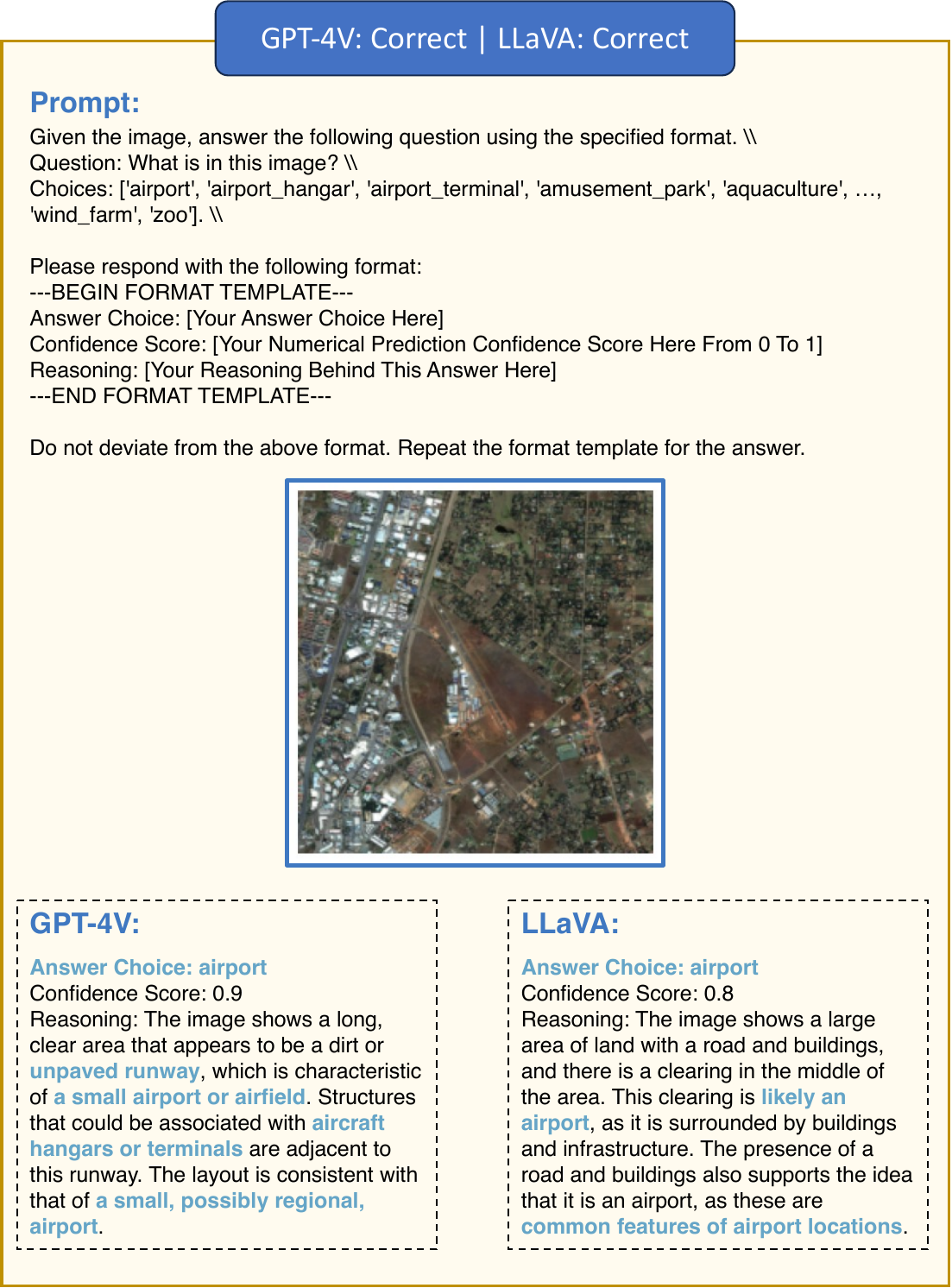}
\caption[Natural Distribution Shift: Case 15 on Fmow]{Natural Distribution Shift: Case 15 - Evaluating the 'Airport' Category in the Region\_2 Domain of the Fmow Dataset.
In this case study, both GPT-4V and LLaVA models are tasked with responding to an identical text prompt accompanied by an image.
Each model correctly identifies 'airport' as the subject, with GPT-4V exhibiting a higher confidence score of 0.9, as opposed to LLaVA's 0.8. 
GPT-4V stands out for its detailed analysis, identifying key elements like runways, aircraft, and terminals, indicative of an airport. 
Remarkably, GPT-4V further distinguishes the airport as small or regional, showcasing its advanced reasoning and contextual interpretation abilities.
}
\label{642890680375}
\end{figure*}

\begin{figure*}[htb!]
\centering
\includegraphics[width=0.96\textwidth]{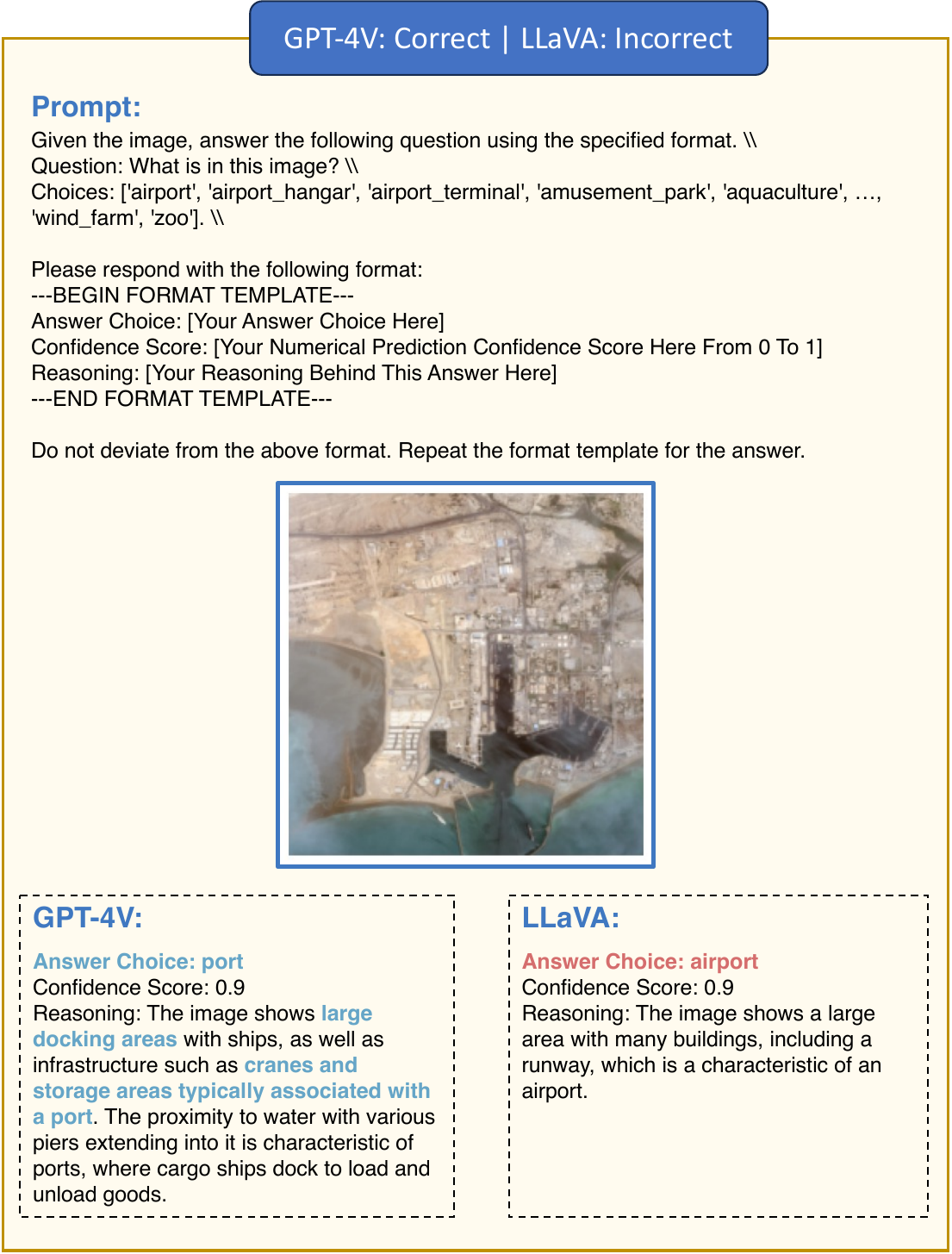}
\caption[Natural Distribution Shift: Case 16 on Fmow]{Natural Distribution Shift: Case 16 - Evaluating the 'Port' Category in the Region\_0 Domain of the Fmow Dataset.
In this instance, GPT-4V accurately identifies the location as a port, citing the presence of docking areas, ships, cranes as key indicators. 
Its reasoning is thorough, focusing on specific port-related characteristics. 
Conversely, LLaVA incorrectly classifies the same image as an airport, referencing runways and buildings, but lacks the detailed analysis of maritime infrastructure present in GPT-4V's description. 
}
\label{728427872260}
\end{figure*}

\begin{figure*}[htb!]
\centering
\includegraphics[width=0.96\textwidth]{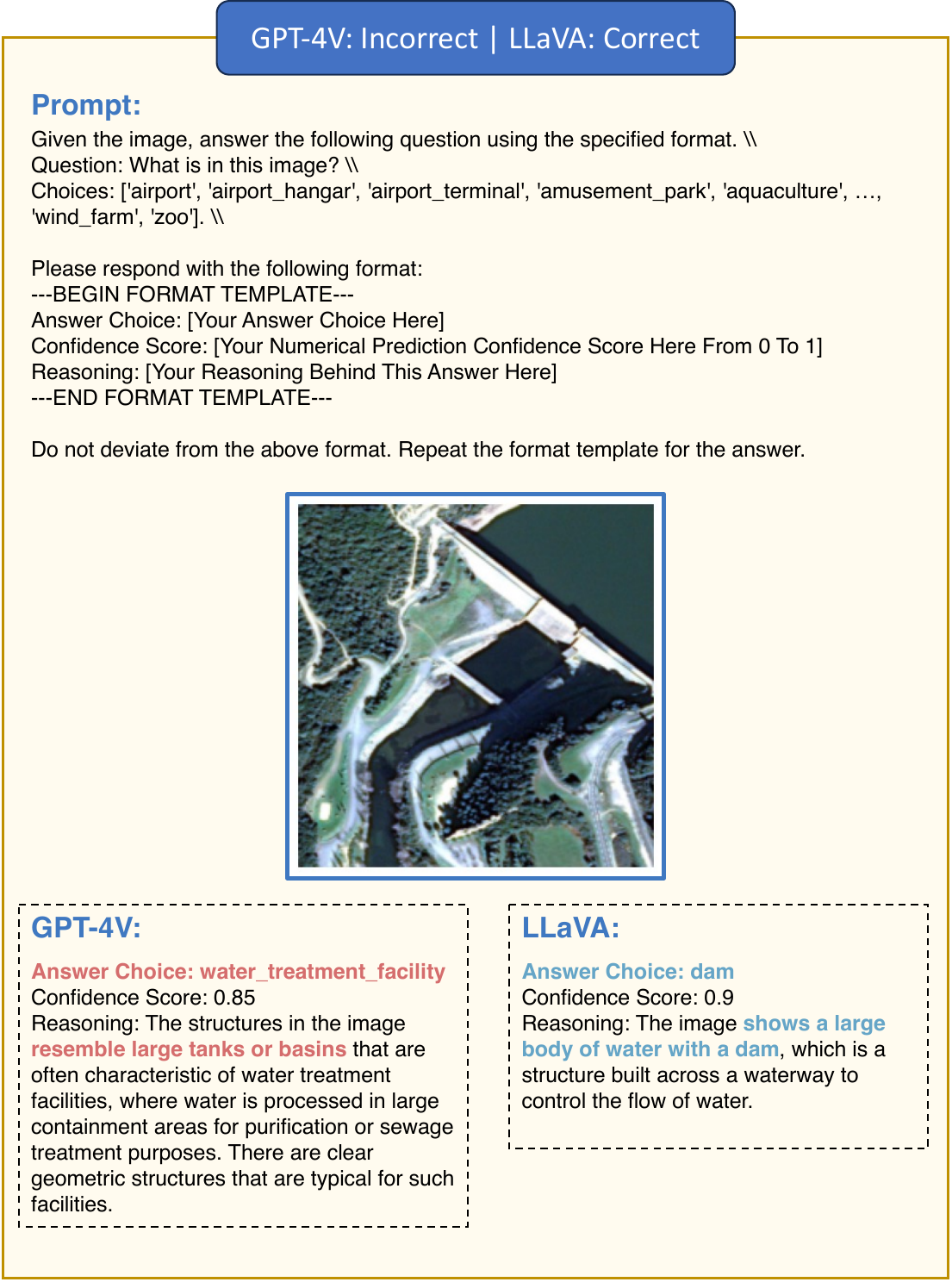}
\caption[Natural Distribution Shift: Case 17 on Fmow]{Natural Distribution Shift: Case 17 - Evaluating the 'Dam' Category in the Region\_3 Domain of the Fmow Dataset.
In this image, GPT-4V incorrectly identifies the scene as a water treatment facility, citing the presence of large tanks or basins typically found in such settings.
Conversely, LLaVA correctly classifies the image as a dam, accurately recognizing the large body of water and the structure controlling its flow, with a confidence score of 0.9. 
}
\label{498038129155}
\end{figure*}

\begin{figure*}[htb!]
\centering
\includegraphics[width=0.96\textwidth]{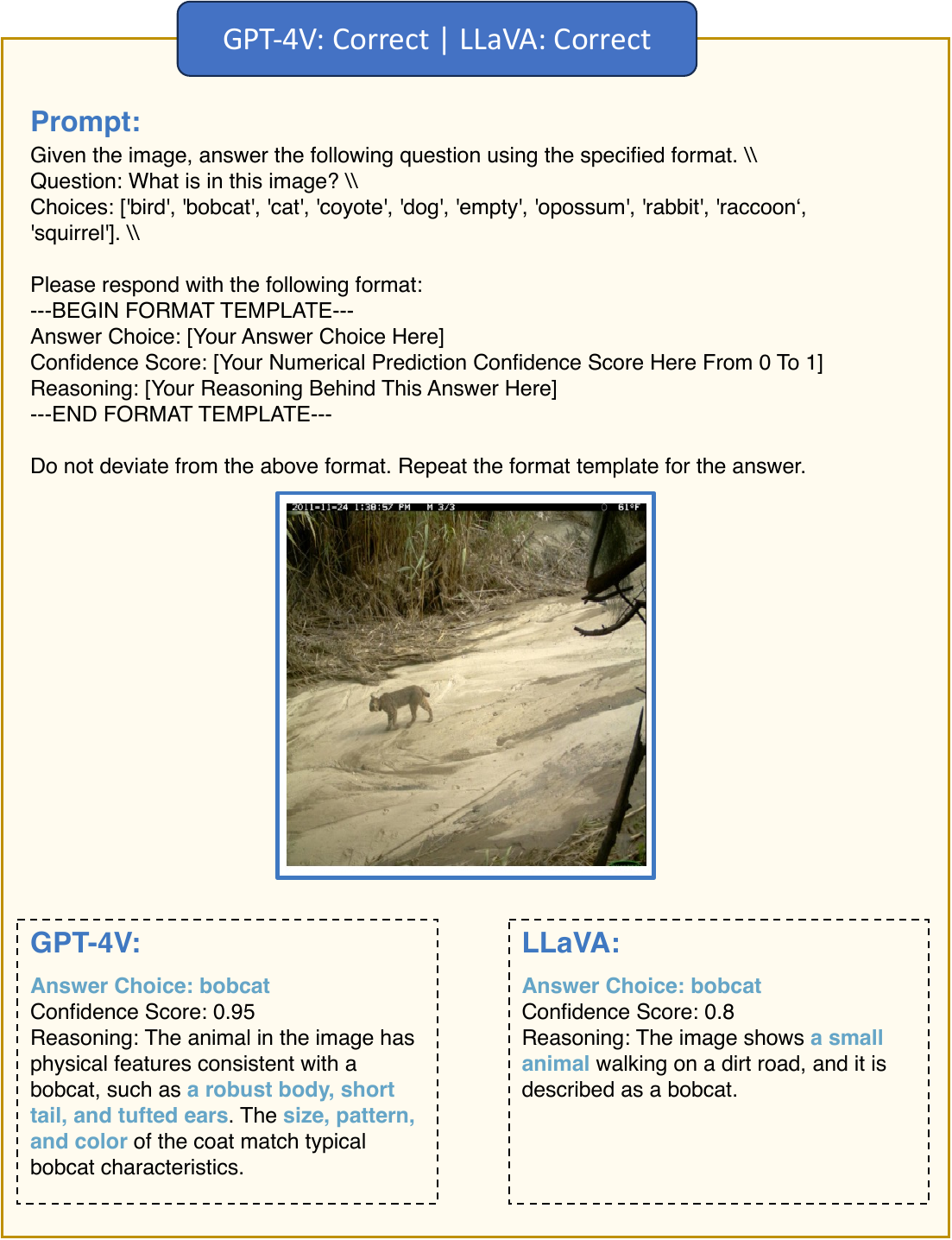}
\caption[Natural Distribution Shift: Case 18 on TerraIncognita]{Natural Distribution Shift: Case 18 - Analyzing the 'Bobcat' Category in Location\_46 Domain of the TerraIncognita Dataset. 
In this evaluation, GPT-4V and LLaVA models respond to a uniform text prompt accompanied by a wildlife image. Both models accurately identify a 'bobcat' as the subject. GPT-4V shows a higher confidence score of 0.95, compared to 0.8 by LLaVA. GPT-4V's reasoning is notable for its detailed analysis, focusing on distinctive physical features of the bobcat, such as a robust body, short tail, and tufted ears, which are challenging to discern even for humans. It also augments its response by detailing the size, pattern, and color of the bobcat, which are crucial for accurate identification. In contrast, LLaVA's identification is based on the general observation of a small animal, a criterion that could apply to multiple species, thereby lacking the specificity demonstrated by GPT-4V.
}
\label{057815033513}
\end{figure*}

\begin{figure*}[htb!]
\centering
\includegraphics[width=0.96\textwidth]{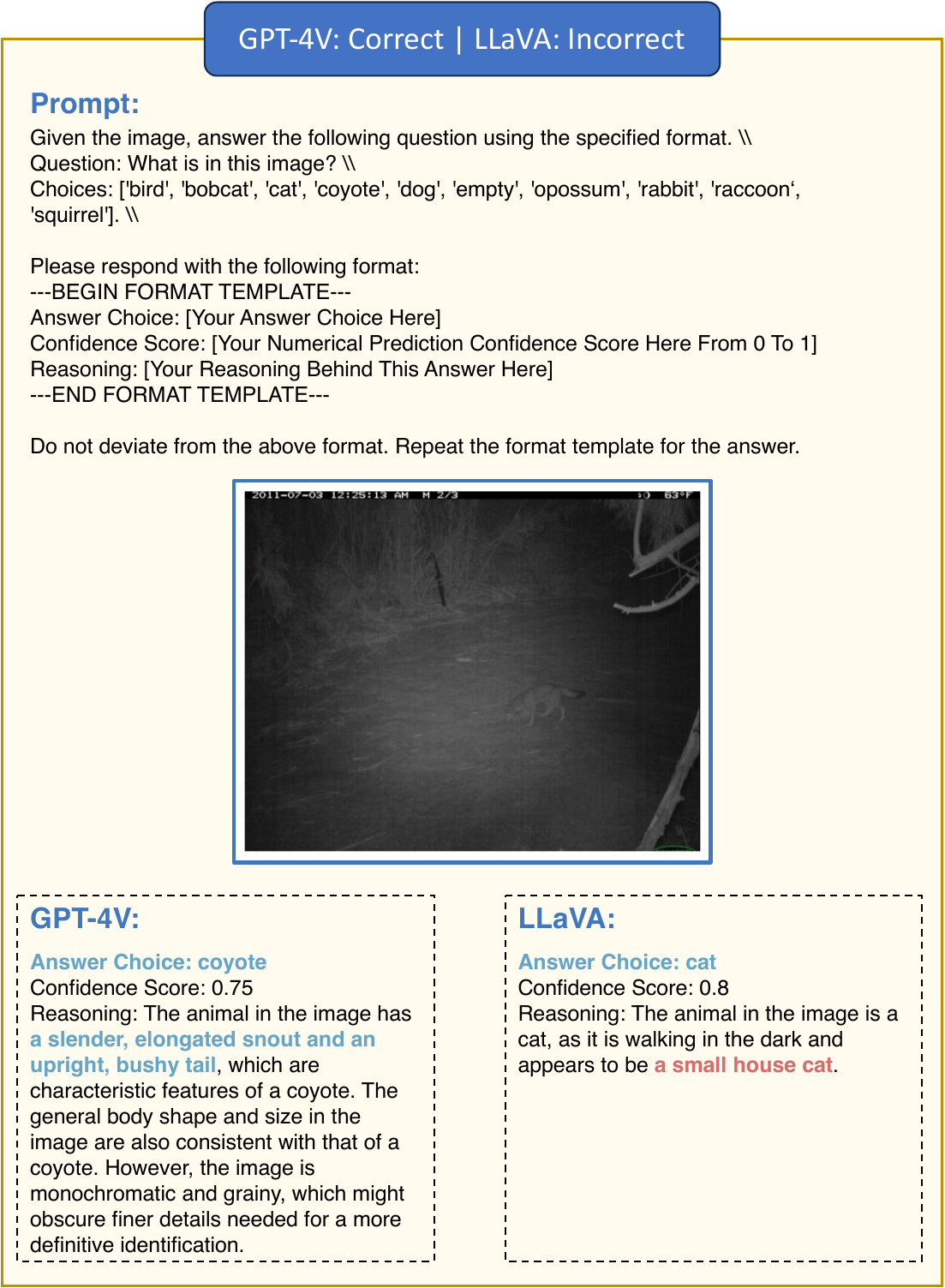}
\caption[Natural Distribution Shift: Case 19 on TerraIncognita]{Natural Distribution Shift: Case 19 - Analyzing the 'Coyote' Category in Location\_46 Domain of the TerraIncognita Dataset. 
In this image, GPT-4V accurately identifies the animal as a coyote, noting its slender, elongated snout and upright, bushy tail, and assigning a confidence score of 0.75. 
It carefully considers the monochromatic and grainy nature of the image that may obscure finer details. 
In contrast, LLaVA incorrectly classifies the animal as a cat with a confidence score of 0.8, failing to recognize the distinct features of a coyote.
}
\label{157994307651}
\end{figure*}

\begin{figure*}[htb!]
\centering
\includegraphics[width=0.96\textwidth]{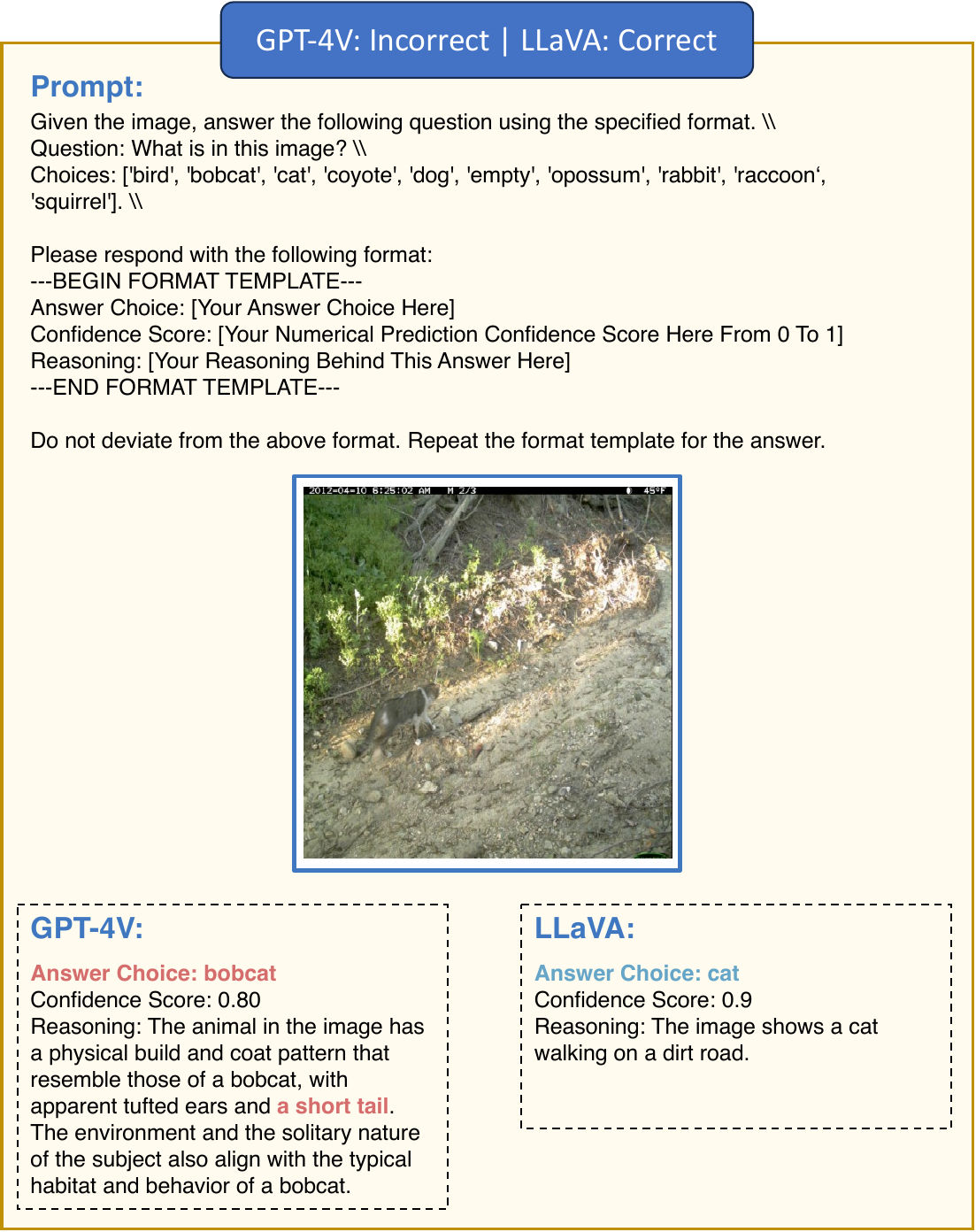}
\caption[Natural Distribution Shift: Case 20 on TerraIncognita]{Natural Distribution Shift: Case 20 - Analyzing the 'Cat' Category in Location\_38 Domain of the TerraIncognita Dataset. 
In this case, LLaVA correctly identifies the animal as a 'cat' with a higher confidence score of 0.9, whereas GPT-4V, with a confidence score of 0.8, mistakenly identifies the animal as a 'bobcat'.
The detailed reasoning of GPT-4V highlights its misclassification: it points to features such as a perceived short tail and tufted ears, typically characteristic of a bobcat, leading to its incorrect conclusion. 
This case illustrates the nuances and challenges of wildlife species recognition, particularly in distinguishing between visually similar animals in natural environments. 
The confidence score outputted by GPT-4V, despite its misclassification in this instance, can serve as a valuable metric, offering insights into the model's decision-making process and potentially guiding reliance on its conclusions.

}
\label{332920742799}
\end{figure*}

\begin{figure*}[htb!]
\centering
\includegraphics[width=0.96\textwidth]{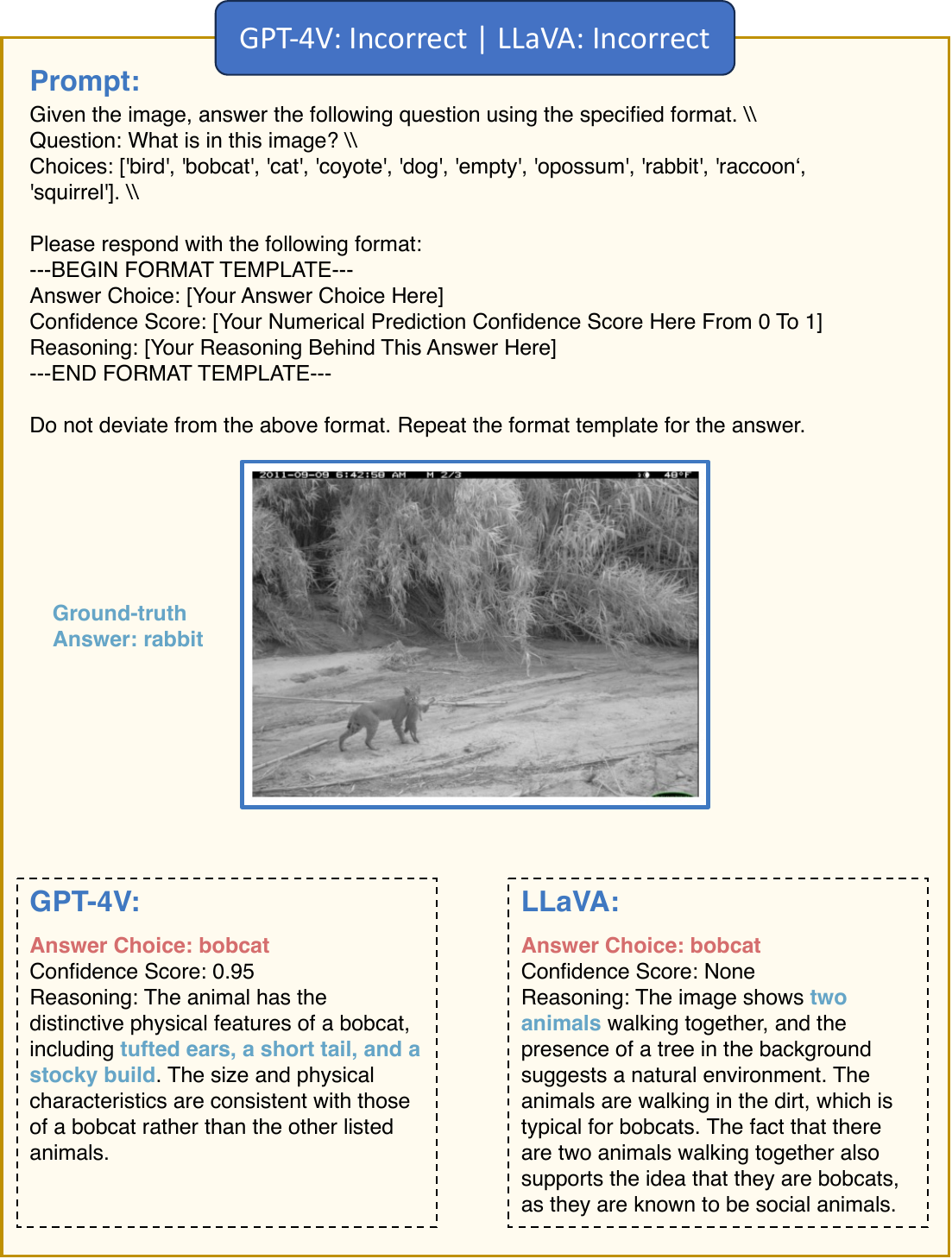}
\caption[Natural Distribution Shift: Case 21 on TerraIncognita]{Natural Distribution Shift: Case 21 - Analyzing the 'Rabbit' Category in Location\_43 Domain of the TerraIncognita Dataset. 
This image presents a complex wildlife scenario where a bobcat is seen capturing and biting a rabbit. Both GPT-4V and LLaVA misinterpret the scene by identifying only the bobcat. This case underlines the intricacies of wildlife recognition, particularly when multiple animals interact in a single frame. The primary focus on the bobcat, while ignoring the rabbit, points to the nuanced challenges in accurately interpreting dynamic natural scenes.
}
\label{242694218402}
\end{figure*}

The diverse array of case studies presented in Figures~\ref{775225593585}, \ref{896371250907}, \ref{289347172711}, \ref{015926979441}, \ref{546069757003}, \ref{340345078522}, \ref{169039403841}, \ref{293135981812}, \ref{044345318906}, \ref{211039753973}, \ref{642890680375}, \ref{728427872260}, \ref{057815033513}, \ref{157994307651} and \ref{332920742799} showcase the adeptness of GPT-4V and LLaVA in navigating the challenges posed by different datasets, including PACS, VLCS, Office-Home, DomainNet, Fmow, and TerraIncognita. 
These examples not only demonstrate GPT-4V's proficiency in accurately recognizing natural distribution shifts in a zero-shot setting but also highlight its ability to adapt to various visual domains and object classifications. 
Additionally, Figures~\ref{830473335836}, \ref{728390971249}, \ref{060139264231}, \ref{351223127016}, \ref{498038129155} and \ref{242694218402} provide insights into instances where GPT-4V does not perform optimally, shedding light on the model's limitations and areas for improvement.

A key observation emerging from these case studies is the nuanced capability of GPT-4V to discern intricate details within images. 
For instance, GPT-4V exhibits its adeptness at identifying textual elements in Figure~\ref{775225593585}. 
Figure~\ref{169039403841} demonstrates a keen eye for specific features, such as the metallic nature and the spout of a kettle, highlighting its attention to detail. 
Furthermore, in Figure~\ref{057815033513}, GPT-4V distinguishes finer characteristics like a short tail and tufted ears in identifying a bobcat, a task that poses a challenge even for human observers.

\subsection{Medical Images}
\begin{table*}[htb!]
\setlength{\abovecaptionskip}{0.cm}
  \caption{Main results of zero-shot generalization performance across distribution shifts on medical and molecule datasets. 
Specifically, CLIP refers to clip-vit-base-patch16, LLaVA refers to llava-v1.5-13b, Gemini refers to gemini-pro-vision, GPT-4V refers to gpt-4-vision-preview.
}
  \label{medical_science}
\centering
\scalebox{0.85}{
\begin{tabular}{lcccccc}
\toprule
Dataset & Camelyon17 & HAM10000 & NIH-Chest  & COVID & DrugOOD\_Assay & DrugOOD\_Scaffold\\ \midrule
Category & medical & medical & medical & medical & molecule & molecule\\
Prediction & tumor & skin diseases & lung disease  & pneumonia types & bioassays & bioassays \\
Domain & hospital & hospital & hospital & hospital & assay & scaffold \\
\#domains & 5 & 4  & 2  & 2 & 81 & 12543 \\
\#classes & 2 & 7 & 15  & 3 & 2 & 2  \\
Examples & \includegraphics[width=2cm]{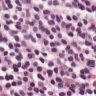}
& \includegraphics[width=2cm]{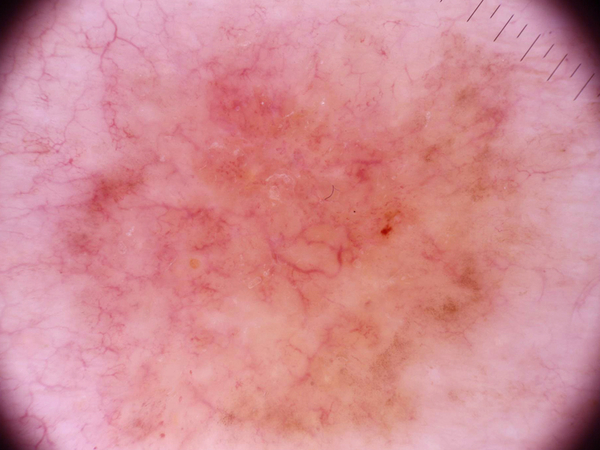} & \includegraphics[width=2cm]{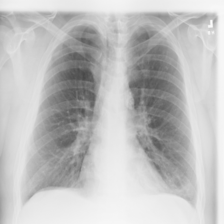} & \includegraphics[width=2cm]{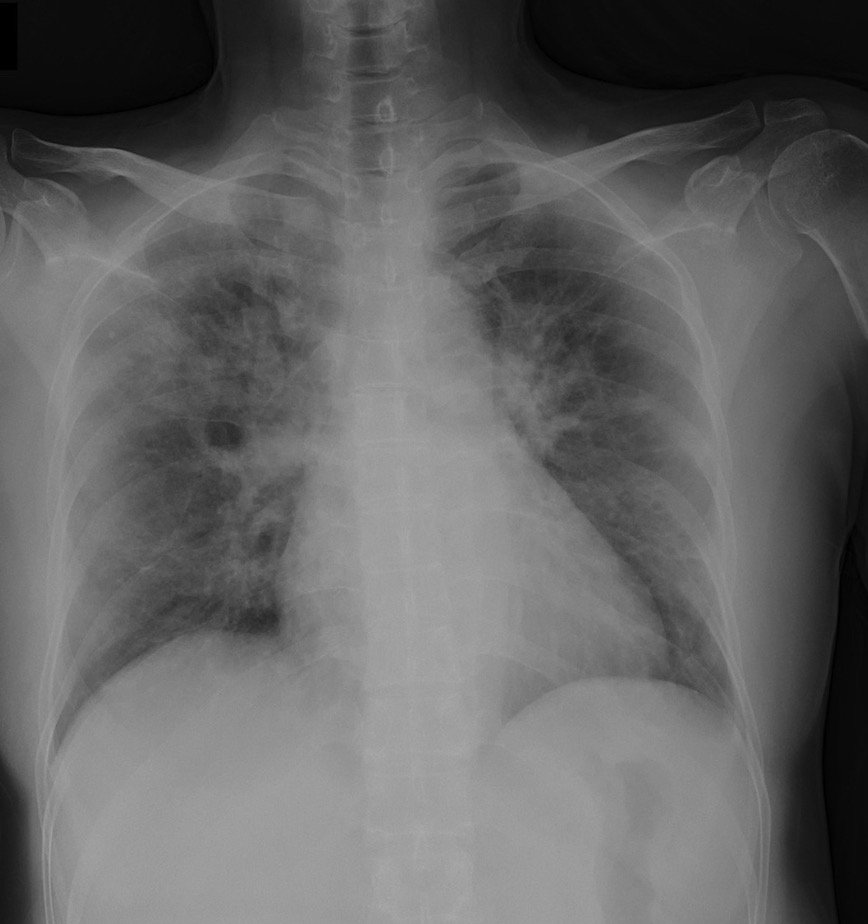}& \includegraphics[width=2cm]{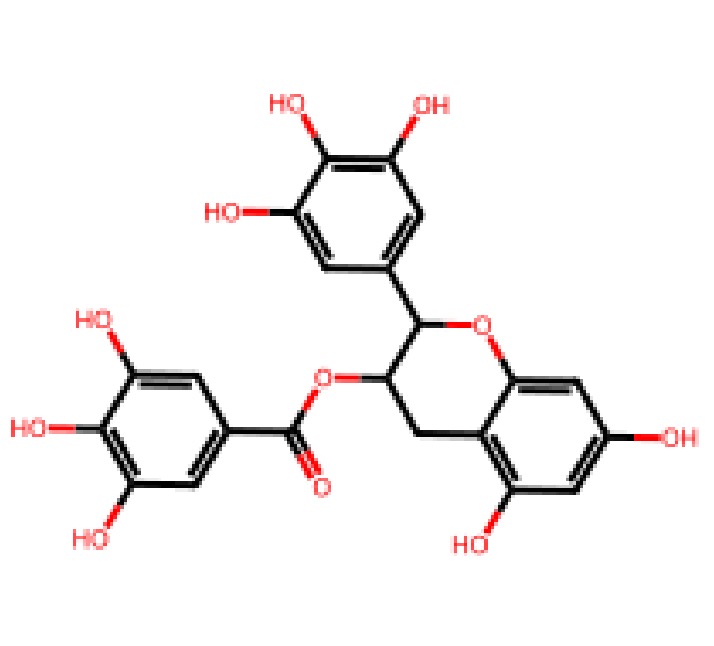} & \includegraphics[width=2cm]{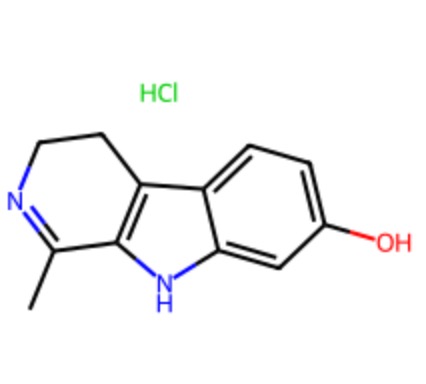} \\
\midrule

& \multicolumn{6}{>{\columncolor{LightBlue}}c}{\textit{random samples (180 cases)}} \\

\multirow{2}{*}{CLIP}  & \cellcolor{gray!20} 0.506  & \cellcolor{gray!20} 0.161 & \cellcolor{gray!20} 0.0  78 &\cellcolor{gray!20}  0.360 &\cellcolor{gray!20}  0.517 &\cellcolor{gray!20}  0.533\\

& \cellcolor{gray!5} 91/180 &\cellcolor{gray!5}  29/180 &\cellcolor{gray!5}  14/180  &\cellcolor{gray!5}  36/100 &\cellcolor{gray!5}  93/180 &\cellcolor{gray!5}  96/180\\ \midrule
\multirow{2}{*}{LLaVA} & \cellcolor{gray!20} 0.508 &\cellcolor{gray!20}  0.100 &\cellcolor{gray!20}  0.044 &\cellcolor{gray!20}  0.450 &\cellcolor{gray!20}  0.517 &\cellcolor{gray!20}  0.533 \\
& \cellcolor{gray!5}  92/180 &\cellcolor{gray!5}  18/180 &\cellcolor{gray!5}  8/180 &\cellcolor{gray!5}  45/100 &\cellcolor{gray!5}  93/180 & \cellcolor{gray!5} 96/180\\ \midrule
\multirow{2}{*}{GPT-4V} & \cellcolor{gray!20} 0.518 & \cellcolor{gray!20} 0.302 &\cellcolor{gray!20} 0.055 &\cellcolor{gray!20} 0.354 &\cellcolor{gray!20} 0.494 &\cellcolor{gray!20} 0.472\\
& \cellcolor{gray!5}  72/139 &\cellcolor{gray!5}  49/162 &\cellcolor{gray!5}  6/108 &\cellcolor{gray!5}  28/79 &\cellcolor{gray!5}  89/180 &\cellcolor{gray!5}  68/144\\
\midrule

\multirow{2}{*}{Gemini} & \cellcolor{gray!20} 0.534 &\cellcolor{gray!20}  0.305 &\cellcolor{gray!20}  0.104 & \cellcolor{gray!20} 0.64 & \cellcolor{gray!20} 0.467 & \cellcolor{gray!20} 0.459\\
 & \cellcolor{gray!5} 94/176 & \cellcolor{gray!5} 53/174 & \cellcolor{gray!5} 17/163 & \cellcolor{gray!5} 64/100 &  \cellcolor{gray!5} 84/180 &  \cellcolor{gray!5} 79/172\\
\midrule 
& \multicolumn{6}{>{\columncolor{LightBlue}}c}{\textit{random samples (1800 cases)}} \\

\multirow{2}{*}{CLIP}  & \cellcolor{gray!20} 0.497  & \cellcolor{gray!20} 0.226 & \cellcolor{gray!20} 0.076  &\cellcolor{gray!20}  0.490 &\cellcolor{gray!20}  0.521 &\cellcolor{gray!20}  0.477\\

& \cellcolor{gray!5} 894/1800 &\cellcolor{gray!5}  406/1800 &\cellcolor{gray!5}  137/1800  &\cellcolor{gray!5}  882/1800 &\cellcolor{gray!5}  924/1772 &\cellcolor{gray!5}  858/1800\\ \midrule
\multirow{2}{*}{LLaVA} & \cellcolor{gray!20} 0.508 &\cellcolor{gray!20}  0.160 &\cellcolor{gray!20}  0.089 &\cellcolor{gray!20}  0.420 &\cellcolor{gray!20}  0.521 &\cellcolor{gray!20}  0.477 \\
& \cellcolor{gray!5}  914/1800 &\cellcolor{gray!5}  288/1800 &\cellcolor{gray!5}  160/1800 &\cellcolor{gray!5}  756/1800 &\cellcolor{gray!5}  923/1772 & \cellcolor{gray!5} 859/1800\\ \midrule
\multirow{2}{*}{GPT-4V} & \cellcolor{gray!20} 0.513 & \cellcolor{gray!20} 0.341 &\cellcolor{gray!20} 0.084 &\cellcolor{gray!20} 0.313 &\cellcolor{gray!20} 0.488 &\cellcolor{gray!20} 0.514\\
& \cellcolor{gray!5}  923/1799 &\cellcolor{gray!5}  548/1606 &\cellcolor{gray!5}  45/535 &\cellcolor{gray!5}  380/1216 &\cellcolor{gray!5}  414/848 &\cellcolor{gray!5}  647/1258\\

\midrule 
\multirow{2}{*}{Gemini} & \cellcolor{gray!20} 0.532 &\cellcolor{gray!20}  0.335 &\cellcolor{gray!20}  0.119 & \cellcolor{gray!20} 0.515 & \cellcolor{gray!20} 0.490 & \cellcolor{gray!20} 0.508\\
 & \cellcolor{gray!5} 940/1766 & \cellcolor{gray!5} 572/1705 & \cellcolor{gray!5} 206/1729 & \cellcolor{gray!5} 926/1798 &  \cellcolor{gray!5} 869/1772 &  \cellcolor{gray!5} 914/1800\\
 
\midrule

 & \multicolumn{6}{>{\columncolor{LightOrange}}c}{\textit{failure cases}} \\

\multirow{2}{*}{CLIP}  & \cellcolor{gray!20} 0.000  &\cellcolor{gray!20} 0.000 &\cellcolor{gray!20} 0.000 &\cellcolor{gray!20} 0.000 &\cellcolor{gray!20} 0.000 &\cellcolor{gray!20} 0.000\\
& \cellcolor{gray!5}  0/180 & \cellcolor{gray!5} 0/180 &\cellcolor{gray!5}  0/180 &\cellcolor{gray!5}  0/100 &\cellcolor{gray!5}  0/180 &\cellcolor{gray!5}  0/180\\ \midrule
\multirow{2}{*}{LLaVA}  & \cellcolor{gray!20} 0.028 &\cellcolor{gray!20} 0.067 &\cellcolor{gray!20} 0.056 &\cellcolor{gray!20} 0.510 &\cellcolor{gray!20} 0.000 &\cellcolor{gray!20} 0.006\\
& \cellcolor{gray!5}  5/180 &\cellcolor{gray!5}  12/180 &\cellcolor{gray!5}  10/180 &\cellcolor{gray!5}  51/100 &\cellcolor{gray!5}  0/180 &\cellcolor{gray!5}  1/180\\ \midrule
\multirow{2}{*}{GPT-4V} & \cellcolor{gray!20} 1.000 &\cellcolor{gray!20} 0.308 &\cellcolor{gray!20} 0.102 &\cellcolor{gray!20} 0.543 &\cellcolor{gray!20} 1.000 &\cellcolor{gray!20} 1.000\\
& \cellcolor{gray!5}  157/157 & \cellcolor{gray!5}  49/159 & \cellcolor{gray!5}  6/59 & \cellcolor{gray!5}  38/70 & \cellcolor{gray!5}  179/179 & \cellcolor{gray!5}  180/180\\
\midrule

\multirow{2}{*}{Gemini} & \cellcolor{gray!20} 1.000 &\cellcolor{gray!20}  0.28 &\cellcolor{gray!20}  0.093 & \cellcolor{gray!20} 0.93 & \cellcolor{gray!20} 1.000 & \cellcolor{gray!20} 1.000 \\
 & \cellcolor{gray!5} 176/176 & \cellcolor{gray!5} 49/175 & \cellcolor{gray!5} 14/150 & \cellcolor{gray!5} 93/100 &  \cellcolor{gray!5} 180/180 & \cellcolor{gray!5} 180/180\\

\bottomrule 

\end{tabular}}
\end{table*}

\subsubsection{Task Introduction}

We investigate the classification capabilities of different models in medical imaging applications under scenarios of distributional shifts. Distributional shifts are particularly common in the field of medical imaging, as changes in imaging technology, patient demographic characteristics, and disease manifestation can significantly alter the data distribution. Exploring the generalizability of the GPT-4 vision large model in medical image analysis tasks holds significant practical value.

In this part, we examine the following medical datasets, each with its distinct characteristics and challenges:

\begin{itemize}
    \item \textbf{Camelyon17~\citep{bandi2018detection}:} The dataset contains 450,000 patch samples, which were derived from 50 whole-slide images (WSIs) featuring breast cancer metastases in lymph node sections. These WSIs were sourced from five different hospitals in the Netherlands, contributing 10 WSIs each. Pathologists meticulously annotated each WSI to identify tumor regions, and these annotations were used to create segmentation masks. These masks, in turn, provided the basis for assigning labels to each individual patch in the dataset.
    \item \textbf{HAM10000~\citep{tschandl2018ham10000}:} The dataset is a critical resource for research in skin lesion analysis, particularly focusing on generalization tasks. This dataset features a wide variety of dermatoscopic images, including numerous skin lesion types such as melanoma, basal cell carcinoma, and benign nevi. It is especially valuable for training and evaluating machine learning models on skin cancer detection and diagnosis. The diversity of images, sourced from different populations and equipment, makes HAM10000 ideal for studying and improving OOD generalization in medical imaging algorithms. This aspect is crucial for developing robust models capable of performing accurately across varied and unseen data, reflecting real-world clinical scenarios.

    \item \textbf{NIH-Chest~\citep{wang2017chestx}:} The NIH Chest X-ray Dataset, a substantial medical imaging collection from the National Institutes of Health, is pivotal for research in out-of-distribution (OOD) generalization and distribution shift challenges in medical imaging. Comprising over 112,000 frontal-view X-ray images from more than 30,000 patients, this dataset is annotated with 14 common thoracic pathologies, such as pneumonia and lung nodules. Its vast and diverse array of patient images, captured under various clinical settings and conditions, provides an exceptional resource for developing and testing machine learning models, particularly in assessing and improving their robustness and performance in the face of distributional shifts and OOD data, which are common obstacles in real-world medical diagnostics.

    \item \textbf{COVID~\citep{han2021semi}:} This dataset serves as a resource for pneumonia detection, encompassing samples of normal cases, typical pneumonia, and COVID-19 pneumonia. The data, sourced from various hospitals due to collection methodologies, exhibit distributional shifts. We utilize this dataset to assess model performance in pneumonia detection tasks under conditions of distributional shift, reflecting real-world variations in medical data collection and patient demographics.

\end{itemize}

These datasets encompass a diverse array of medical scenarios and tasks, while also presenting a variety of distribution shifts. This diversity positions them as prime candidates for assessing the zero-shot generalization abilities of the GPT-4V model, with comparative analysis against CLIP, LLaVA, and Gemini. Table~\ref{medical_science} offers a comprehensive overview of each dataset, highlighting crucial elements like prediction types, domain specifics, the range of domains and classes, along with representative examples. 

\subsubsection{Comparative Accuracies Across Datasets and Domains}
Table~\ref{medical_science} outlines the accuracies and correct-to-total case ratios for three models (CLIP, LLaVA, and GPT-4V) across four medical datasets, incorporating both random samples and failure cases identified in CLIP. This subsection is dedicated to examining GPT-4V's zero-shot generalization abilities within medical datasets.

\textbf{GPT-4V's Performance in Random Samples:}
According to Table~\ref{medical_science}, it is observed that the performance of GPT-4V, Gemini, CLIP, and LLaVA on medical image classification tasks is quite average. For instance, on the Camelyon17 dataset, the performances of GPT-4V, Gemini, CLIP, and LLaVA are 0.513, 0.532, 0.497, and 0.508, respectively. This suggests that the data from these datasets may not have been present in the training sets of these four models, highlighting a potential gap in their pre-training data and indicating the need for further model training or adaptation to improve performance in these specific medical tasks.

\textbf{GPT-4V in Handling CLIP's Failure Cases:} To assess GPT-4V’s capabilities in more challenging scenarios, we examine its performance in CLIP’s failure cases. On the HAM10000 dataset, GPT-4V achieved an accuracy of 0.308, surpassing LLaVa's 0.067. There were also varying degrees of accuracy improvements on the NIH-Chest and COVID datasets. These results demonstrate GPT-4V's robustness in handling complex and challenging visual tasks, maintaining stable performance even in scenarios where CLIP struggled. 

\subsubsection{Case Demonstration}
\begin{figure*}[htb!]
\centering
\includegraphics[width=0.96\textwidth]{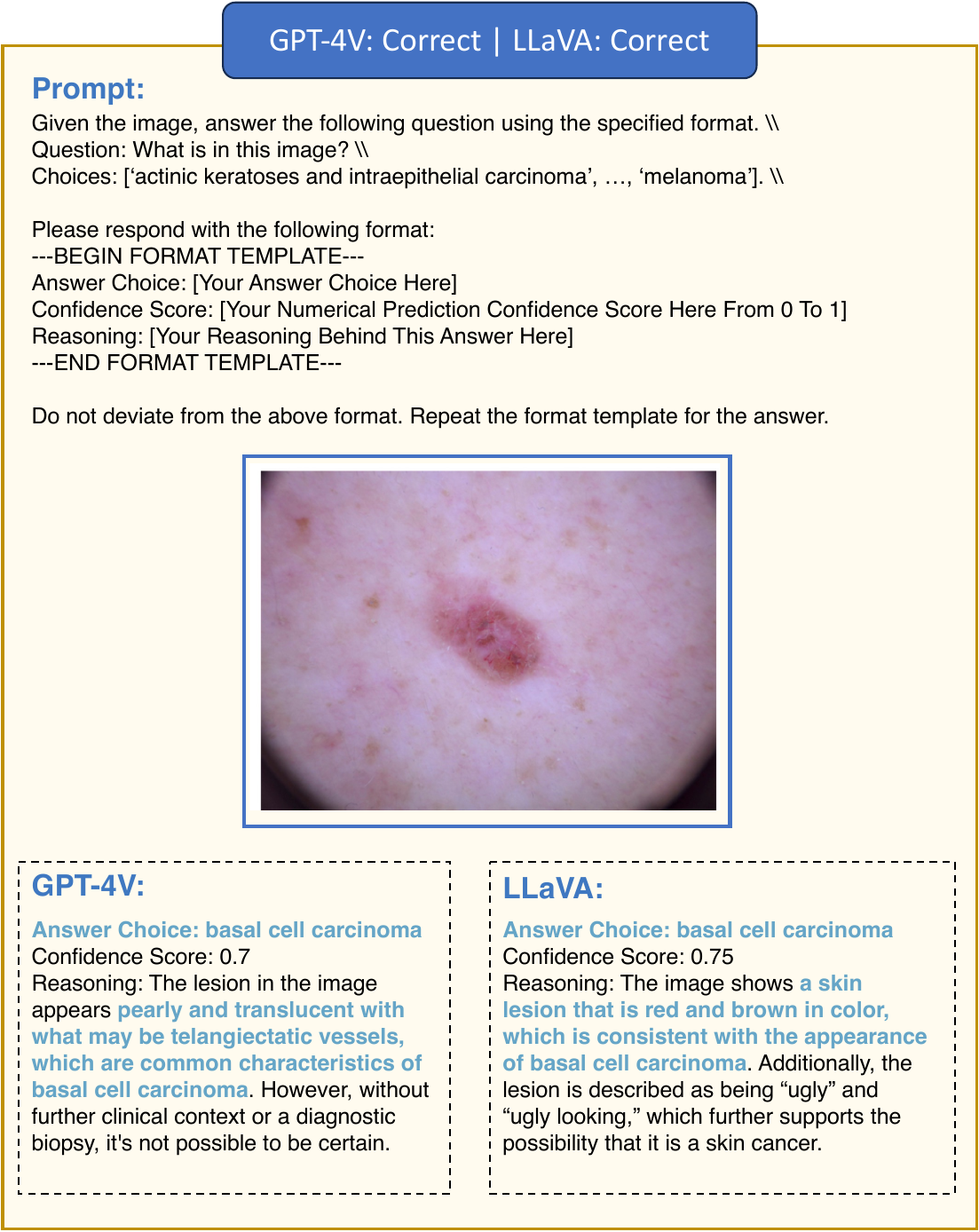}
\caption[Medical Distribution Shift: Case 1 on HAM10000]{Medical Distribution Shift: Case 1 - Analyzing the `basal cell carcinoma' Category in the vidir\_modern Domain of the HAM10000 Dataset. In this case study, both GPT-4V and LLaVA models are tasked with responding to an identical text prompt accompanied by an image.
Each model correctly identifies `basal cell carcinoma' as the subject, with LLaVA exhibiting a higher confidence score of 0.75, as opposed to GPT-4V's 0.7. 
GPT-4V stands out for its detailed analysis, identifying key elements like pearly, translucent, and telangiectatic vessels, indicative of basal cell carcinoma. LLaVA gives an analysis mainly in terms of color appearance. 
}
\label{HAM10000_random_case}
\end{figure*}

\begin{figure*}[htb!]
\centering
\includegraphics[width=0.96\textwidth]{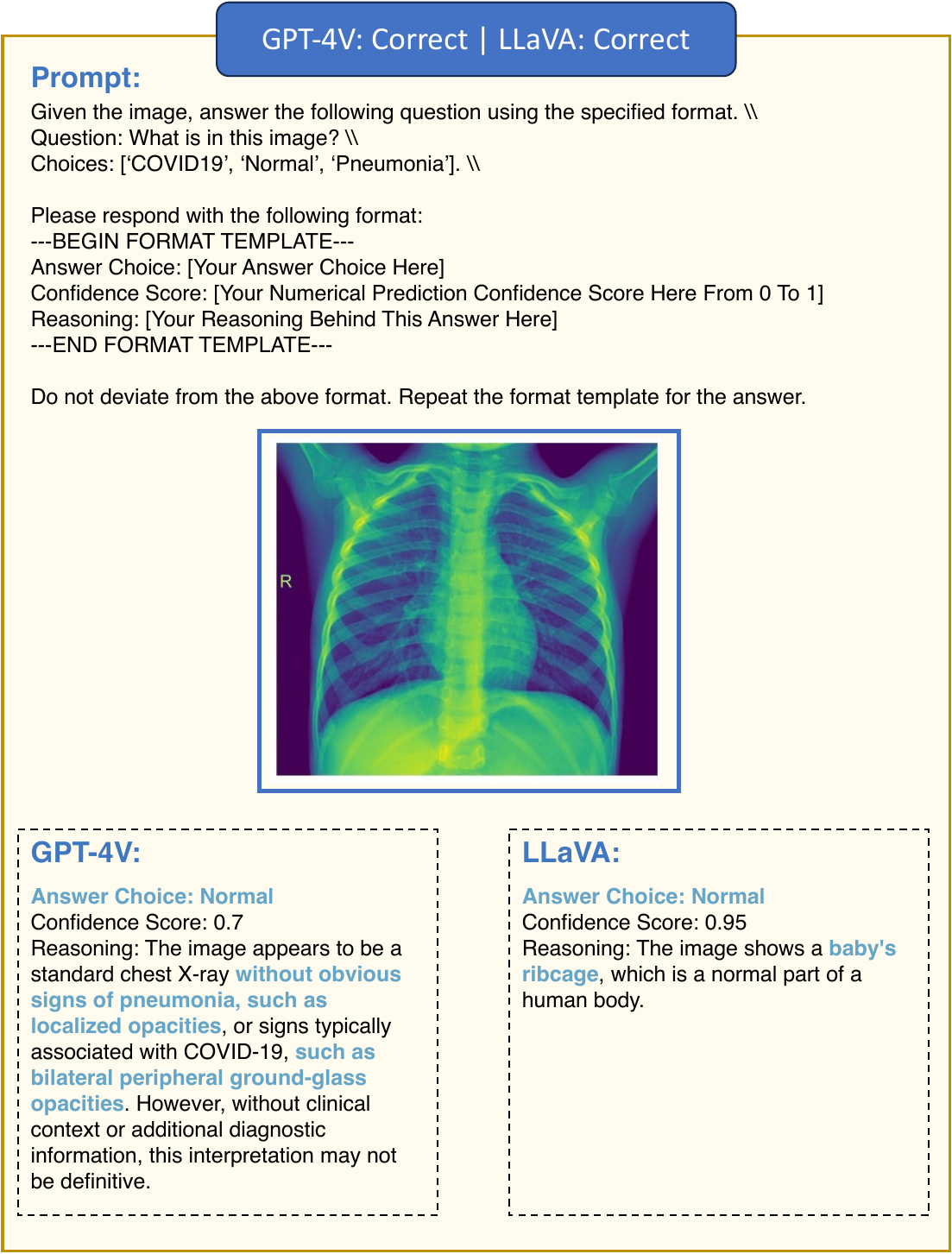}
\caption[Medical Distribution Shift: Case 2 on COVID]{Medical Distribution Shift: Case 2 - Analyzing the `Normal' Category in the hostial1 Domain of the COVID Dataset. In this case study, both GPT-4V and LLaVA models are tasked with responding to an identical text prompt accompanied by an image.
Each model correctly identifies `Normal' as the subject, with LLaVA exhibiting a higher confidence score of 0.95, as opposed to GPT-4V's 0.7. 
GPT-4V stands out for its detailed analysis, identifying key elements like  localized opacities, and bilateral peripheral ground-glass opacities, indicative of a normal sample. LLaVA states that the image is of a baby's rib cage, but does not give a proper reason for why it is categorized as normal, despite the high confidence score of 0.95.}
\label{COVID_random_case}
\end{figure*}

The diverse array of case studies presented in Figures~\ref{HAM10000_random_case} and \ref{COVID_random_case} showcase the adeptness of GPT-4V and LLaVA in navigating the challenges posed by different datasets, including HAM10000, NIH-Chest, and COVID. 

\subsection{Scientific Images}
\subsubsection{Task Introduction}
Our research investigates the performance of various computational models in scientific fields, with a focus on predicting molecular properties amid distributional shifts due to variations in scaffolds and assays. Such shifts, resulting from changes in molecular scaffolds and assay conditions, profoundly affect the nature of scientific datasets. Assessing how advanced models like GPT-4 can adapt to these variations is vital for enhancing their predictive accuracy and reliability in the dynamic landscape of molecular science, where the intricate interplay of molecular structure and assay environments shapes data diversity and complexity.

In this part, we examine the following scientific datasets, each with its distinct characteristics and challenges:

DrugOOD~\citep{ji2023drugood} is a comprehensive dataset curator and benchmarking tool specifically designed for AI-aided drug discovery (AIDD). It focuses on the critical challenge of drug-target binding affinity prediction, involving both macromolecules (protein targets) and small molecules (drug compounds). Unlike traditional fixed datasets, DrugOOD offers automated data curation with customizable scripts, rich domain annotations, realistic noise annotations, and robust benchmarking of state-of-the-art OOD algorithms. It is particularly useful for testing graph-based out-of-distribution learning problems, crucial in molecular data modeled as irregular graphs. DrugOOD\_Assay and DrugOOD\_Scaffold can be obtained by splitting the domains with assays and scaffolds. 
\begin{itemize}
    \item \textbf{DrugOOD\_Assay~\citep{ji2023drugood}:} In the DrugOOD\_Assay, domains are delineated based on the assay. This means that samples generated from the same assay are classified into the same domain, reflecting the unique environmental conditions of each assay. Due to these varying conditions, activity values measured across different assays exhibit a natural distribution shift. Consequently, the model is challenged to perform on data from bioassay environments it has not previously seen, testing its ability to generalize and maintain accuracy in the face of diverse and novel assay environments.
    \item \textbf{DrugOOD\_Scaffold~\citep{ji2023drugood}:} In the DrugOOD\_Scaffold dataset, the domains are defined based on different molecular scaffolds. Molecules with the same molecular scaffold are grouped into the same domain, following the approach outlined by \citep{koh2021wilds, hu2021kinasemd}. This structuring emphasizes the importance for models to have the capability to generalize effectively to unseen domains that are characterized by novel scaffolds, thereby enabling accurate predictions across a broad spectrum of molecular structures.
\end{itemize}

These datasets encompass a diverse array of scientific scenarios, while also presenting a variety of distribution shifts. This diversity positions them as prime candidates for assessing the zero-shot generalization abilities of the GPT-4V model, with comparative analysis against CLIP and LLaVA. Table~\ref{medical_science} offers a comprehensive overview of each dataset, highlighting crucial elements like prediction types, domain specifics, the range of domains and classes, along representative examples. 

\subsubsection{Performance Across Datasets and Domains}
The results show that, in both the DrugOOD\_Assay and DrugOOD\_Assay datasets, GPT-4V, CLIP, and LLaVA failed. They were ineffective in accurately predicting the categories of molecules. The reasons for their failures could be attributed to three main factors: First, the complexity of the scientific task. Second, these datasets were not included in the training sets of these three models. Third, the ambiguity in data labeling, for instance, the labels `inactive' and `active' in scientific datasets are different from natural dataset labels like `elephant' or `bike'. The use of `inactive' and `active' as class labels is more ambiguous and lacks specific meaning. In conclusion, it is understandable that the zero-shot classification capabilities of these three models are poor.

\subsubsection{Prompt Engineering Trick}

\begin{figure}[tb!]
\centering
\includegraphics[width=.9\textwidth]{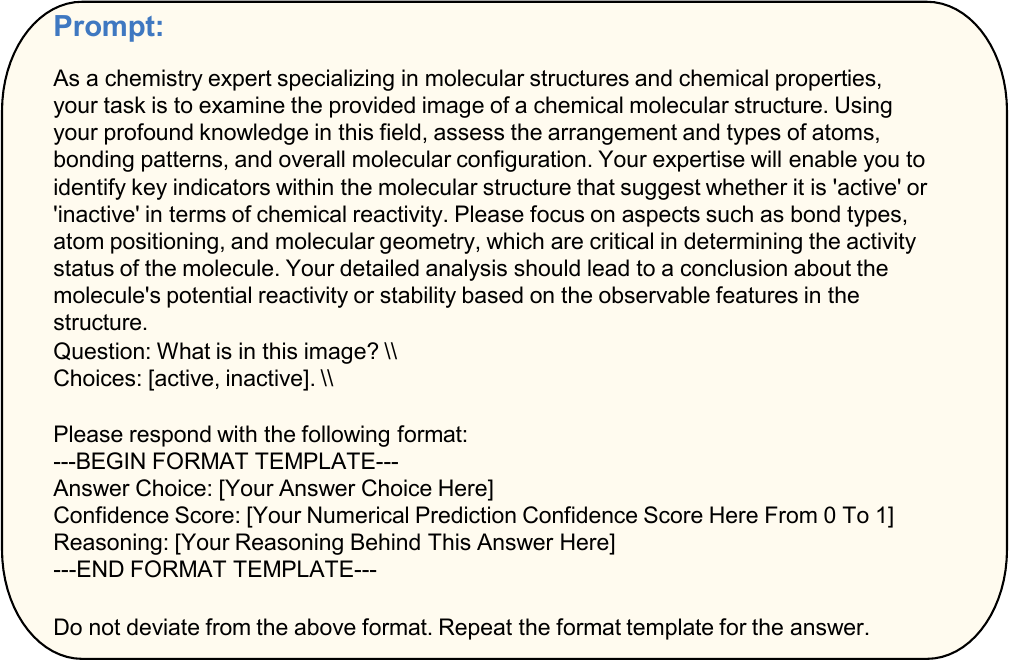}
\caption[Illustration of a structured prompt format]
{An illustration of a structured prompt format used in the PACS dataset, showcasing a specific approach for image-based questioning and response formatting.
The format includes a question about the image's content, a list of answer choices, and a template for answering, including an answer, confidence score, and the reasoning process.}
\label{prompt_engineer}
\end{figure}

This study explores the significant role of the Prompt Engineering Trick in enhancing performance in scientific image classification tasks. Specifically, we applied this technique in the task of chemical structure-activity classification, achieving a notable improvement in classification accuracy from 51.4\% to 52.5\%. This approach involves introducing meticulously designed prompts, such as instructing the model to analyze molecular structure images in the role of a chemistry expert, as shown in Figure~\ref{prompt_engineer}. We required the model to not only identify atomic arrangements and bonding patterns in the images but also to interpret the overall configuration of the molecule to determine its chemical reactivity as either active or inactive. This method not only improved classification accuracy but also made the model's reasoning process more logical and interpretable. This research demonstrates that carefully designed prompts can significantly enhance the performance and understanding of machine learning models in specific tasks.

\subsubsection{Case Demonstration}
\begin{figure*}[htb!]
\centering
\includegraphics[width=0.9\textwidth]{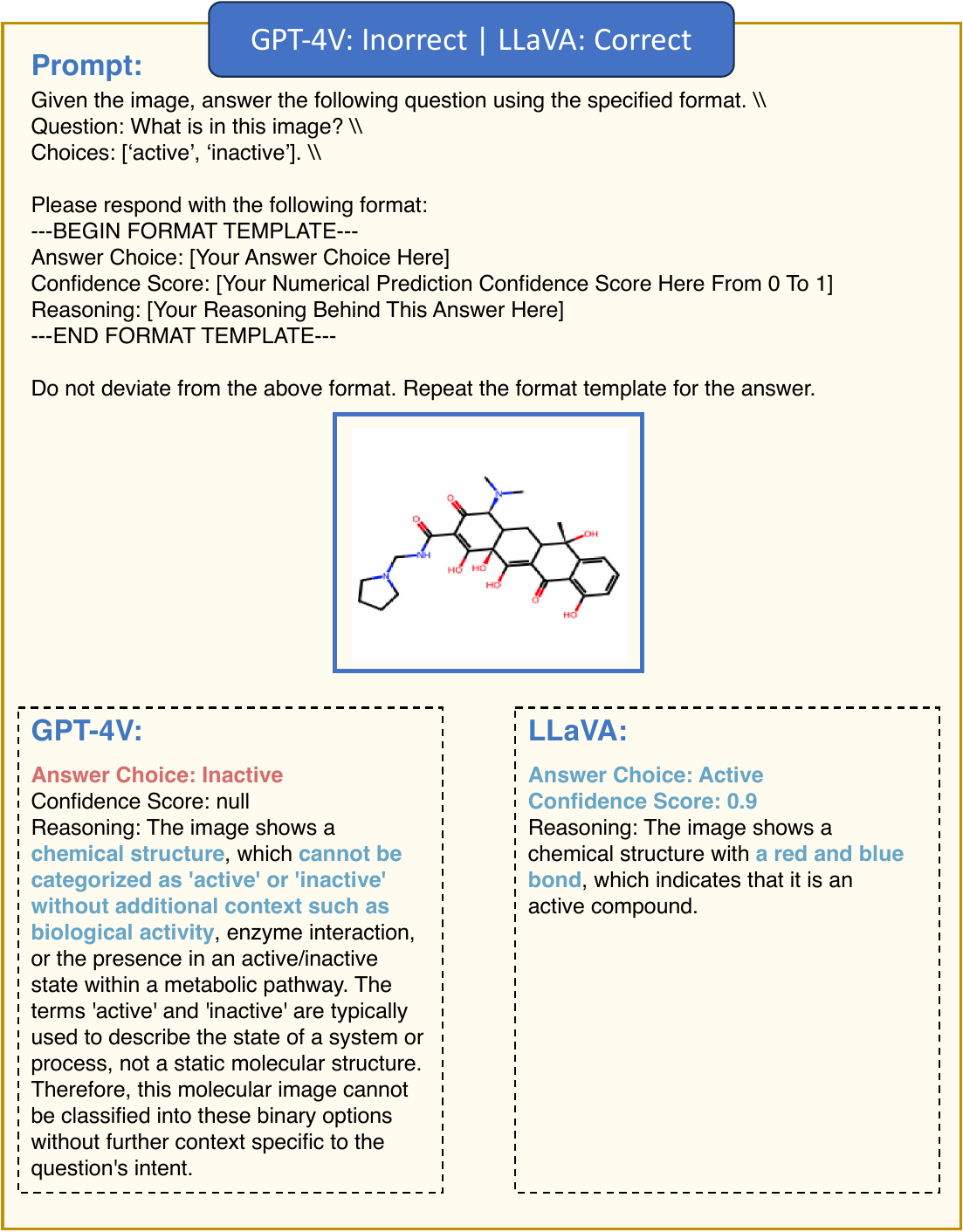}
\caption[Scientific Distribution Shift: Case 1 on DrugOOD\_assay]{Scientific Distribution Shift: Case 1 - Analyzing the `active' Category in the ID-75 Domain of the DrugOOD\_assay Dataset. In this case study, both GPT-4V and LLaVA models are tasked with responding to an identical text prompt accompanied by an image. GPT-4V incorrectly predicts without confidence score while LLaVA correctly predicts with high confidence. Although the GPT-4V predicts error categories, it does not give high confidence. According to reason, GPT-4V can recognize this image as a chemical structure. The statement that it cannot be categorized as `active' or `inactive' without other contexts such as biological activity, enzyme interactions, or active/inactive states in metabolic pathways makes sense. The limitations of the DrugOOD dataset itself are also noted here, i.e., it is not reasonable to simply categorize the data as `active' or `inactive'. Conversely, LLaVA, despite giving correct predictions, is very unreliable in its reasoning. It is wrong to classify them as `active' or `inactive' by the color of the bond.
}
\label{drugood_assay_random_case}
\end{figure*}

The representative case study presented in Figure~\ref{drugood_assay_random_case} showcases the adeptness of GPT-4V and LLaVA in navigating the challenges. The results in Figures~\ref{drugood_assay_random_case} show that GPT-4V does not perform well in predicting molecular properties. Although LLaVA can correctly predict the molecular properties, its reasoning is not convincing, suggesting that LLaVA's correct predictions are merely guesses without any solid basis. In contrast, although GPT-4V does not make accurate predictions, it does not provide a confidence level, and its reasoning is more logical. Therefore, to some extent, GPT-4V is more reliable than LLaVA.

\section{Adaptability to Controlled Data Perturbations}
\label{057362393941}
To assess GPT-4V's adaptability to entirely new distribution shifts, our methodology encompasses two distinct strategies: (1) Noise Injection: We introduce Gaussian noise into the PACS, VLCS, and Office-Home datasets to artificially create variations in data distribution. (2) Domain Shift Generation: Utilizing ControlNet~\cite{zhang2023adding}, we generate datasets that exhibit domain shifts. These datasets are designed to significantly differ from those used in the pretraining phase and are not encountered by the model during its initial training. This approach allows us to systematically evaluate GPT-4V's performance across datasets that vary substantially from the pretraining data.

\subsection{Gaussian Noise}

\begin{table*}[htb]
\setlength{\abovecaptionskip}{0.cm}
  \caption{Main results of zero-shot generalization performance across distribution shifts created by adding Gaussian noise. Specifically, CLIP refers to clip-vit-base-patch16, LLaVA refers to llava-v1.5-13b, Gemini refers to gemini-pro-vision, GPT-4V refers to gpt-4-vision-preview.
}
  \label{gaussian}
\centering
\scalebox{0.77}{
\begin{tabular}{lccc|ccc}
\toprule
Dataset & PACS\_gaussian & VLCS\_gaussian & Office-Home\_gaussian & PACS\_gaussian & VLCS\_gaussian & Office-Home\_gaussian\\ \midrule

& \multicolumn{3}{c}{\textit{\textbf{random samples}}} 
& \multicolumn{3}{c}{\textit{\textbf{failure cases}}} \\
\multirow{2}{*}{CLIP} & \cellcolor{gray!20} 0.961 & \cellcolor{gray!20} 0.799 &\cellcolor{gray!20} 0.741 & \cellcolor{gray!20} 0.000  &\cellcolor{gray!20} 0.000 & \cellcolor{gray!20}0.000 \\
& \cellcolor{gray!5} 1729/1800 &\cellcolor{gray!5}  1439/1800 &\cellcolor{gray!5}  1334/1800 &\cellcolor{gray!5}  0/180 &\cellcolor{gray!5}  0/180 &\cellcolor{gray!5}  0/180 \\ \midrule
\multirow{2}{*}{LLaVA} & \cellcolor{gray!20} 0.985 &\cellcolor{gray!20} 0.857 &\cellcolor{gray!20} 0.682 & \cellcolor{gray!20} 0.784 &\cellcolor{gray!20} 0.589 &\cellcolor{gray!20} 0.433 \\
& \cellcolor{gray!5} 1773/1800 & \cellcolor{gray!5} 1542/1800 & \cellcolor{gray!5} 1229/1800 & \cellcolor{gray!5} 105/134 & \cellcolor{gray!5} 106/180 & \cellcolor{gray!5} 78/180 \\ \midrule
\multirow{2}{*}{GPT-4V} & \cellcolor{gray!20} 0.972 & \cellcolor{gray!20} 0.810 &\cellcolor{gray!20} 0.874 &\cellcolor{gray!20}  0.707 &\cellcolor{gray!20} 0.568 &\cellcolor{gray!20} 0.790 \\
& \cellcolor{gray!5} 1750/1800 &\cellcolor{gray!5} 1043/1287 &\cellcolor{gray!5} 1550/1773 &\cellcolor{gray!5} 70/99 &\cellcolor{gray!5} 100/176 &\cellcolor{gray!5} 132/167 \\ \midrule
\multirow{2}{*}{Gemini} & \cellcolor{gray!20} 0.989 & \cellcolor{gray!20} 0.841 &\cellcolor{gray!20} 0.921 &\cellcolor{gray!20} 0.850 &\cellcolor{gray!20} 0.540 &\cellcolor{gray!20} 0.851 \\
& \cellcolor{gray!5} 1729/1749 &\cellcolor{gray!5} 1414/1682 &\cellcolor{gray!5} 1480/1607 &\cellcolor{gray!5} 102/120 &\cellcolor{gray!5} 81/150 &\cellcolor{gray!5} 137/161 \\
\bottomrule 
\end{tabular}}
\end{table*}

\begin{table*}[htb]
\setlength{\abovecaptionskip}{0.cm}
  \caption{Main results of zero-shot generalization performance across distribution shifts created by ControlNet. Specifically, CLIP refers to clip-vit-base-patch16, LLaVA refers to llava-v1.5-13b, Gemini refers to gemini-pro-vision, GPT-4V refers to gpt-4-vision-preview.
}
  \label{control}
\centering
\scalebox{0.83}{
\begin{tabular}{lccc|ccc}
\toprule
Dataset & PACS\_unseen & VLCS\_unseen & Office-Home\_unseen & PACS\_unseen & VLCS\_unseen & Office-Home\_unseen\\ \midrule



& \multicolumn{3}{c}{\textit{\textbf{random samples}}} 
& \multicolumn{3}{c}{\textit{\textbf{failure cases}}} \\

\multirow{2}{*}{CLIP} & \cellcolor{gray!20} 0.992  & \cellcolor{gray!20} 0.924 &\cellcolor{gray!20} 0.722 & \cellcolor{gray!20} 0.000  &\cellcolor{gray!20} 0.000 &\cellcolor{gray!20} 0.000\\
& \cellcolor{gray!5} 1786/1800 &\cellcolor{gray!5} 1633/1768 &\cellcolor{gray!5} 1299/1800 &\cellcolor{gray!5} 0/16 &\cellcolor{gray!5} 0/135 &\cellcolor{gray!5} 0/180 \\ \midrule
\multirow{2}{*}{LLaVA} & \cellcolor{gray!20} 0.996 &\cellcolor{gray!20} 0.962 &\cellcolor{gray!20} 0.618 & \cellcolor{gray!20} 0.813 &\cellcolor{gray!20} 0.726 &\cellcolor{gray!20} 0.250 \\
& \cellcolor{gray!5} 1793/1800 &\cellcolor{gray!5} 1700/1768 &\cellcolor{gray!5} 1113/1800 & \cellcolor{gray!5} 13/16 &\cellcolor{gray!5} 98/135 &\cellcolor{gray!5} 45/180\\ \midrule
\multirow{2}{*}{GPT-4V} & \cellcolor{gray!20} 0.989 & \cellcolor{gray!20} 0.932 & \cellcolor{gray!20} 0.755 & \cellcolor{gray!20} 0.875 & \cellcolor{gray!20} 0.880 & \cellcolor{gray!20} 0.611\\
& \cellcolor{gray!5} 731/739 &\cellcolor{gray!5} 1096/1176 &\cellcolor{gray!5} 935/1238 & \cellcolor{gray!5} 14/16 &\cellcolor{gray!5} 117/133 & \cellcolor{gray!5} 110/180\\\midrule
\multirow{2}{*}{Gemini} & \cellcolor{gray!20} 0.995 & \cellcolor{gray!20} 0.942 & \cellcolor{gray!20} 0.794 & \cellcolor{gray!20} 0.733 & \cellcolor{gray!20} 0.770 & \cellcolor{gray!20} 0.579\\
& \cellcolor{gray!5} 1763/1772 &\cellcolor{gray!5} 1627/1728 &\cellcolor{gray!5} 1283/1615 & \cellcolor{gray!5} 11/15 &\cellcolor{gray!5} 97/126 & \cellcolor{gray!5} 95/164\\
\bottomrule 
\end{tabular}}
\end{table*}
\subsubsection{Comparative Accuracies Across Domains}
Table~\ref{gaussian} outlines the accuracies and correct-to-total case ratios for four models (CLIP, LLaVA, Gemini, and GPT-4V) across PACS\_gaussian, VLCS\_gaussian, and Office-Home\_gaussian, incorporating both random samples and failure cases identified in CLIP. This subsection is dedicated to examining GPT-4V's zero-shot generalization abilities within datasets with distribution shifts.

\textbf{GPT-4V's Performance in Random Samples:} Focusing initially on datasets encompassing a broad range of domains and categories, like Office-Home\_gausssion, GPT-4V showcases remarkable generalization capabilities. Its impressive accuracy rate of 87.4\% in Office-Home\_gausssion is a testament to GPT-4V's adeptness in managing distribution shifts, especially those with Gaussian noise. In the PACS\_Gaussian dataset, all three models exhibit strong performance, each surpassing an accuracy rate of 95\%. This uniformity in performance hints that PACS\_gausssion might have been a part of the foundational training data for these models.

\textbf{GPT-4V in Handling CLIP's Failure Cases:} To evaluate GPT-4V's performance in more challenging scenarios, we examined its response to cases where CLIP had failed. In datasets with a wide range of categories, such as Office-Home\_gausssion, GPT-4V demonstrated significant resilience. For instance, in Office-Home\_gausssion, GPT-4V achieved an accuracy rate of 79.0\%, surpassing LLaVA's 35.7\%. In both PACS\_gausssion and VLCS\_gausssion datasets, GPT-4V consistently outperformed LLaVA. These results highlight GPT-4V's robustness in handling complex and challenging visual scenarios, even in situations where CLIP encountered difficulties.

\subsubsection{Case Demonstration}
The diverse array of case studies presented in Induced Distribution Shift: Cases 1 and 2 of the Appendix showcase the adeptness of GPT-4V and LLaVA in navigating the challenges posed by different datasets, including PACS\_gaussian, Office-Home\_gaussian, and VLCS\_gaussian. These examples not only demonstrate GPT-4V's proficiency in accurately recognizing natural distribution shifts under Gaussian noise incorporation but also highlight its ability to adapt to various visual domains and object classifications.

\subsection{Style Change with ControlNet}

\subsubsection{Comparative Accuracies Across Domains}
Table~\ref{control} outlines the accuracies and correct-to-total case ratios for three models (CLIP, LLaVA, and GPT-4V) across PACS\_unseen, VLCS\_unseen, and Office-Home\_unseen, incorporating both random samples and failure cases identified in CLIP. This subsection is dedicated to examining GPT-4V's zero-shot generalization abilities within datasets with domain shift created by ControlNet.

\textbf{GPT-4V's Performance in Random Samples:} Focusing initially on datasets encompassing a broad range of domains and categories, like Office-Home\_unseen, GPT-4V showcases remarkable generalization capabilities. Its impressive accuracy rate of 75.5\% in Office-Home\_unseen is a testament to GPT-4V's adeptness in managing distribution shifts created by ControlNet. In the PACS\_unseen and VLCS\_unseen, all three models exhibit strong performance, each surpassing an accuracy rate of 90\%. This uniformity in performance hints that PACS\_unseen and VLCS\_unseen might have been a part of the foundational training data for these models.

\textbf{GPT-4V in Handling CLIP's Failure Cases:} To evaluate GPT-4V's performance in more challenging scenarios, we examined its response to cases where CLIP had failed. In datasets with a wide range of categories, such as Office-Home\_unseen, GPT-4V demonstrated significant resilience. For instance, in Office-Home\_unseen, GPT-4V achieved an accuracy rate of 61.1\%, surpassing LLaVA's 25.0\%. In both PACS\_unseen and VLCS\_unseen datasets, GPT-4V consistently outperforms LLaVA. These results highlight GPT-4V's robustness in handling challenging visual scenarios, even in situations where CLIP encountered difficulties.

\subsubsection{Case Demonstration}
\begin{figure*}[htb!]
\centering
\includegraphics[width=0.96\textwidth]{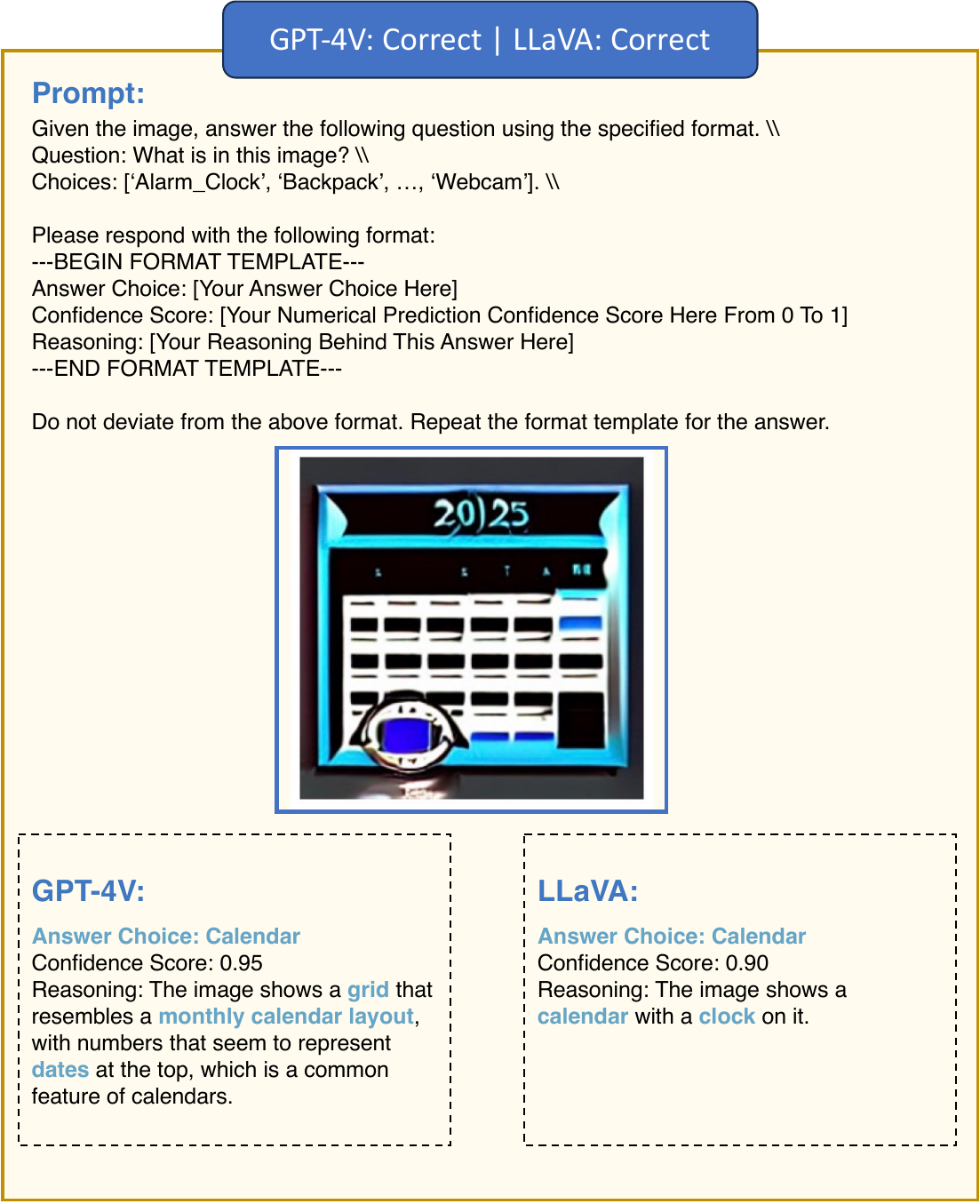}
\caption[Induced Distribution Shift: Case 3 on PACS\_unseen]{Induced Distribution Shift: Case 3 - Analyzing the `dog' Category in the art\_painting Domain of the PACS\_unseen Dataset. In this case study, both GPT-4V and LLaVA models are tasked with responding to an identical text prompt accompanied by an image. Both GPT-4V and LLaVA predicted correctly, however, GPT-4V gave a confidence level of 1.00 and LLaVA did not give a confidence level. Both GPT-4V and LLaVA give a more nuanced and reliable reason.}
\label{PACS-new-random}
\end{figure*}

\begin{figure*}[htb!]
\centering
\includegraphics[width=0.96\textwidth]{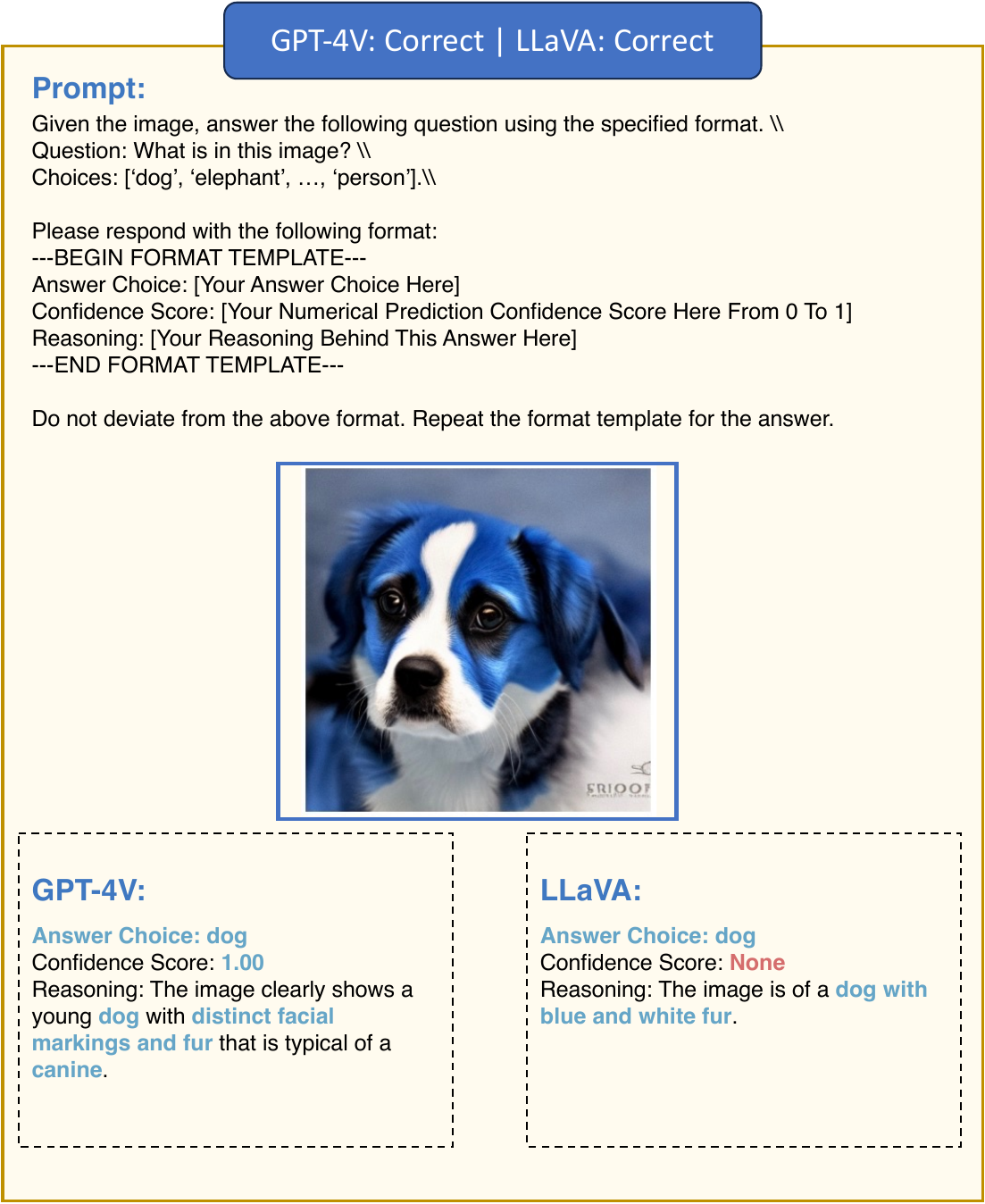}
\caption[Induced Distribution Shift: Case 4 on Office-Home\_unseen]{Natural Distribution Shift: Case 4 - Analyzing the `Calendar' Category in the Clipart Domain of the Office-Home\_unseen Dataset. In this case study, both GPT-4V and LLaVA models are tasked with responding to an identical text prompt accompanied by an image. Both GPT-4V and LLaVA predicted correctly, but GPT-4V gave higher confidence. GPT-4V and LLaVA focus on different details. GPT-4V focuses on the grid and dates associated with Calendar, whereas LLaVA focuses on another object `clock' in the image.}
\label{officehome-new-random}
\end{figure*}

\begin{figure*}[htb!]
\centering
\includegraphics[width=0.96\textwidth]{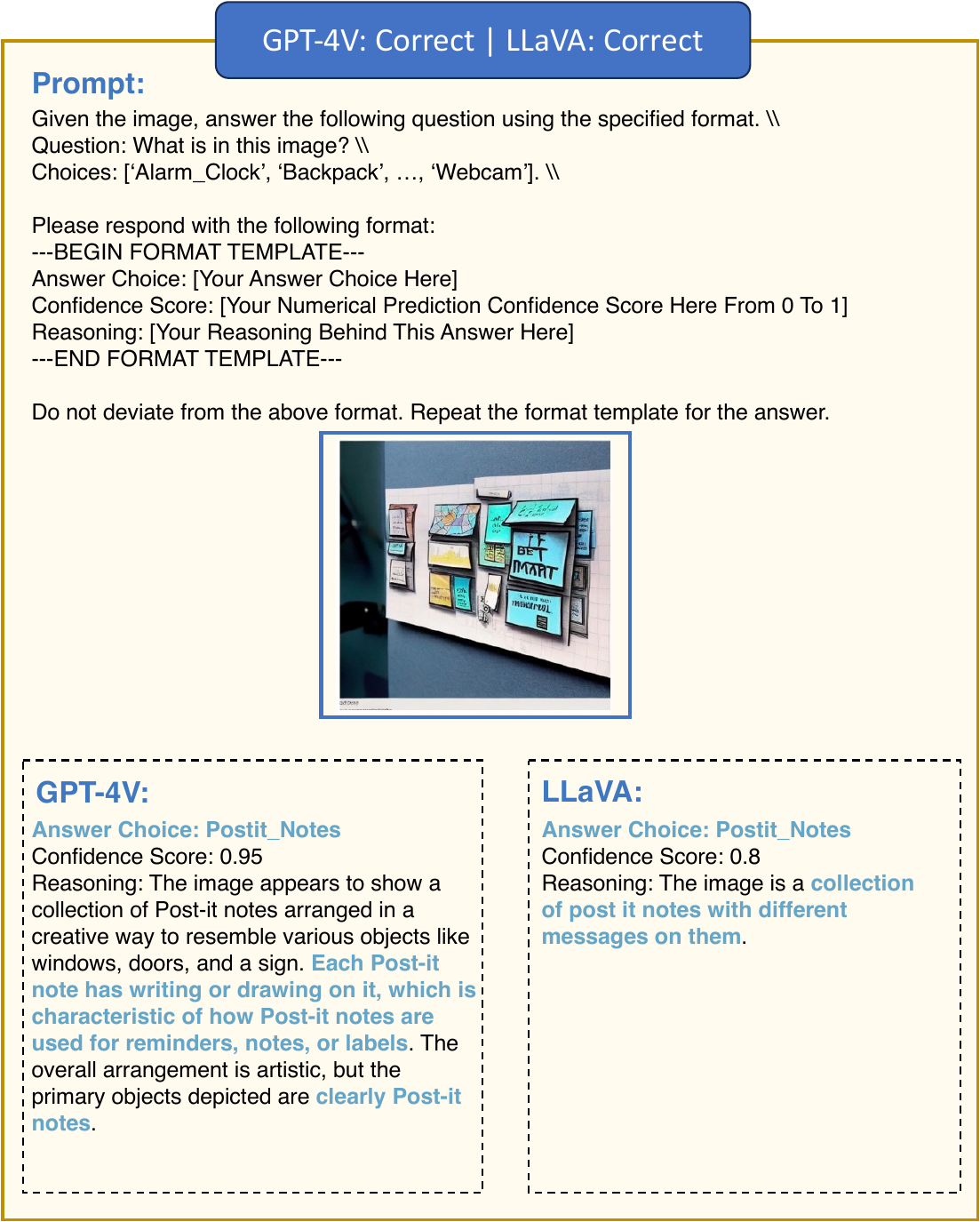}
\caption[Induced Distribution Shift: Case 5 on Office-Home\_unseen]{Natural Distribution Shift: Case 5 - Analyzing the `Postit\_Notes' Category in the Art Domain of the Office-Home\_unseen Dataset. In this case study, both GPT-4V and LLaVA models are tasked with responding to an identical text prompt accompanied by an image. Both GPT-4V and LLaVA predicted correctly, but GPT-4V gave higher confidence with more detailed description. GPT-4V focuses on the writing and drawing associated with Postit\_Notes, whereas LLaVA focuses on massages on it. This also demonstrates that GPT-4V and LLaVA have a certain degree of generalization ability on unseen data with domain shifts, with GPT-4V possessing stronger explanatory capabilities.}
\label{Office-Home-unseen1}
\end{figure*}

\begin{figure*}[htb!]
\centering
\includegraphics[width=0.96\textwidth]{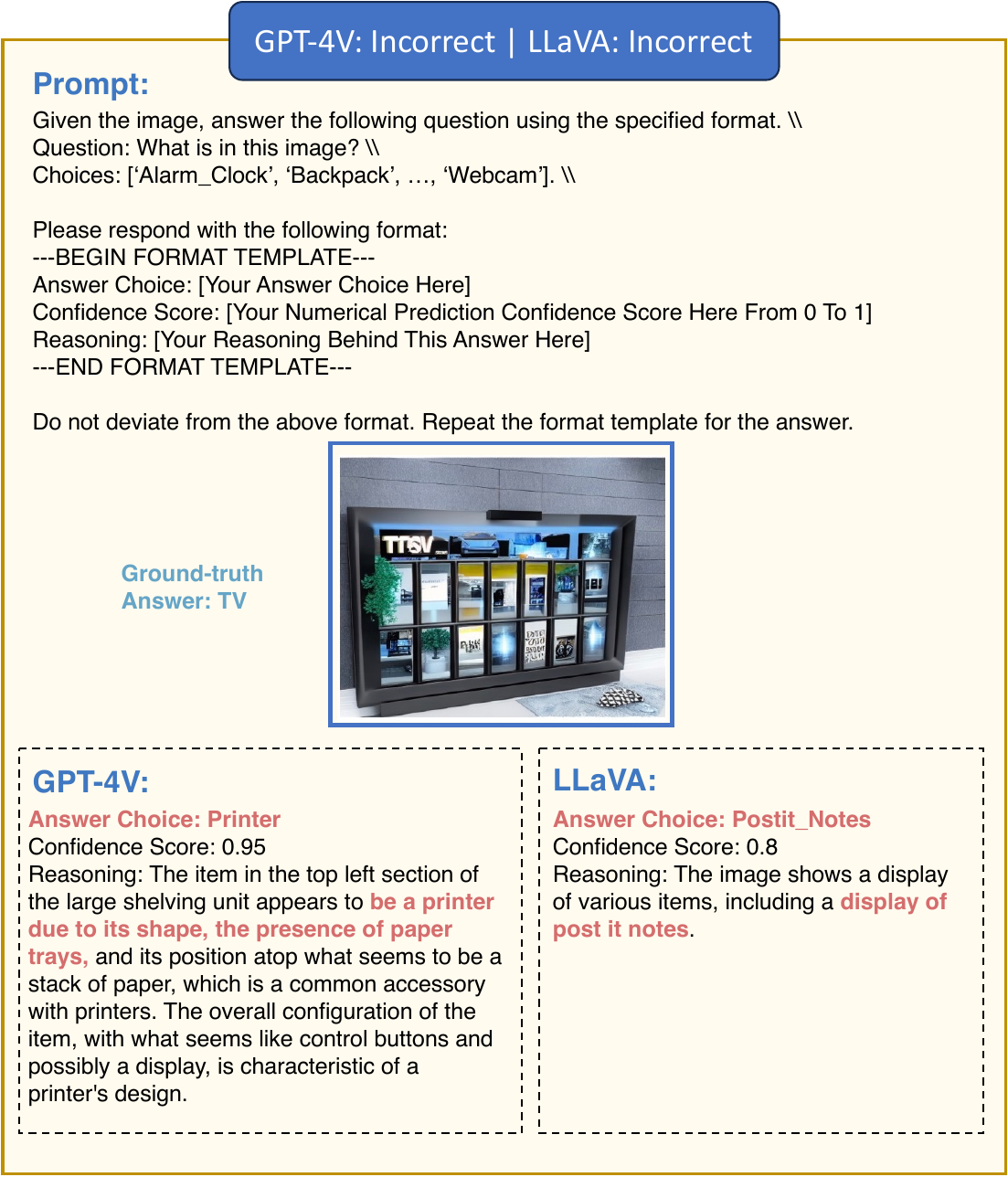}
\caption[Induced Distribution Shift: Case 6 on Office-Home\_unseen]{Natural Distribution Shift: Case 6 - Analyzing the `TV' Category in the Real World Domain of the Office-Home\_unseen Dataset. This image presents a TV displaying complex content. Due to the misleading nature of the complex content, both GPT-4V and LLaVA made mistakes. GPT-4V misidentified the TV as a Printer, while LLaVA misidentified it as Post-it Notes. This result demonstrates that both GPT-4V and LLaVA still have difficulties in predicting complex samples accurately.
}
\label{Office-Home-unseen2}
\end{figure*}

\begin{figure*}[htb!]
\centering
\includegraphics[width=0.96\textwidth]{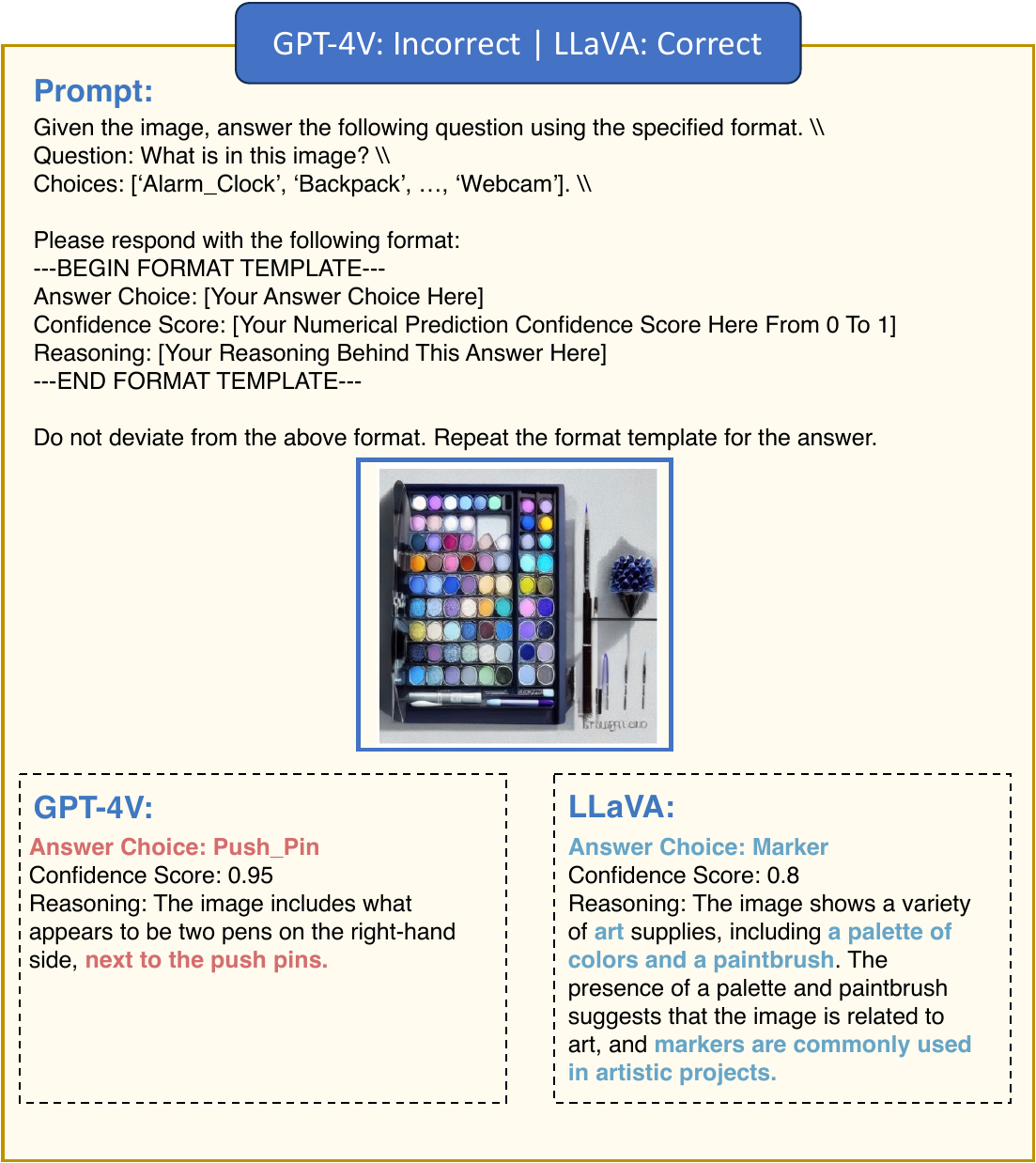}
\caption[Induced Distribution Shift: Case 7 on Office-Home\_unseen]{Natural Distribution Shift: Case 7 - Analyzing the `Marker' Category in the Art Domain of the Office-Home\_unseen Dataset. In this case, LLaVA correctly identifies the object as a `Marker' with a confidence score of 0.8, whereas GPT-4V, with a confidence score of 0.95, mistakenly identifies the object as a `Push\_Pin'. Due to the presence of an object resembling a Push\_Pin in the image, GPT-4V identified the image as a Push\_Pin. Meanwhile, not only did LLaVA correctly predict, but it also provided a description related to its prediction: a palette of colors and a paintbrush.
}
\label{Office-Home-unseen3}
\end{figure*}

\begin{figure*}[htb!]
\centering
\includegraphics[width=0.96\textwidth]{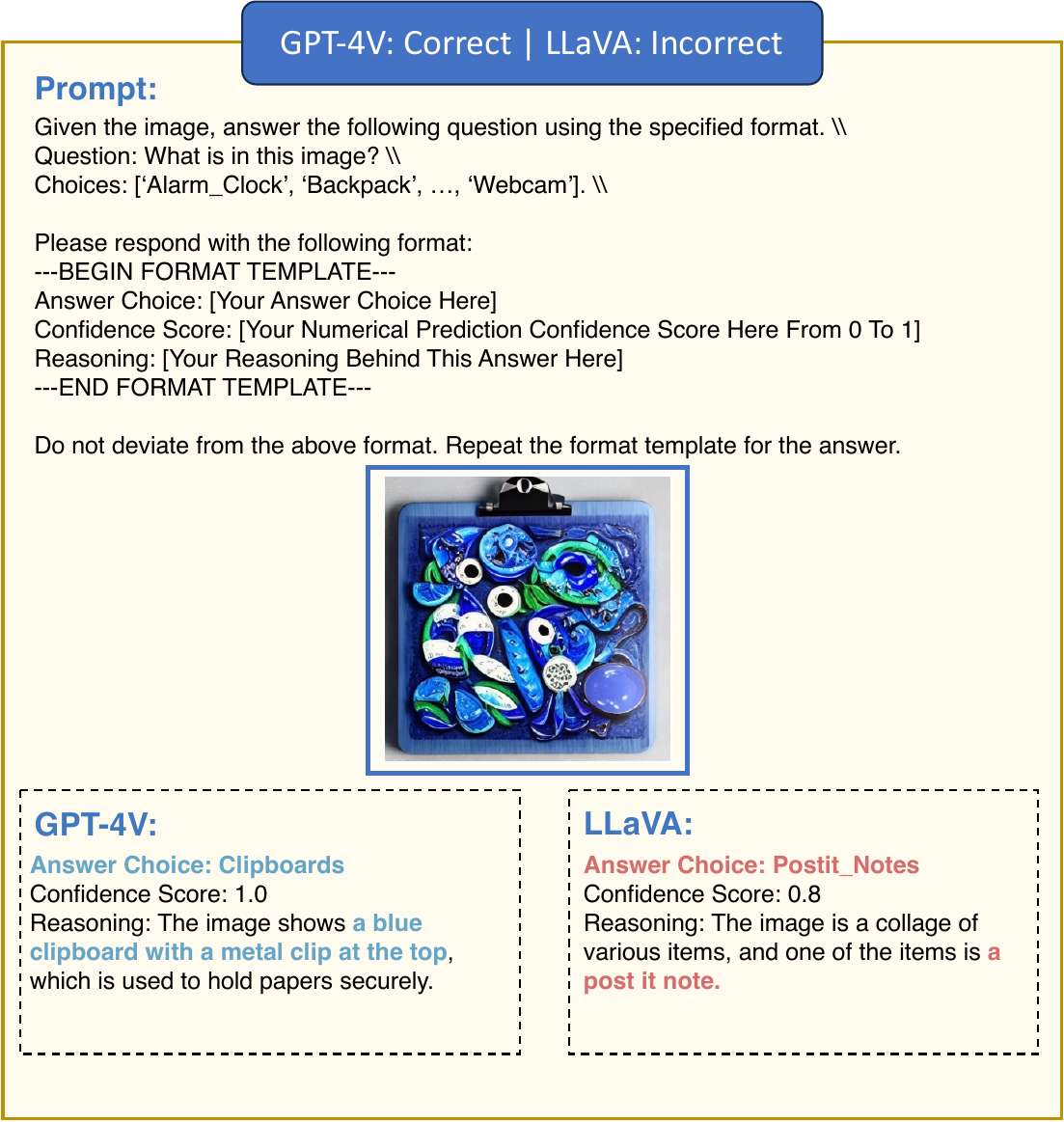}
\caption[Induced Distribution Shift: Case 8 on Office-Home\_unseen]{Natural Distribution Shift: Case 8 - Analyzing the `Clipboards' Category in the Clipart Domain of the Office-Home\_unseen Dataset. In this image, GPT-4V accurately identifies the object as Clipboards, noting a metal clip at the top of object, and assigning a confidence score of 1.0. GPT-4V successfully captured the key element `clip,' which helped in identifying the object as Clipboards. In contrast, LLaVA incorrectly classifies the object as Postit\_Notes with a confidence score of 0.8, failing to recognize the key element `clip' of Clipboards.}
\label{Office-Home-unseen444}
\end{figure*}

The diverse array of case studies presented in Figure~\ref{officehome-new-random}, \ref{PACS-new-random}, and \ref{Office-Home-unseen1} showcase the adeptness of GPT-4V and LLaVA in navigating the challenges posed by different datasets, including PACS\_unseen, Office-Home\_unseen, and VLCS\_unseen. These examples not only demonstrate GPT-4V's proficiency in accurately recognizing natural distribution shifts created by ControlNet incorporation but also highlight its ability to adapt to various visual domains and object classifications. However, under certain complex samples, such as Figure~\ref{Office-Home-unseen2}, \ref{Office-Home-unseen3}, and \ref{Office-Home-unseen444}, both GPT-4V and LLaVA still have their limitations. They are prone to being misled by irrelevant factors in the image, leading to incorrect predictions. 

\section{Exploiting In-Context Learning for Domain Bridging}
\label{018520640299}

Addressing distribution shifts traditionally involves fine-tuning pre-trained foundational models with source domain data to facilitate effective adaptation to target domains. While this approach can be effective, it often requires significant computational resources and time, especially for large foundational models \citep{hu2021lora}. 
Against this backdrop, our research shifts focus to the exploration of in-context learning capabilities of large multimodal models, with a specific emphasis on GPT-4V. This approach presents a novel method for simulating traditional domain generalization paradigms.

In-context learning, as defined by GPT-3 \citep{brown2020language}, involves conditioning the model on a set of natural language instructions alongside a few task demonstrations. 
The model is then expected to apply this learned context to complete further instances of the task, primarily through predicting subsequent sequences. 
This methodology leverages the model's inherent ability to infer and apply patterns from limited information without any parameter update, a significant difference from conventional fine-tuning techniques. 
This ability of large foundation models to demonstrate emergent capabilities through in-context learning has been increasingly recognized and highlighted in recent studies \citep{wei2022chain,ouyang2022training,wei2022emergent,wang2022self,kojima2022large}. 
Our study aims to assess how effectively GPT-4V utilizes in-context learning to navigate distribution shifts across diverse domains~\citep{ahuja2023closer,gupta2023context}. 

\subsection{In-context Setup}

For our in-context learning exploration, we focus on the Camelyon17~\citep{bandi2018detection}, COVID~\citep{han2021semi}, DrugOOD\_Assay~\citep{ji2023drugood} and NIH-Chest~\citep{wang2017chestx} datasets. 
These datasets were chosen due to GPT-4V's previously observed underperformance, perhaps because the pre-training data distribution rarely includes scientific datasets like medical and protein.
We wish the in-context learning that simulates conventional domain adaptation/generalization would enhance adaptability to certain tasks.
In our experimental setup, we randomly select two classes within two domains of each dataset, designating them as source and target domains.
From the source domain, we choose two representative examples for each class, like normal and typical pneumonia in the COVID dataset or normal and tumor in the Camelyon17 dataset, as illustrated in Figure \ref{351022164706}. 
To demonstrate the potential of in-context learning as an effective approach for adapting large multimodal models to distribution shifts, we have intentionally limited our experiment to just one source domain and two examples. 
This decision is primarily driven by the constraints related to token cost.
This setup emulates the concept of traditional out-of-distribution generalization but contrasts with it by leveraging the model's innate ability to adapt to new contextual information while maintaining its original parameterization \citep{brown2020language}.

Below, we illustrate an example of an in-context prompt applied to the Camelyon17 dataset. 
This dataset is distinguished by its binary classification system, encompassing two distinct classes: `normal' and `tumor'. 
In contrast to the basic prompt in Figure \ref{291135271633}, we explicitly annotate the class labels for the two in-context examples provided to GPT-4V, i.e., `The first image is normal and the second image is tumor'. 
Furthermore, the prompt's inquiry is subtly altered to `What is the third image?', thereby aligning the model's focus with the specific task of classification based on the provided contextual examples.
The response format template is set the same as the previous basic prompt.

\begin{center}
\fbox{%
\begin{minipage}{0.97\linewidth}
\textbf{Text Prompt with In-Context Examples:\\ }
Given the image, answer the following question using the specified format. \\
\textit{The first image is \{class\_1\} and the second image is \{class\_2\}.} \\
Question: What is the third image? \\
Choices:[‘class\_1', ‘class\_2']. \\
Please respond with the following format:\\
... \\
\end{minipage}%
}    
\end{center}

\begin{figure}[htb!]
\centering
\includegraphics[width=.9\textwidth]{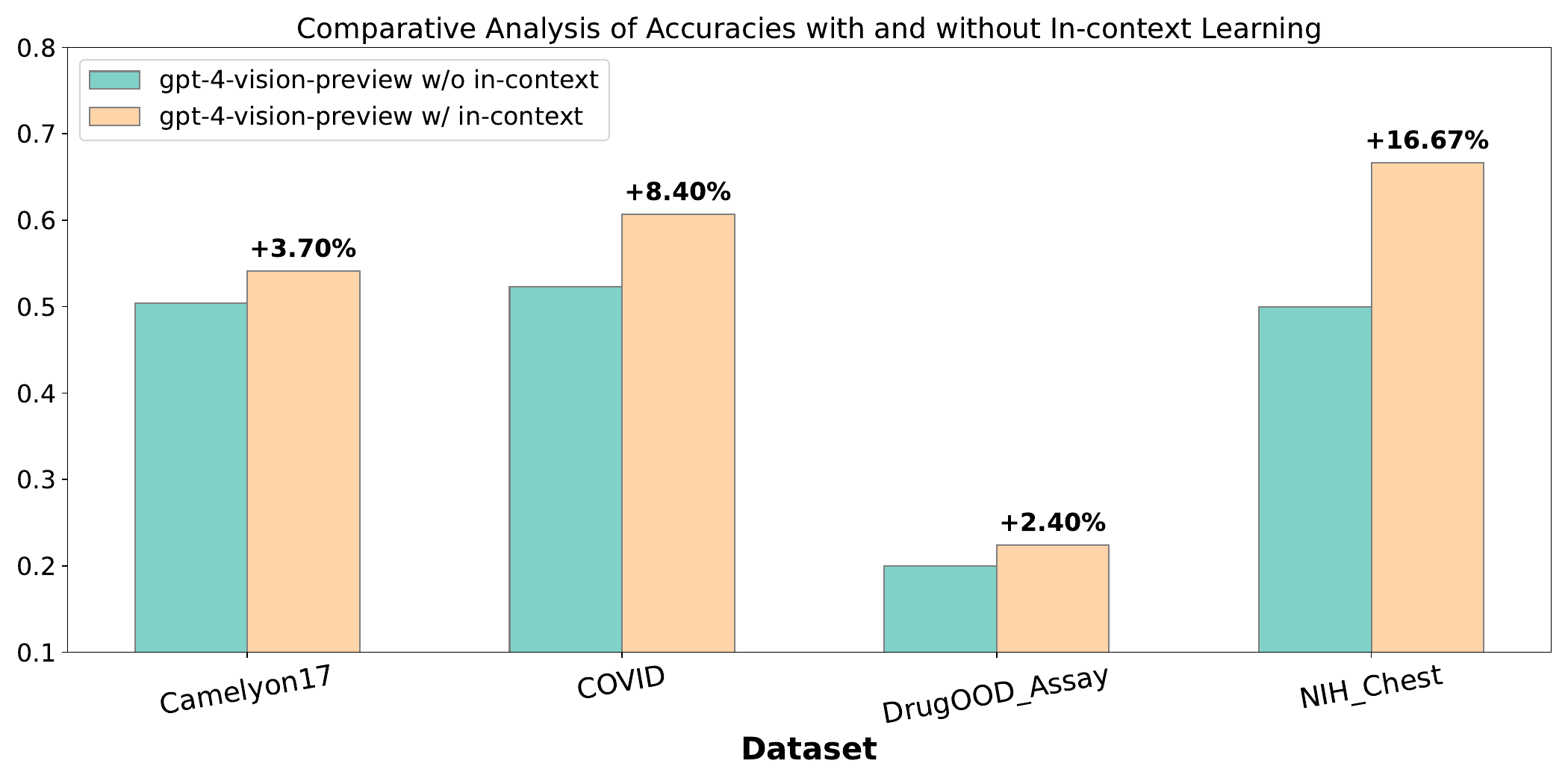}
\caption[In-context learning on GPT-4V for Domain Bridging]
{Improvements in target domain performance with in-context learning on GPT-4V across Camelyon17, COVID, DrugOOD\_Assay and NIH\_Chest datasets.}
\label{580835271317}
\end{figure}

\subsection{In-context Performance} 

In Figure~\ref{580835271317}, we illustrate the impact of in-context learning when applied to the baseline GPT-4V model, specifically within the target domain. 
This approach demonstrates consistent performance enhancements across four distinct datasets. 
In particular, the application of in-context learning yields improvements of 3.7\%, 8.4\%, 2.4\%, and 16.67\% for the Camelyon17, COVID, DrugOOD\_Assay, and NIH\_Chest datasets, respectively. 
These results highlight the potential of in-context learning in boosting model adaptability, especially in situations characterized by distribution shifts.

The observed variability in performance gains across these datasets suggests a correlation between the inherent task complexity and the unique data distributions of each dataset. 
This aspect of the results prompts further investigation into the nuances of in-context learning and its differential impact based on dataset characteristics.

In our experimental setup, two examples were randomly selected from the source domain for the in-context learning process. 
However, a more deliberate selection of in-context examples could potentially lead to even greater improvements in model performance \citep{huang2023machine}. 
This possibility opens avenues for future research, where the strategic choice of in-context examples could be explored as a means to optimize the efficacy of in-context learning.

 \subsection{In-context Case Demonstration}
 This section showcases selected cases to demonstrate the enhancement of inference performance through in-context examples.
 
\begin{figure}[tb!]
\centering
\includegraphics[width=.84\textwidth]{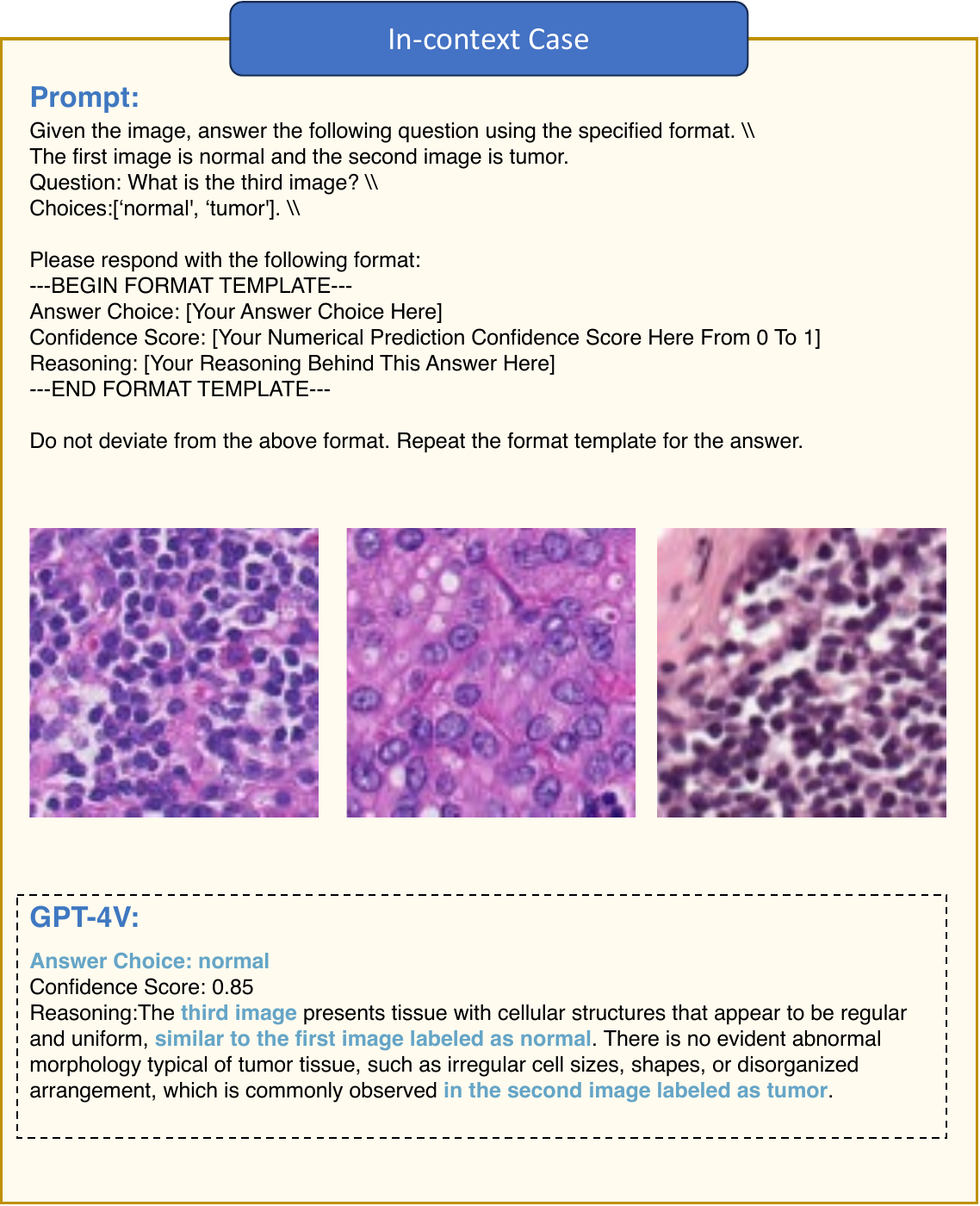}
\caption[In-context Case Demonstration: Case 1 on Camelyon17]
{Demonstration of GPT-4V's inference process when exposed to in-context learning with examples from the Camelyon17 dataset.
The experiment involves using two representative images from the source domain (hospital\_2), one labeled `normal' and the other `tumor', followed by a test image from the target domain (hospital\_3).
GPT-4V, conditioned with these in-context examples, distinguishes between regular and uniform tissue patterns in the `normal' image and abnormal, irregular cell sizes in the `tumor' image. 
It then applies this contextual understanding to accurately infer the class of the test image from hospital\_3. 
This process showcases GPT-4V's ability to leverage in-context cues for effective domain bridging.}
\label{351022164706}
\end{figure}

 \textbf{GPT-4V’s Interpretation of In-context Examples:}
Figure~\ref{351022164706} features a case study within the Camelyon17 dataset. The procedure includes presenting GPT-4V with two annotated images from a source domain (hospital\_2): one denoted as 'normal' and the other as 'tumor'. 
These are followed by a test image from a different domain (hospital\_3). 
Conditioned with this contextual information, GPT-4V effectively discerns between the regular, uniform tissue patterns in the 'normal' image and the abnormal, irregular cell structures in the 'tumor' image. 
It then applies this discernment to precisely classify the test image from hospital\_3. 
This case exemplifies how GPT-4V employs in-context examples to bridge different domains, enhancing its interpretive accuracy.

\begin{figure}[hb!]
\centering
\includegraphics[width=.84\textwidth]{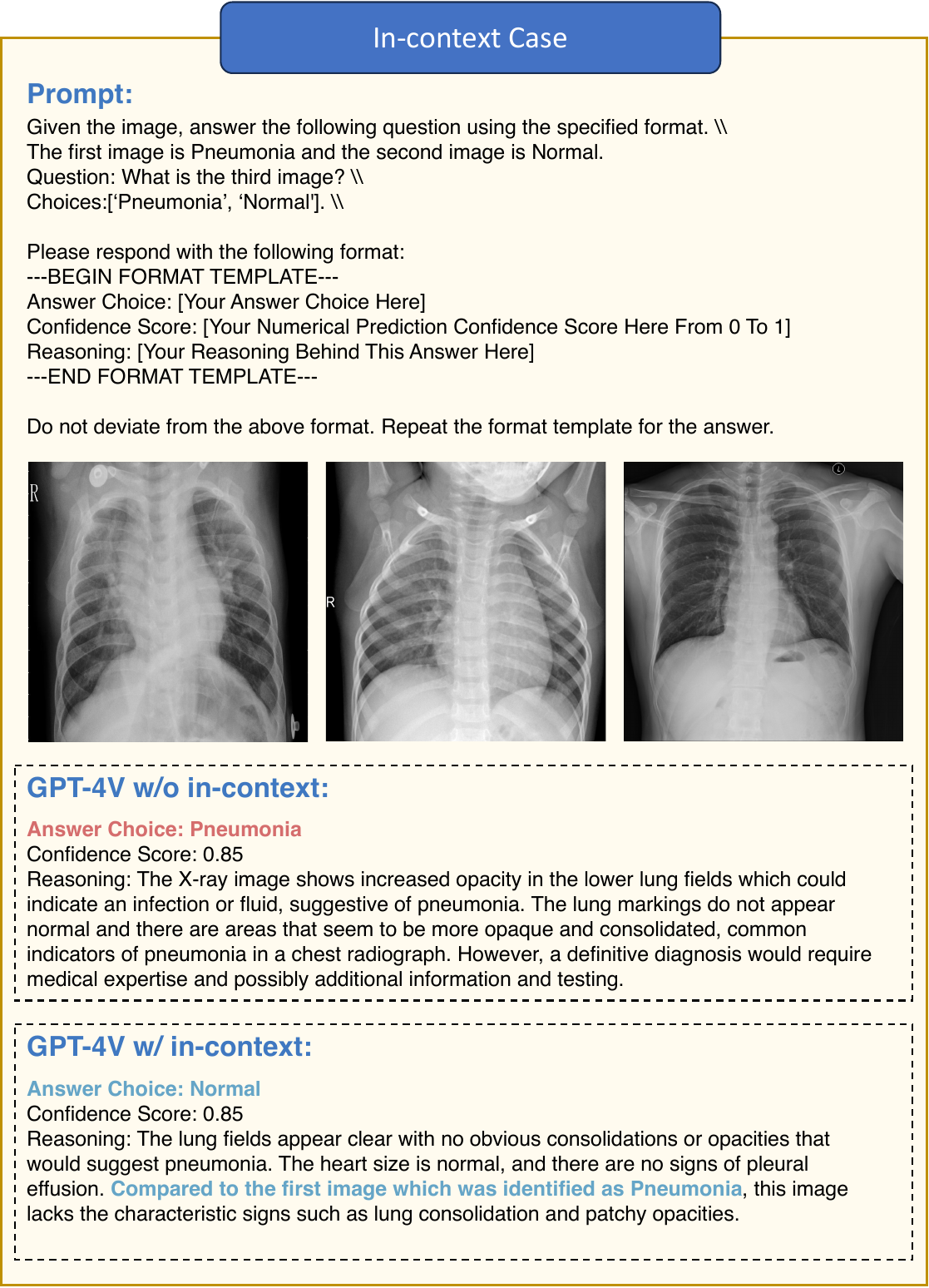}
\caption[In-context Case Demonstration: Case 2 on COVID]
{
This figure illustrates GPT-4V's inference process under in-context learning, using selected examples from the COVID dataset. 
The process involves initially presenting two annotated images from the source domain: one labeled as 'Pneumonia' and the other as 'Normal.' Subsequently, GPT-4V is tasked with classifying a test image from the target domain. The model, having been conditioned with these specific in-context examples, evaluates and categorizes the test image by drawing comparisons to the previously presented pneumonia and normal X-ray images.}
\label{447670842416}
\end{figure}

\textbf{The Impact of In-context Examples:}
Figure~\ref{447670842416} explores the influence of in-context learning on GPT-4V's performance in classifying chest X-ray images. 
The figure presents a comparative analysis of the model’s accuracy with and without in-context learning. 
Initially, GPT-4V incorrectly classifies a test image as `Pneumonia’ with a confidence score of 0.85, when no contextual information is provided. 
However, when conditioned with two in-context examples from the source domain, one labeled 'Pneumonia' and the other `Normal,' the model's performance shifts markedly. 
With in-context learning, the model compares the third image with the first 'Pneumonia' figure and accurately categorizes the same test image as 'Normal' with an identical confidence score. 
This stark difference underscores the significant role that in-context learning plays in enhancing the model's diagnostic precision, particularly in discerning subtle distinctions in medical imaging.
 
\section{Conclusion}
Our investigation into the adaptability and generalization capabilities of GPT-4V, a leading multimodal foundation model, marks a significant advancement in our understanding of AI systems' robustness against distribution shifts. Through rigorous evaluation and comparison with models like CLIP, LLaVA, and Gemini across 13 diverse datasets in natural, medical, and molecular domains, we have delineated the capability boundaries of GPT-4V, uncovering both its strengths and limitations in various complex scenarios. Our findings reveal that while GPT-4V demonstrates notable adaptability and zero-shot generalization capabilities, its performance varies significantly across different scenarios of distribution shifts. 
This variation underscores the importance of continuous assessment and enhancement of foundation models to cope with evolving data landscapes. While we have made significant strides in understanding and improving the adaptability of foundation models like GPT-4V, our journey toward creating truly robust and versatile AI foundation models is ongoing.

\clearpage

\begin{figure*}[htb!]
\centering
\includegraphics[width=0.96\textwidth]{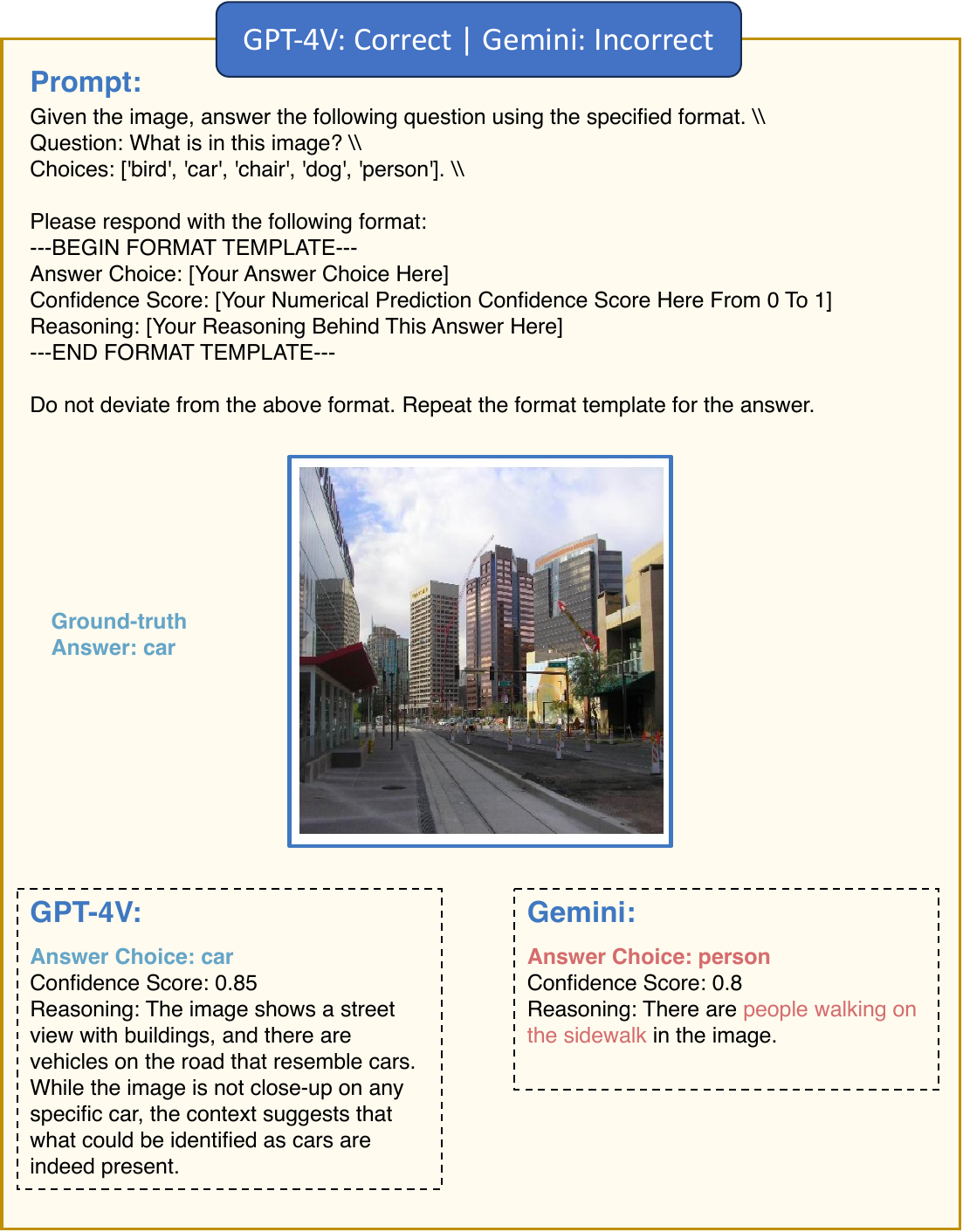}
\caption[Natural Distribution Shift: Case 22 on VLCS]{Natural Distribution Shift: Case 22 - Analyzing the `Car' Category in the SUN09 Domain of the VLCS Dataset. This instance illustrates a challenging scenario where GPT-4V accurately identifies the presence of cars within a street view, noting vehicles on the road amidst buildings with a confidence score of 0.85. Conversely, Gemini, with a confidence score of 0.8, incorrectly identifies a person, focusing on individuals walking on the sidewalk. This comparison highlights the nuanced differences in model perception and interpretation within complex urban environments, emphasizing the critical role of context in AI's visual comprehension.}
\label{711867515269}
\end{figure*}

\begin{figure*}[htb!]
\centering
\includegraphics[width=0.96\textwidth]{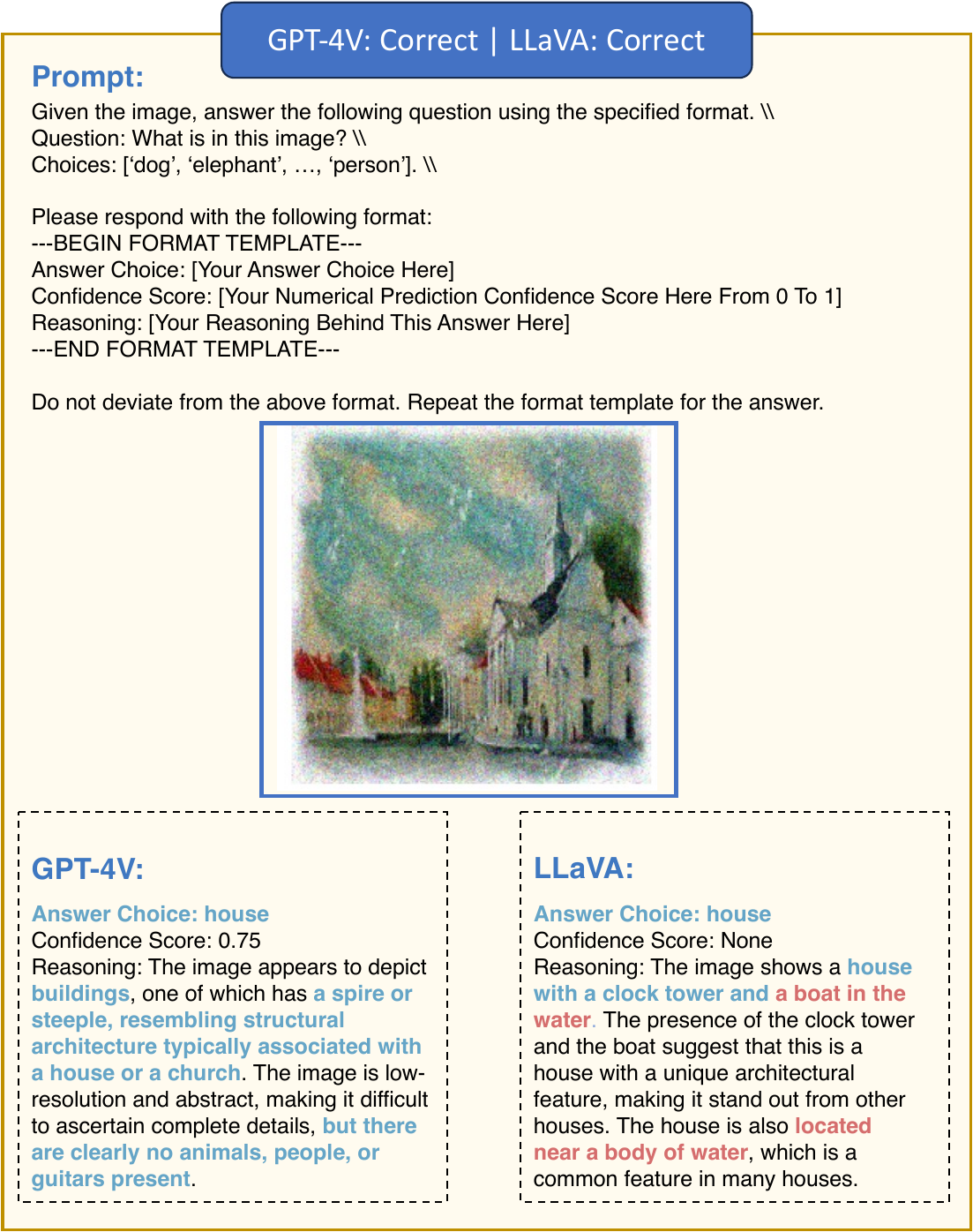}
\caption[Induced Distribution Shift: Case 1 on Office-Home\_gaussion]{Induced Distribution Shift: Case 1 - Analyzing the `Fork' Category in the Product Domain of the Office-Home\_gaussion Dataset. In this case study, both GPT-4V and LLaVA models are tasked with responding to an identical text prompt accompanied by an image. Both GPT-4V and LLaVA predicted correctly, but GPT-4V gave higher confidence levels as well as more detailed explanations such as tines, a handle, and several pointed prongs. These are the basic characteristics of a fork. However, the reason given by LLaVA is rough.}
\label{Office-Home-gaussion-random_case}
\end{figure*} 

\begin{figure*}[htb!]
\centering
\includegraphics[width=0.96\textwidth]{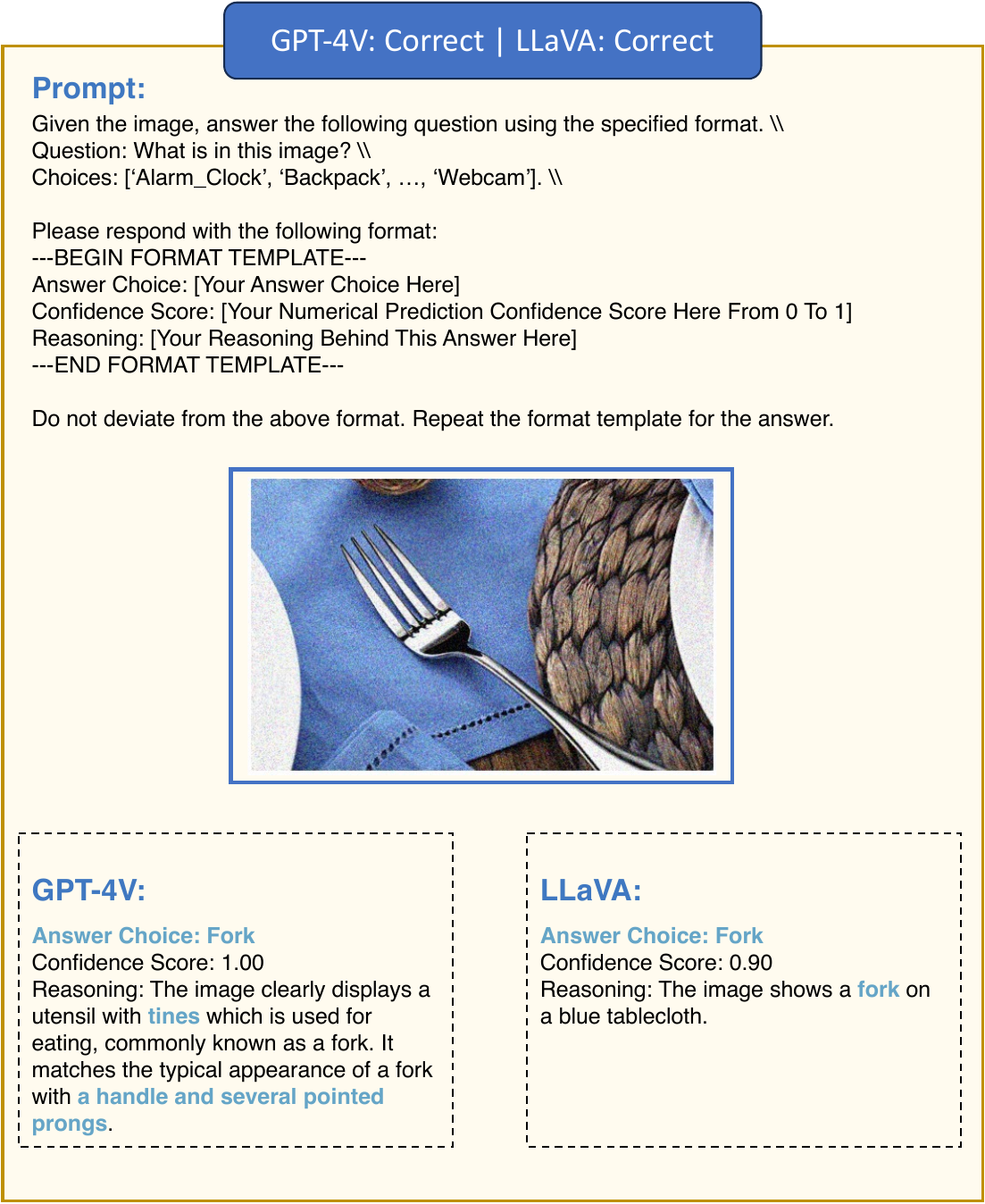}
\caption[Induced Distribution Shift: Case 2 on PACS\_gaussion]{Induced Distribution Shift: Case 2 - Analyzing the `house' Category in the art\_painting Domain of the PACS\_gaussion Dataset. In this case study, both GPT-4V and LLaVA models are tasked with responding to an identical text prompt accompanied by an image. Both GPT-4V and LLaVA predicted correctly, however, GPT-4V gave a confidence level of 0.75 and LL a VA did not give a confidence level. GPT-4V gave some more detailed information in the reason, such as spire and steeple. On the contrary, LLaVA gave a partially incorrect description in the reason, such as boat and water.}
\label{PACS-gaussion-random_case}
\end{figure*} 



\clearpage

\bibliographystyle{plainnat}
\bibliography{references} 

\end{document}